\documentclass[10pt,twocolumn,letterpaper]{article}

\usepackage[preprint]{cvpr}      

\usepackage[table,usenames,dvipsnames]{xcolor} 

\usepackage{graphicx} 
\usepackage{caption} 
\usepackage[framemethod=TikZ]{mdframed} 
\usepackage{helvet}
\usepackage[most]{tcolorbox} 
\usepackage{colortbl} 
\usepackage{multirow} 
\usepackage{enumitem}
\usepackage{fontawesome}

\usepackage{mdframed}

\newcommand{\increasenoparent}[1]{\textcolor{ForestGreen}{+#1}}

\newcommand{\decreasenoparent}[1]{\textcolor{red}{-#1}}

\newcommand{\changelinkcolor}[1]{\hypersetup{linkcolor=#1}}

\newcommand{\CameraReady}{\textcolor{black}}
\usepackage{colortbl, xcolor}

\definecolor{shadecolor}{gray}{0.9} 
\definecolor{examplegray}{gray}{0.95} 
\definecolor{lightgray}{gray}{0.9}
\definecolor{darkblue}{RGB}{31, 78, 121}   
\definecolor{darkgreen}{RGB}{78, 121, 31}  
\definecolor{lightgray}{RGB}{236,236,236}  
\definecolor{framecolor}{RGB}{100,100,100} 
\definecolor{darkgray}{RGB}{64,64,64}
\definecolor{darkred}{RGB}{139,0,0}
\definecolor{darkyellow}{RGB}{236,165,40}

\definecolor{lightblue}{RGB}{173,216,230} 
\definecolor{skyblue}{RGB}{135,206,235}   
\definecolor{steelblue}{RGB}{70,130,180}  

\definecolor{apricot}{RGB}{255,224,178}   
\definecolor{orange}{RGB}{255,165,0}      
\definecolor{darkorange}{RGB}{255,140,0}  

\definecolor{lightgreen}{RGB}{144,238,144} 
\definecolor{olivedrab}{RGB}{107,142,35}  
\definecolor{forestgreen}{RGB}{34,139,34} 

\definecolor{lightgrey}{RGB}{211,211,211} 
\definecolor{darkgrey}{RGB}{169,169,169}  
\definecolor{dimgrey}{RGB}{105,105,105}   

\definecolor{lightpurple}{RGB}{230, 230, 250} 

\newtcolorbox{ebox}[2][]{
  colback=lightgray,
  colframe=darkgray,
  colbacktitle=darkgray,
  coltitle=white,
  title=#2,
  #1
}

\usepackage{pifont} 

\newcommand{\colorrect}[1]{\textcolor{#1}{\ding{110}}}

\usepackage[dvipsnames]{xcolor}

\definecolor{cvprblue}{rgb}{0.21,0.49,0.74}
\usepackage[pagebackref,breaklinks,colorlinks,citecolor=cvprblue]{hyperref}

\def\paperID{10865} 
\def\confName{CVPR}
\def\confYear{2024}

\newcommand{\datasetname}{\texttt{\textbf{Glitch}}Bench\xspace}

\title{ \datasetname: Can large multimodal models detect video game glitches? }

\author{%
Mohammad Reza Taesiri\(^{1}\), Tianjun Feng\(^{1}\), Anh Totti Nguyen\(^{2}\), Cor-Paul Bezemer\(^{1}\)\\
{\small \(^{1}\)University of Alberta, \texttt{\{mtaesiri, robbie020428, bezemer\}@ualberta.ca}}\\
{\small \(^{2}\)Auburn University, \texttt{anh.ng8@gmail.com}}
}

\begin{document}
\maketitle
\begin{abstract}

Large multimodal models (LMMs) have evolved from large language models (LLMs) to integrate multiple input modalities, such as visual inputs.
This integration augments the capacity of LLMs for tasks requiring visual comprehension and reasoning.
However, the extent and limitations of their enhanced abilities are not fully understood, especially when it comes to real-world tasks.
To address this gap, we introduce \datasetname, a novel benchmark derived from video-game quality assurance tasks, to test and evaluate the reasoning capabilities of LMMs.
Our benchmark is curated from a variety of unusual and glitched scenarios from video games and aims to challenge both the visual and linguistic reasoning powers of LMMs in detecting and interpreting out-of-the-ordinary events.
Our evaluation shows that \datasetname presents a new, interesting challenge to  state-of-the-art LMMs.
Code and data are available at: \href{https://glitchbench.github.io/}{https://glitchbench.github.io/}
\end{abstract}
    
\section{Introduction}
\label{sec:intro}

\begin{figure}[t] 
\begin{ebox}{GPT-4V fails to focus on the rain inside the room}
\centering
\includegraphics[width=\textwidth]{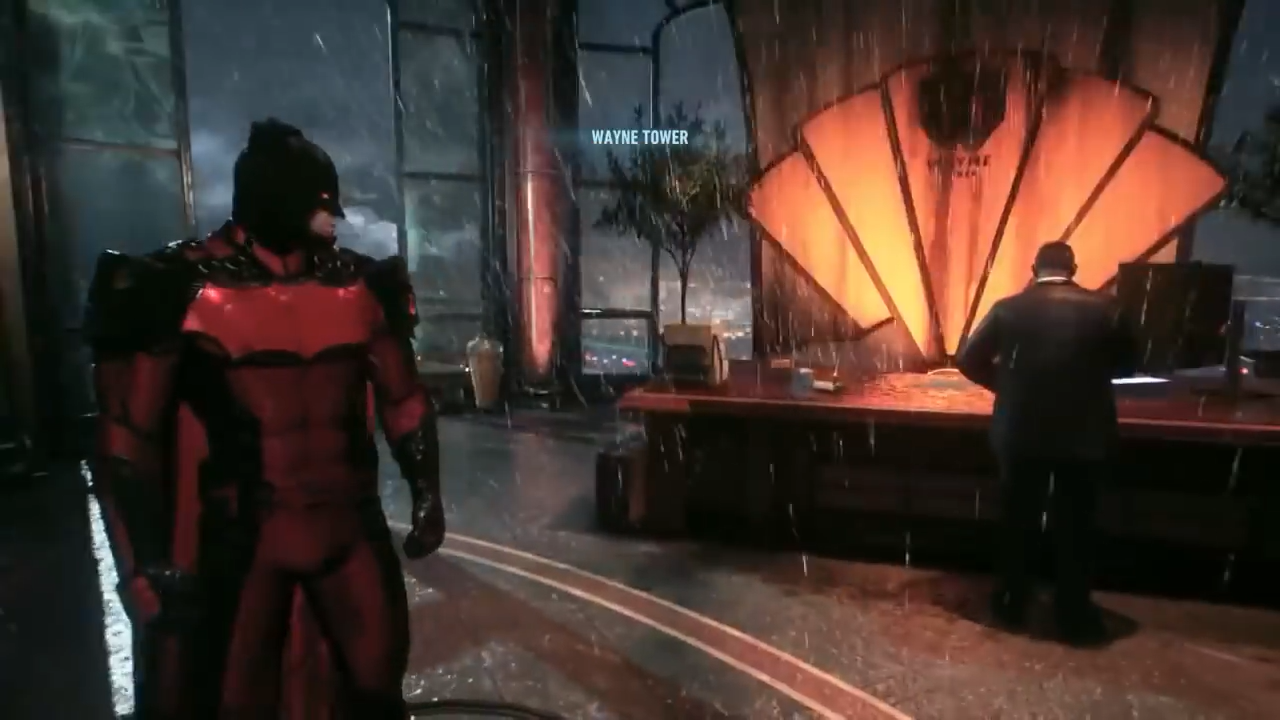}
\raggedright
\textbf{\textcolor{darkblue}{Question:}} What is wrong with this image?
\\
\textbf{\textcolor{darkgreen}{Ground truth:}} It is raining inside the building
\\
\textbf{\textcolor{darkred}{GPT-4V:}} 
There's nothing inherently ``wrong'' with this image; however, if you're referring to inconsistencies or peculiarities that might not align with the established Batman canon, the red Batman suit is unusual since Batman is traditionally known in black or very dark colors. The use of red could be an artistic choice for the game, an alternate suit available within the game's universe, or indicative of a different character who takes on the Batman mantle.
\end{ebox}
\caption{The image depicts a screenshot in which it rains inside a room. While the rain should be what is wrong with the image, GPT-4V fails to reason correctly and instead focuses on the color of Batman's costume. Note that the ground truth is never presented as part of the prompt in our study.}
\label{fig:teaser}
\end{figure}

\begin{figure*}[t!]
\centering

\begin{subfigure}{0.325\textwidth}
  \includegraphics[width=\linewidth]{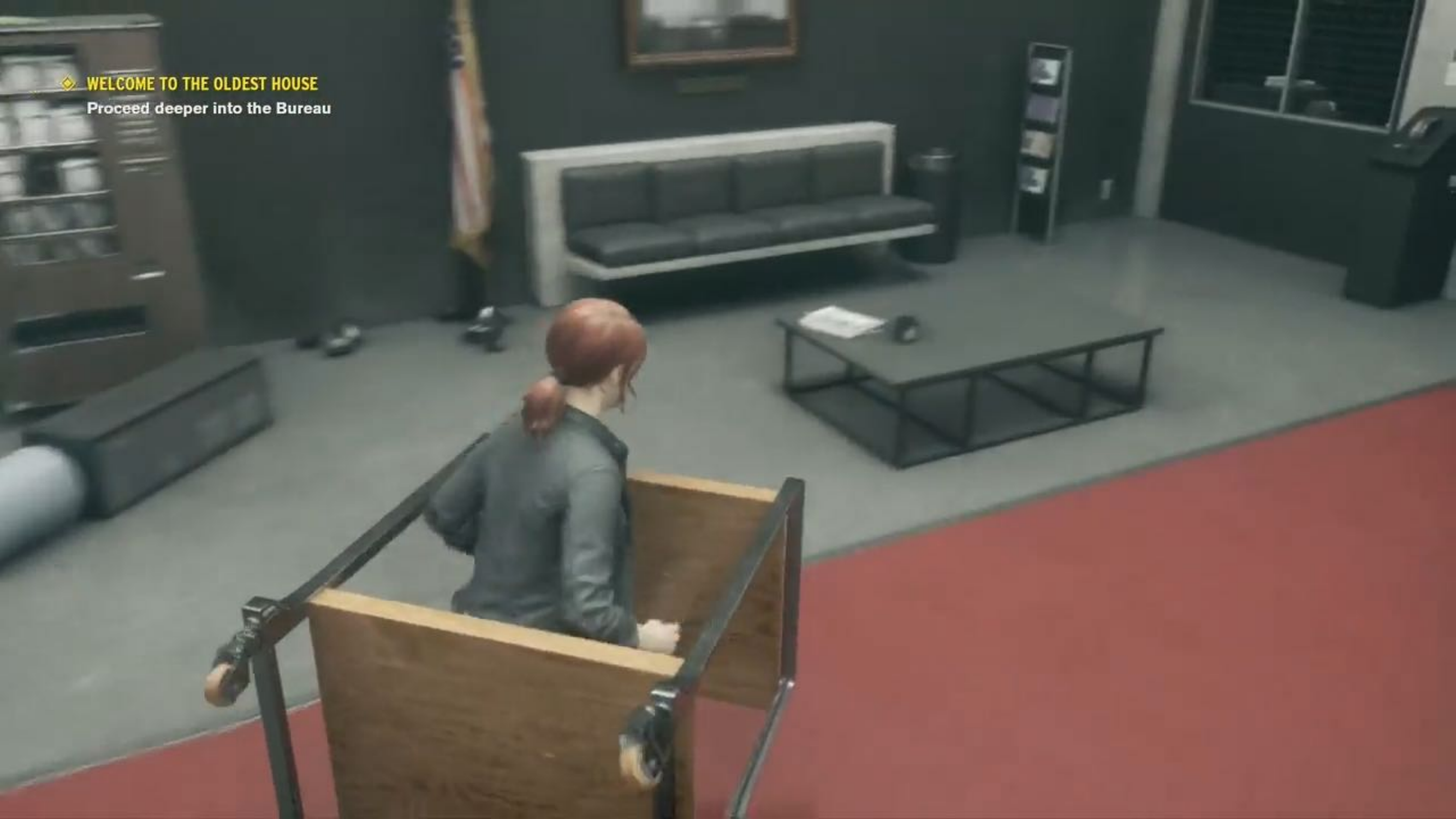}
  \caption{A person stuck in a piece of furniture}
\end{subfigure}
\hfill 
\begin{subfigure}{0.325\textwidth}
  \includegraphics[width=\linewidth]{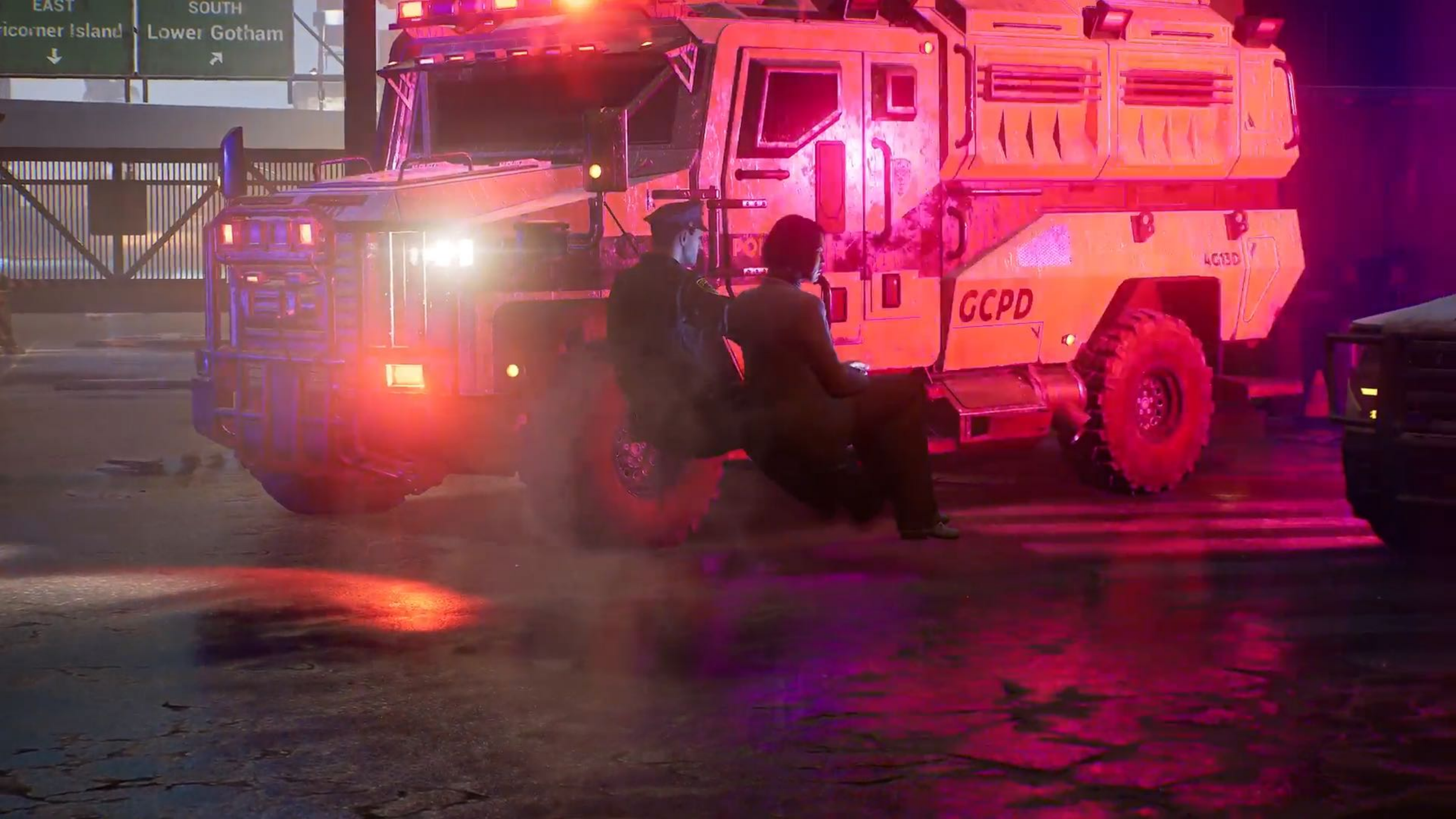}
  \caption{Two people driving an invisible car}
  \label{fig:invisible_car}
\end{subfigure}
\hfill
\begin{subfigure}{0.325\textwidth}
  \includegraphics[width=\linewidth]{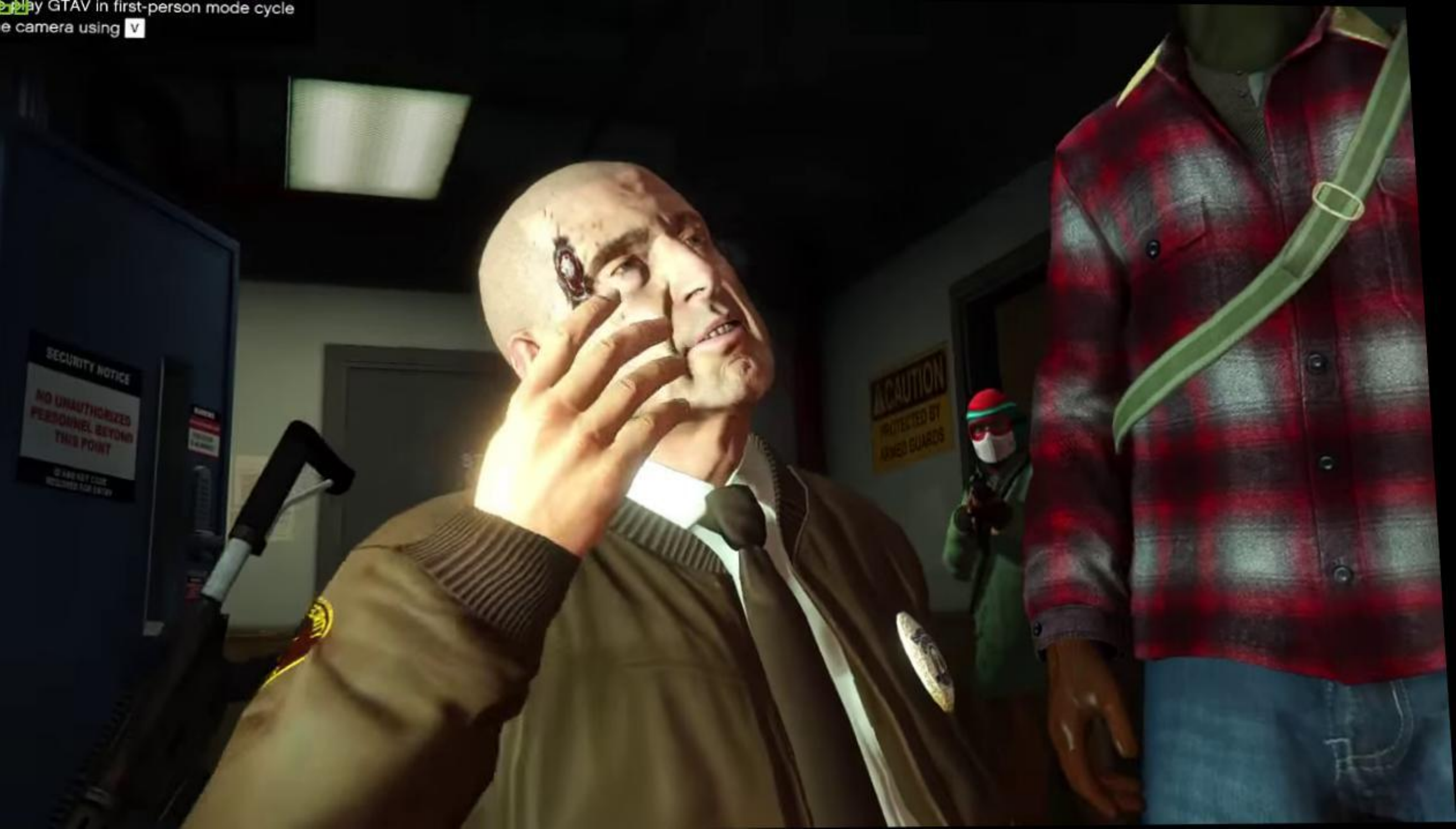}
  \caption{A rifle floating in the air}
\end{subfigure}

\vspace{\baselineskip} 

\begin{subfigure}{0.325\textwidth}
  \includegraphics[width=\linewidth]{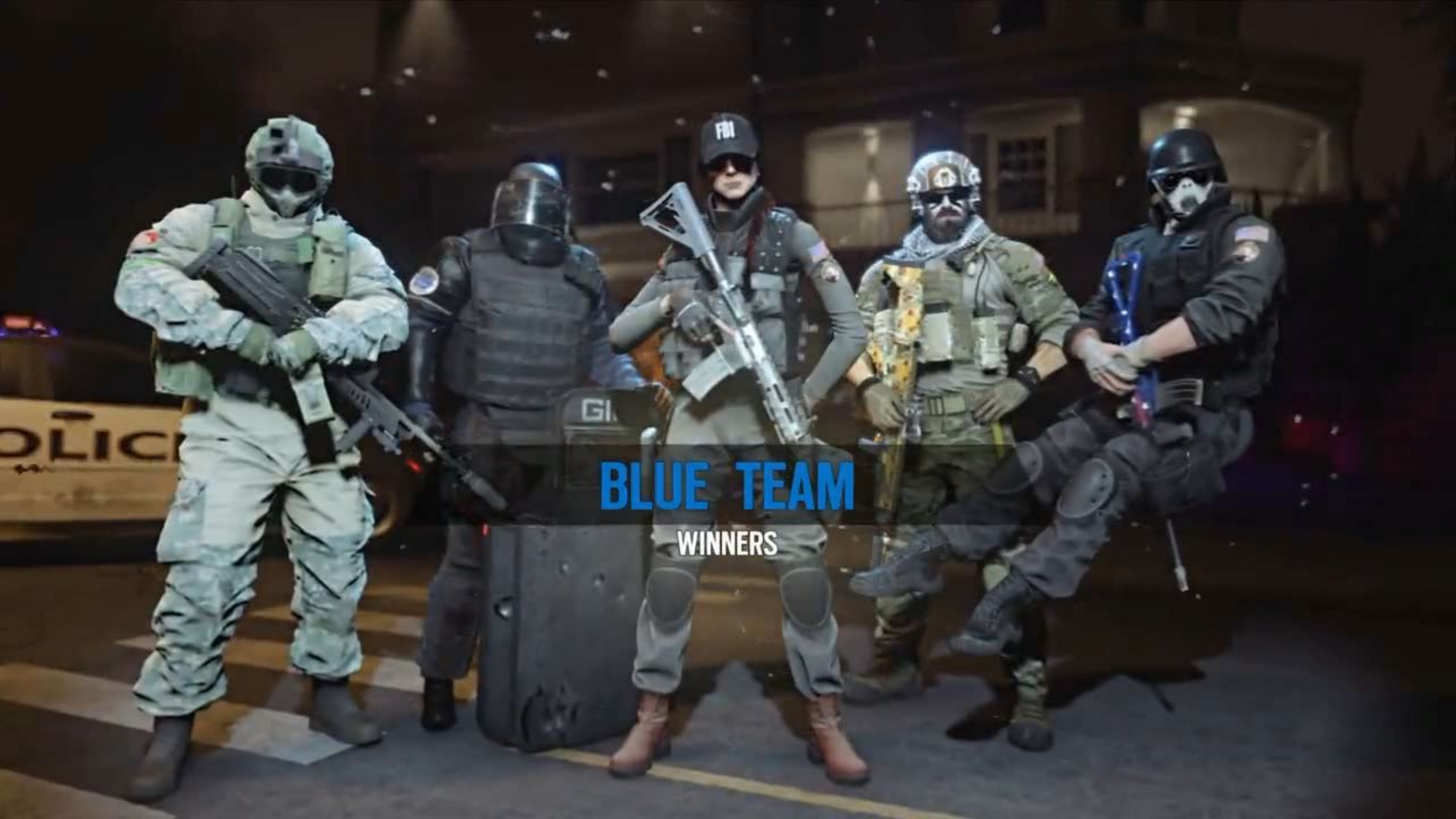}
  \caption{A person is floating in the air}
  \label{fig:person_float}
\end{subfigure}
\hfill
\begin{subfigure}{0.325\textwidth}
  \includegraphics[width=\linewidth]{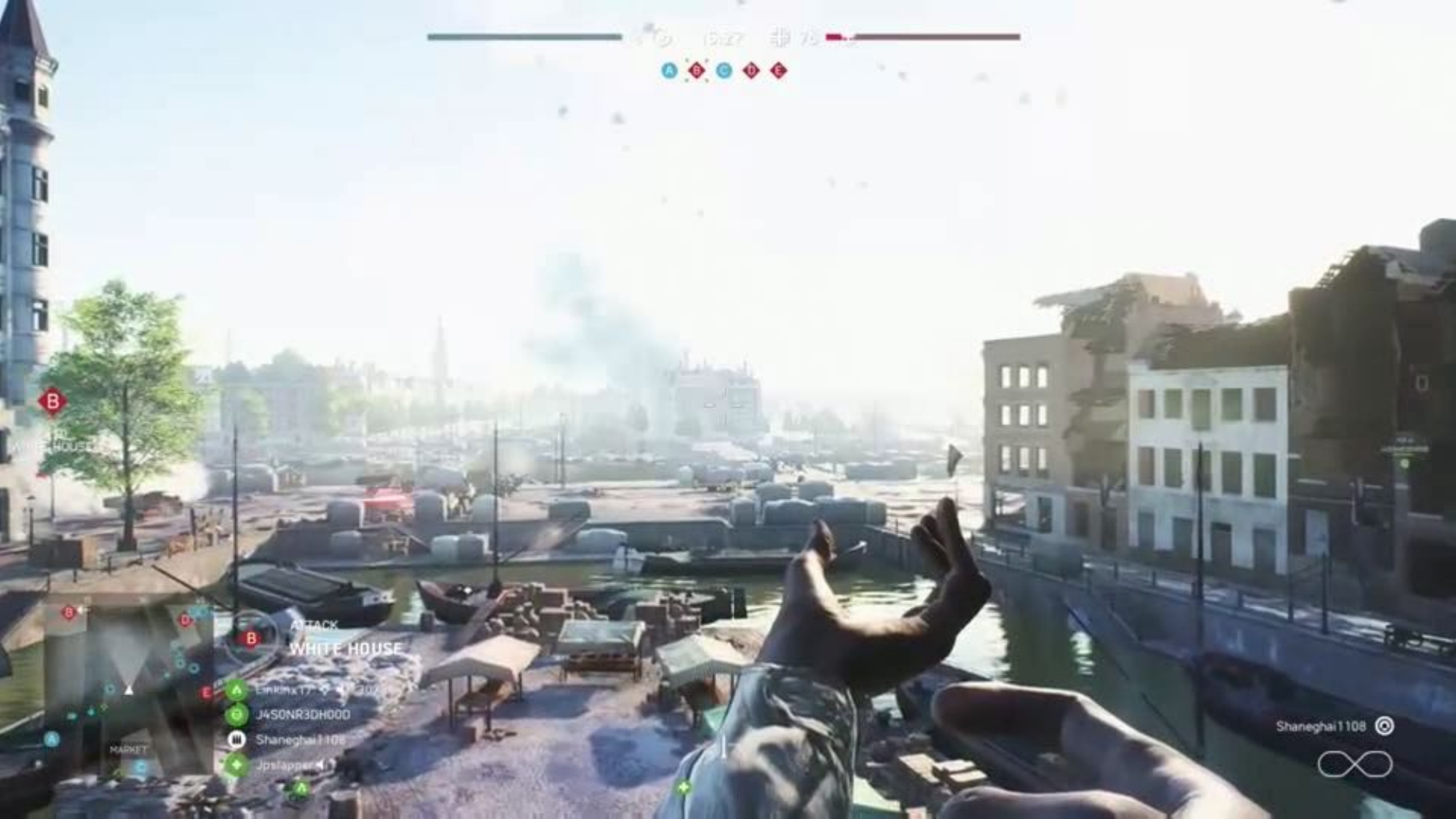}
  \caption{The gun in the hand is missing}
   \label{fig:missing_gun}
\end{subfigure}
\hfill
\begin{subfigure}{0.325\textwidth}
  \includegraphics[width=\linewidth]{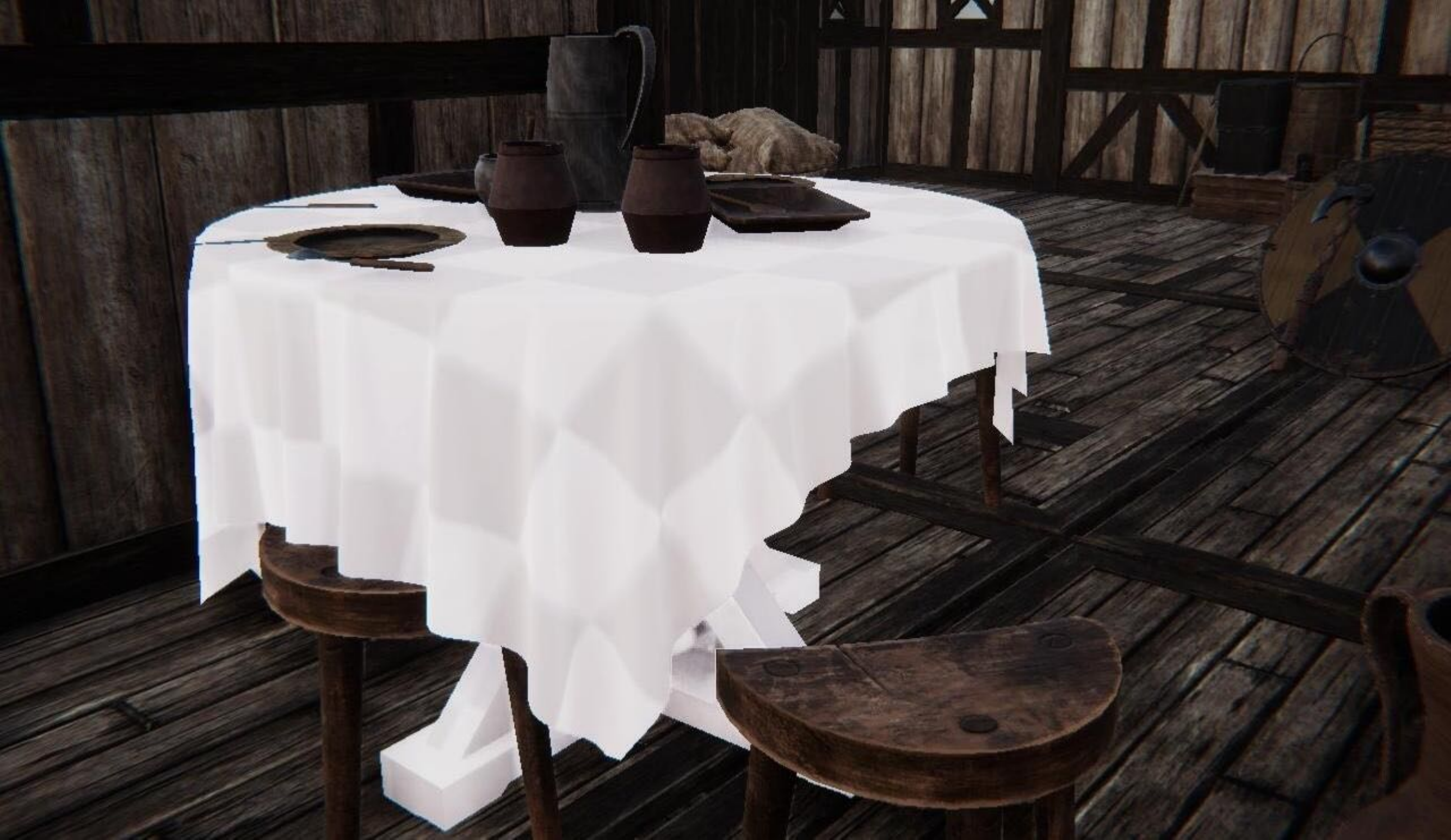}
  \caption{The table cover has a placeholder texture}
  \label{fig:missing_texture}
\end{subfigure}

\caption{Sample images from the \datasetname  showing glitches in various games with distinct styles. Samples (a)--(e) are captured from online videos, while sample (f) is generated inside the Unity game engine.}
\label{fig:sample_glitches}
\end{figure*}

The video game industry boasts an estimated annual revenue of USD 217 billion~\cite{GrandViewResearch2023} with a total of 3.2 billion gamers worldwide in 2022 \cite{gamers_epic}.
Automatically detecting in-game glitches is, therefore, a highly demanding task, but that remains a long-standing challenge~\cite{rahman2023weak, taesiri2022clip,taesiri2020video, ling2020using, zheng2019wuji, rani2023deep, chen2021glib, wilkins2022learning, nantes2008framework}.
A \textit{glitch} is an unexpected frame that occurs within a game due to either an unforeseen software bug, player actions, or unanticipated interactions between game elements and does \emph{not} result in a program crash.
From a computer vision perspective, glitch detection involves recognizing an extremely wide spectrum of long-tail video frames, from rendering (\eg, 3D objects with missing textures; \cref{fig:missing_texture}), unrealistic physics (\eg, two people sitting in an invisible car; \cref{fig:invisible_car}), to semantic errors (\eg, indoor rain; \cref{fig:teaser}).
Unlike software bugs that can be caught by examining the code alone, glitches are more non-trivial to detect because they are mostly the result of a one-time event that unexpectedly happens in-game.


A holy grail of game quality assurance is to build a general glitch detector that works for any game of any genre and mechanics.
We set the first step toward this goal by building \datasetname, an evaluation benchmark of 593 glitches, leveraging the public's crowd knowledge from the game community's reports on \url{reddit.com/r/GamePhysics}.
The glitches span across 205 games of various genres.
Each glitch has a video clip, a representative frame, a one-line description, and a reference to a corresponding Reddit thread where gamers discussed the error.

Large image-text, multimodal models (LMMs), such as GPT-4V~\cite{gpt4v}, are improving at an unprecedentedly fast pace.
They excel in many existing tasks, including object detection~\cite{yang2023set, liu2023llava}, multi-step reasoning~\cite{bubeck2023sparks, li2023large, alayrac2022flamingo, awadalla2023openflamingo}, and detailed image captioning~\cite{gpt4v, lin2023sphinx, liu2023improved, yang2023re, peng2023kosmos}.
Testing LMMs on \datasetname may yield important findings not only to the game industry but also to the Artificial Intelligence (AI) community because glitch detection requires a combination of knowledge and understanding of image aesthetics, computer graphics, physics and commonsense reasoning (skills that are often tested individually in a benchmark~\cite{bitton2023visit}).

In this paper, we evaluate how well LMMs perform in detecting glitches from a single frame.
Our main findings and contributions include:

\begin{enumerate}
    \item We introduce \datasetname, which contains 330 glitch-free and 593 glitch screens taken from 205 games for evaluating LMMs (\cref{sec:method}).
    
    \item We evaluate 11 state-of-the-art LMMs, including GPT-4V~\cite{gpt4v} and LLaVA~\cite{liu2023improved} on our benchmark and in comparison with the performance on 6 other common benchmarks (\cref{sec:benchmarking}).

    \item LMMs are better at detecting glitches that violate simple physical laws (\eg, a car flying in the air) than other more subtle glitches (\eg, human limbs in an implausible pose; \cref{fig:gpt4fails_basketball}).

    \item The state-of-the-art model on \datasetname is GPT-4V with 43.4\% accuracy. In the extensive captioning setup, we estimated the upper limits of models, and GPT-4V can achieve an accuracy of 64.9\%, which is almost twice that of LLaVA, the second-best model (30.5\%).

    \item In sum, there exists a headroom of 30--35\% on \datasetname for future LMM models to improve, presenting an interesting challenge to the AI community.
    
\end{enumerate}

\section{Related Work}
\label{sec:relwork}

\subsection{Multimodal, image-text datasets}

Recently, there has been rapid development of large multimodal models that can process multiple modalities, including visual and textual inputs. 
Existing datasets that come with human-generated image captions, such as COCO Caption~\cite{chen2015microsoft}, Nocaps~\cite{agrawal2019nocaps}, CapFilt:~\cite{li2022blip} and Flickr30k~\cite{plummer2015flickr30k}, can serve as a simple way to evaluate language models. By providing the image, we can ask a model to describe it and then compare the generated caption with the ground truth~\cite{liu2023improved, ye2023mplug, liu2023visual}.
Image captioning is a narrow domain and can be extended into visual question answering (VQA) by asking questions related to an image.
Datasets like GQA~\cite{hudson2019gqa}, OK-VQA~\cite{marino2019ok}, VQAv2~\cite{goyal2017making}, and Vizwiz~\cite{gurari2018vizwiz} contain image-question pairs to probe the visual reasoning and understanding of LMMs.

Building upon simple VQAs, several benchmarks aim to increase the complexity of tasks over different dimensions. TextVQA~\cite{singh2019towards}, OCR-VQA \cite{mishra2019ocr} and TextCap\cite{sidorov2020textcaps} propose questions about the text shown in the image.
ScienceQA~\cite{lu2022learn} and MathVista~\cite{lu2023mathvista} focus on scientific topics and charts, while VCR~\cite{zellers2019recognition} and Sherlock~\cite{zellers2019recognition} focus on commonsense reasoning.
Moreover, AI2D~\cite{hiippala2021ai2d}  is directed at questions concerning scientific diagrams, and IconQA~\cite{lu2021iconqa} targets the comprehension of abstract diagrams. Each of these benchmarks is designed to push the boundaries of VQA systems by introducing specialized content that requires advanced reasoning and understanding.

There are also comprehensive evaluation frameworks that assess multimodal language models across a wider spectrum of capabilities. These evaluations extend beyond visual and textual reasoning to encompass a variety of skills such as generation, question answering, adherence to instructions, and the application of commonsense logic. Notable among these are SEED-Bench~\cite{li2023seed} , MME~\cite{fu2023mme}, MMBench~\cite{liu2023mmbench}, MM-Vet~\cite{yu2023mm}, VisIT-Bench~\cite{bitton2023visit}, which collectively serve to provide a robust measure of a model's proficiency in handling tasks that integrate multiple modalities.

Unlike traditional datasets that contain queries about elements present in the image, our approach is novel in directing models to discern the atypical aspects, \ie, glitches, with no linguistic hints provided.
We show an image to the model and ask it to report unusual aspects of it. Such questions require a more integrated approach to visual and linguistic processing within an LMM to formulate a response.

\subsection{Vision-language Stress Testing}

Out-of-distribution (OOD) datasets have become a cornerstone for evaluating the capabilities and progress of machine learning models. In standard image classification, in particular the ImageNet~\cite{russakovsky2015imagenet} dataset, the introduction of datasets~\cite{hendrycks2021many, hendrycks2021many, hendrycks2019benchmarking, taesiriimagenet} has underscored the importance of robustness and generalization in model evaluation. 
As we move from simple image classification tasks to more complex multimodal tasks, there is an increasing need for similar OOD datasets that can comprehensively test the generalization abilities of LMMs.

There are several studies that stress test various aspects of vision from different angles, such as compositional and spatial reasoning~\cite{kamath2023s, gokhale2022benchmarking, thrush2022winoground, jiang2022bongard}, objects placed out of context and implausible scenes~\cite{CHOI2012853, bitton2023breaking, zhou2023rome}, and the exploitation of language and vision priors~\cite{liu2023hallusionbench, frank2021vision}.

The closest benchmark to ours is Whoops~\cite{bitton2023breaking}, which is designed to challenge commonsense knowledge and reasoning in LMMs. 
However, our dataset differs in several ways:  (1) The tasks in \datasetname come from real-world tasks, specifically video game quality assurance, and are not artificially created to test models. 
(2) Whoops requires cultural and background knowledge to answer; for example, \textit{A panda bear is catching salmon fish} is unusual since pandas subsist almost entirely on bamboo. In contrast, our dataset contains samples that contradict basic commonsense and the physics of the world. 
(3) Finally, images in Whoops are synthesized using image-to-text models; they are clear without artifacts, centered in the image, and do not stress the visual side of the image, focusing on the context.
In contrast, for \datasetname, models need to fully scan the image to identify its unusual aspects (\cref{fig:sample_glitches}), and there are many distracting elements present in the image, challenging them to focus on the correct part of the image.

\subsection{Empirical Analysis of Recent LMMs}

With the release of recent proprietary LLMs, such as GPT-4V and Bard~\cite{google2023bard}, some studies attempt to evaluate and report the performance of these models on various benchmarks and tasks~\cite{wu2023early, cui2023holistic, Qin_2023}. The main goal of these studies is to provide a comprehensive evaluation of the models across various well-established tasks and some narrow domains~\cite{wen2023road, yan2023multimodal}. The main difference between our work and these studies is that we propose a general, stress-testing benchmark to measure the generalization power of various LLMs, both proprietary and open source, on a specific, glitch-detection task in the game industry.

\section{\datasetname}
\label{sec:method}

In this section, we describe the creation process of \datasetname, a benchmark aimed at stress-testing visual perception and commonsense reasoning in LMMs, motivated by real-world game quality assurance tasks.

During development, video games go through many stages of testing to reach certain quality standards before release.
However, even after release, they can still exhibit unusual in-game events, or glitches.
Glitches, often viewed as annoying bugs, can also possess a humorous and entertaining aspect. Players frequently report glitches across various social media platforms, particularly on Reddit and YouTube.
A critical aspect of understanding glitches is the requirement of commonsense knowledge about the basic laws of physics of the game's universe, making them a suitable and practical candidate for testing machine learning models. \cref{fig:sample_glitches} shows six samples from \datasetname.

\subsection{Constructing the Dataset}

\datasetname contains two parts: (1)~513 samples shared by players of video games, \ie, frames collected from online sources, and (2)~75 synthetic samples.

\paragraph{Samples shared by players of video games:}
To construct our dataset, we sampled 1,000 videos from the GamePhysics~\cite{taesiri2022clip} dataset. This dataset consists of videos from a \href{https://www.reddit.com/r/GamePhysics/}{subreddit} with the same name, containing gameplay video clips with unusual events and glitches.

Next, we conducted a manual review process to filter videos based on two criteria: (1)~the presence of a glitch in the video, and (2) the potential for humans to detect the glitch from a single frame. The second criterion is key because certain glitches, such as those involving rapid shaking or changes in size over time, cannot be detected from a still image alone.

After applying these filters, we extracted one frame from each remaining video, resulting in a collection of 650 samples.
Our final round of manual reviews revealed two potential issues: (1)~some glitches are not detectable from the extracted image and require more context to understand, and (2)~some images contain the faces of gamers who streamed the content on an online platform (which could cause the LMM to identify these faces as what is wrong with the images).
After removing videos that contain one of these issues, our final glitch set contains 513 images.

\paragraph{Generating synthetic samples with Unity:}
To enhance our dataset, we supplemented samples from the GamePhysics dataset with 75 synthetic examples created inside the Unity game engine.
These samples were specifically designed to mimic a subset of common development-stage bugs~\cite{rahman2023weak, taesiri2020video, ling2020using} that are not readily available in online social media platforms and, hence, to diminish the survivor bias effect. These flaws are often fixed before the public release of a game through the quality assurance process of a game development company and are therefore not often posted on social media.

Our synthetic sample generation process involves the injection of three categories of glitches into each scene: (1)~placeholder textures, (2)~object mesh distortions, and (3)~low-resolution textures.

\paragraph{Glitch-free images:} Our focus is on glitch frames, as they are more challenging to capture and collect. However, to establish a baseline for comparison, we also included a set of glitch-free images. To accomplish this, we randomly selected gameplay walkthroughs from various games on YouTube. From these walkthroughs, we extracted a random subset of frames, resulting in the compilation of a dataset consisting of 330 frames sourced from a diverse array of games. The groundtruth captions for these glitch-free images is \textit{``There is nothing wrong with this image''}.


\subsection{Labeling the Dataset}

For all images, we provide a short description of the glitch present in the image. Our goal is to label the images briefly, highlighting only the unusual elements in simple language. For instance, if an image depicts a character with a contorted physique, the label would simply state, \textit{``character has an unnatural body position''}.

It is important to highlight that some images can be described in many different ways. Diverse phrases such as \textit{``falling from the sky'',} \textit{``suspended in mid-air''}, or \textit{``jumping in the air''} might all refer to a single event.
Instead of handling such cases in the labeling process, in the evaluation process, we incorporate a language model to diminish the effect of this (see~\cref{sec:exp_setup}).

\subsection{Categorizing the Glitch Types in the Images}

In this section, we provide a high-level categorization of glitches in our dataset. While there have been some attempts to provide a taxonomy of video game bugs~\cite{lewis2010went, truelove2021we}, these taxonomies do not provide descriptions that are adequate to automate bug categorization.

We propose a novel human-AI team-based method to build a categorization based on the descriptions of the images. This process is a collaborative effort between GPT-4 and humans, where GPT-4 suggests initial categories, and then humans refine these suggestions by providing feedback or asking the model to re-evaluate its output, harnessing the reflective ability of GPT-4~\cite{shinn2023reflexion}. Finally, we manually bridge the resulting categories to those proposed by Lewis et al.~\cite{lewis2010went} based on the semantics and instances of the glitches in our dataset.

\paragraph{Process:} We prompt GPT-4 with all the glitch descriptions in our dataset and ask it to generate a categorization based on the descriptions and semantics of the glitches. In each subsequent iteration, we provide feedback in one of two ways: (1)~we ask GPT-4 to review its previous answer through reflection, or (2)~we explicitly instruct the model to merge two categories that are semantically similar.
We stop when the model no longer changes its answer through reflection or when we can no longer merge  categories.

In the last step, to assign each image to a category, we prompt GPT-4 with the description of the glitch and the final categories and ask it to assign each image to one of them. The final categories, the number of instances, examples for each category, and the parent category proposed by Lewis et al.~\cite{lewis2010went} are outlined in Table~\ref{tab:bugs_cats}.

\begin{table}[t]
    \centering
    \small
\caption{Categorization of video game glitches in \datasetname. 
Numbers highlighted in \colorrect{lightpurple} show the number of images in each category. Categories highlighted in \colorrect{lightblue} show the corresponding categories proposed by Lewis et al.~\cite{lewis2010went}.}
    \label{tab:bugs_cats}
    \begin{tabular}{p{0.9\linewidth}}
        \toprule
        \textbf{Physics, Collision, and Spawn}\hfill \colorbox{lightpurple}{Images: 422} \\
        \multicolumn{1}{c}{ \colorbox{lightblue}{(Non-Temporal $\rightarrow$ Invalid position)}  } \\ 
        \midrule
         1. Objects and characters floating or stuck in the air (\cref{fig:person_float}). \\
         2. Characters or objects clipping through solid objects like walls, floors, or ground. \\
         3. Vehicles or characters falling under the game map. \\
        \midrule
        \textbf{Animation and Pose} \hfill \colorbox{lightpurple}{Images: 75} \\
        \multicolumn{1}{c}{ \colorbox{lightblue}{ (Non-Temporal $\rightarrow$ Invalid graphical representation) }  } \\ 
        \midrule
        1. Unusual or impossible body poses and positions (\cref{fig:gpt4fails_basketball}).\\
        2. Characters in a T-pose or with distorted body parts.  \\
        3. Incorrect animations for certain actions. \\
        \midrule
        \textbf{Rendering and Texture}\hfill \colorbox{lightpurple}{Images: 67} \\
        \multicolumn{1}{c}{ \colorbox{lightblue}{ (Non-Temporal $\rightarrow$ Invalid graphical representation)  }  } \\ 
        \midrule
         1. Mesh stretches or objects with distorted shapes. \\
         2. Missing textures or objects displaying a ``default'' placeholder texture (\cref{fig:missing_texture}). \\
         3. Objects with low-resolution. \\
        \midrule
        \textbf{Camera, User Interface, and Lighting}\hfill \colorbox{lightpurple}{Images: 26} \\
        \multicolumn{1}{c}{ \colorbox{lightblue}{ (Non-Temporal $\rightarrow$ Invalid value change) } } \\ 
        \midrule
         1. Camera issues such as clipping inside objects or improper character views. \\
         2. In-game menus displaying incorrect elements. \\
         3. Shadows or lighting effects that do not match the environment. \\
        \bottomrule
    \end{tabular}
\end{table}

\section{Experiments}
\label{sec:benchmarking}

\begin{table*}[t]
\centering
\caption{Accuracy of various LMMs on \datasetname. Numbers highlighted in \colorrect{apricot} represent the average results of Q1 and Q2, which are the main results of the benchmark. Numbers related to Q3 serve as a visual perception test to measure the ability of models to report glitches in a relaxed manner. Numbers highlighted in \colorrect{lightblue} show the maximum agreement achievable with ground truth as perceived by Llama-2's judgment (\%).
\CameraReady{Numbers highlighted in \colorrect{LimeGreen} represent the results obtained from GPT-4V on glitch-free images.}
}
\label{tab:main_results} 
\resizebox{\textwidth}{!}{%
\begin{tabular}{@{}lr|rrrrrrrrrrr@{}}
\toprule
\multicolumn{1}{c}{\multirow{2}{*}{\textbf{Question}}} &
  \multicolumn{2}{c}{\textbf{\begin{tabular}[c]{@{}c@{}}GPT-4V\\ \cite{gpt4v} \end{tabular}}} &
  \multicolumn{2}{c}{\textbf{\begin{tabular}[c]{@{}c@{}}LLaVA-1.5\\ \cite{liu2023improved} \end{tabular}}} &
  \multicolumn{2}{c}{\textbf{\begin{tabular}[c]{@{}c@{}}SPHINX\\  \cite{lin2023sphinx} \end{tabular}}} &
  \multicolumn{2}{c}{\textbf{\begin{tabular}[c]{@{}c@{}}InstructBLIP\\ \cite{dai2023instructblip} \end{tabular}}} &
  \multicolumn{1}{c}{\textbf{\begin{tabular}[c]{@{}c@{}}OtterHD\\ \cite{li2023otterhd} \end{tabular}}} &
  \multicolumn{1}{c}{\textbf{\begin{tabular}[c]{@{}c@{}}Qwen\\-VL \cite{bai2023qwen} \end{tabular}}} &
  \multicolumn{1}{c}{\textbf{\begin{tabular}[c]{@{}c@{}}MiniGPT\\-v2 \cite{chen2023minigpt} \end{tabular}}} &
  \multicolumn{1}{c}{\textbf{\begin{tabular}[c]{@{}c@{}}Fuyu\\ \cite{fuyu-8b} \end{tabular}}} \\ \cmidrule(l){2-13}
\multicolumn{1}{c}{} &
  \multicolumn{1}{c}{n/a} &
  \multicolumn{1}{c}{n/a} &
  \multicolumn{1}{c}{7B} &
  \multicolumn{1}{c}{13B} &
  \multicolumn{1}{c}{7B} &
  \multicolumn{1}{c}{13B} &
  \multicolumn{1}{c}{7B} &
  \multicolumn{1}{c}{13B} &
  \multicolumn{1}{c}{8B} &
  \multicolumn{1}{c}{10B} &
  \multicolumn{1}{c}{7B} &
  \multicolumn{1}{c}{8B} \\ \midrule
\textbf{Q1.} What is unusual about this image? &
\cellcolor{LimeGreen}88.2 & 57.2 & 35.2 & 36.3 & 19.2 & 25.3 & 25.3 & 21.9 & 24.8 & 21.2 & 19.1 & 8.6\\
\textbf{Q2.} What is wrong with this image? &
\cellcolor{LimeGreen}95.5 & 29.5 & 23.9 & 34.7 & 30.9 & 30.5 & 13.8 & 8.9 & 23.3 & 9.3 & 17.9 & 8.4 \\ \cmidrule(l){2-13}
\rowcolor{apricot} \multicolumn{1}{r}{\textbf{Average}} &
\cellcolor{LimeGreen}91.6  & 43.4 & 29.6 & 35.5 & 25.0 & 27.9 & 19.6 & 15.4 & 24.0 & 15.2 & 18.5 & 8.5 \\ \midrule
\textbf{Q3.} Describe the image in detail. &
- & 64.9 & 28.0 & 30.5 & 17.5 & 21.9 & 16.0 & 11.8 & 21.6 & 14.0 & 16.0 & 7.6 \\ \midrule
\rowcolor{lightblue} \multicolumn{1}{r}{\textbf{Maximum Agreement}} &
\cellcolor{LimeGreen}95.5  & 64.9 & 35.2 & 36.3 & 30.9 & 30.5 & 25.3 & 21.9 & 24.8 & 21.2 & 19.1 & 8.6\\ \bottomrule
\end{tabular}%
}
\end{table*}

\subsection{Experimental Setup}
\label{sec:exp_setup}

\paragraph{Formulating Questions:}
We designed \datasetname as a free-text response benchmark, in contrast with traditional LMM benchmarks that utilize Yes/No or multiple-choice formats~\cite{fu2023mme, li2023seed}.
We ask models to describe the unusual aspects of an image by answering three questions:
\begin{itemize}
    \item[] \textbf{(Q1)} \textit{What is unusual about this image?}
    \item[] \textbf{(Q2)}  \textit{What is wrong with this image?}
    \item[] \textbf{(Q3)} \textit{Describe the image in detail}
\end{itemize} 
Note that we do not explicitly use the word \textit{glitch} in the question, and we use simple language similar to what a layperson would use.
During the inference, we allow models to come up with their own reasoning, and after the model generates the full response, we record it for further evaluation and comparison with the ground truth.


The rationale for free-text answers is that including an `unusual' event description among choices hints to the LMM, letting it answer while disregarding visual aspects.

We included question \textbf{Q3} to assess whether the models can accurately report any glitches or unusual elements within the image in extensive captioning. Essentially, this question serves as a visual perception test, evaluating whether the models can identify and describe unusual aspects of the image in a more relaxed condition.
For example, in the sample shown in~\cref{fig:teaser}, we test the model to see if it can identify the presence of rain in the room. In this case, it indicates that it is raining outside.

\paragraph{Evaluation:}

\begin{figure}[t]
    \centering
    \includegraphics[width=1\columnwidth]{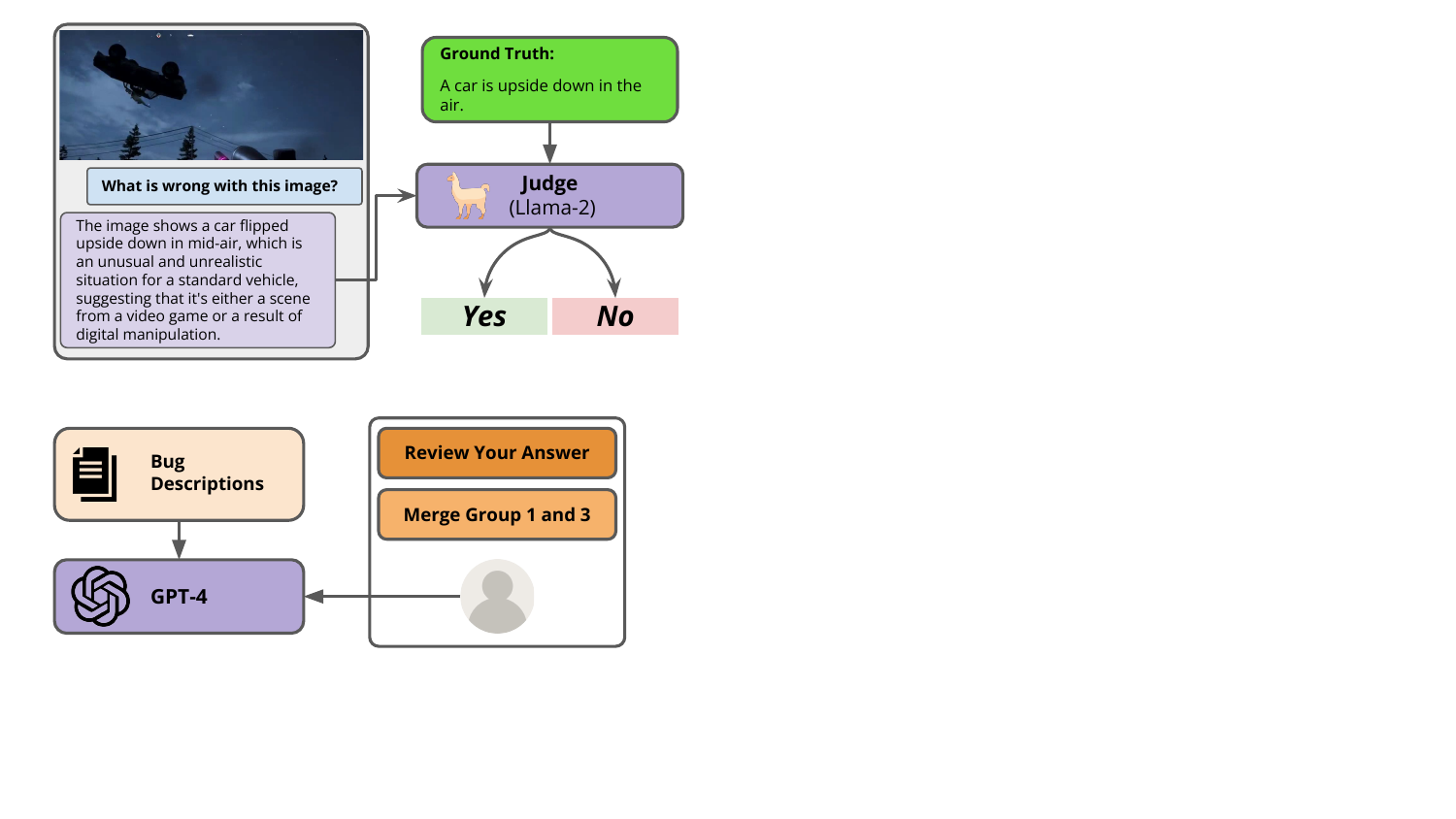}
    \caption{To evaluate a model's response, we ask a judge (the Llama-2-70b-Chat model) to compare it semantically with the ground truth.}
    \label{fig:Judging_process}
\end{figure}

Following recent successes~\cite{yin2023lamm, zheng2023judging, liu2023aligning, bitton2023visit} we employ a language model as a judge to evaluate the model's responses. We use Llama-2-70B-Chat~\cite{touvron2023llama} to compare the model-generated text with the ground truth and determine whether the text conveys the same meaning or mentions the event highlighted by the ground truth (see Fig.~\ref{fig:Judging_process}).

We report the accuracy of each model on each tested question and present the average performance for \textbf{Q1} and \textbf{Q2} as the final benchmark result. \textbf{Q3} serves as the visual perception test, and we report the performance of the models on it separately.

To assess Llama-2's judgment and determine if it can effectively serve as an evaluator, we manually reviewed a subset of responses for each model. For each model, we manually labeled 20 samples, with a total of 220 samples.

\paragraph{Models:}

In total, we evaluated 11 LMMs, including GPT-4V~\cite{gpt4v}, and 10 open source models: LLaVA-1.5 (7B and 13B)~\cite{liu2023improved}, SPHINX (7B and 13B)~\cite{lin2023sphinx}, InstructBLIP (7B and 13B)~\cite{dai2023instructblip}, Qwen-VL-Chat (10B)~\cite{bai2023qwen}, MiniGPT-v2 (7B)~\cite{chen2023minigpt}, OtterHD~\cite{li2023otterhd}, and Fuyo (8B)~\cite{fuyu-8b}.
%
%
We used the default temperature and top-p configurations provided with the model and API. We increased \texttt{max\_token} to get full responses from models. (See~\cref{sec:supp-immplementation_details} for details).

\subsection{Quantitative Results}

Table~\ref{tab:main_results} shows the performance of all the tested models for the three questions. The \colorbox{apricot}{Average} performance on \textbf{Q1} and \textbf{Q2} is the main result of our benchmark.
GPT-4V is the best-performing model, achieving 57.2\% (\textbf{Q1}) and 29.5\% (\textbf{Q2}) and an average of 43.4\%.
Next, LLaVA-1.5-13B achieves an average of 35.5\% and is the best performing open-source model.
These findings show \datasetname is challenging for even state-of-the-art commercial \& open-source models.

\CameraReady{The performance of GPT-4V on glitch-free images is much higher than on glitch images, with an average accuracy of 91.6\%, which suggests that glitch-free images are much easier to handle.}

Models exhibit different performance depending on the questions being asked, but all except for the SPHINX family show better performance when prompted with \textbf{Q1}. Nevertheless, the gap in performance varies, with GPT-4V showing the largest gap of 27.7pp (57.2\% vs. 29.5\%). These results highlight that different prompts steer the behavior of LMMs differently and suggest that multi-step reasoning~\cite{wei2022chain, kojima2022large} could also help LMMs.

Our results also highlight that higher resolutions improve the performance. In particular, SPHINX-13B, which operates at a higher resolution than SPHINX-7B ($448 \times 448$ vs. $224 \times 224$), on average performs \increasenoparent{2.9} pp (27.9\% vs. 25.0\%) better than the base model. 
Similarly, OtterHD, which employs Fuyu as the base model with enhanced flexibility and support for higher image resolutions, outperforms Fuyu on average by \increasenoparent{15.5} (24.0\% vs. 8.5\%).


Asking LMMs to extensively caption the image using \textbf{Q3} only triggers GPT-4V to produce a very verbose response. In many cases, GPT-4V describes many details in the image and can touch upon the unusual aspects of the image.
In this setup, GPT-4V can achieve 64.9\%, which is an increase of \increasenoparent{7.7} over \textbf{Q1} and \increasenoparent{21.5} pp better than the benchmark results. This gap suggests that GPT-4V can \textit{see} many details in the image, but it cannot easily focus on the unusual aspects in the frame, indicating a gap in its reasoning capabilities across different modalities and prompts.

\paragraph{Human evaluation:} Table~\ref{tab:judge_llm_vs_human_kappa} shows the results of comparing between Llama-2 judgments and human evaluations, with the level of agreement for each model measured by Cohen's Kappa~\cite{cohen1960coefficient}. Cohen's Kappa demonstrates varying levels of concordance for each model. GPT-4V (0.80), InstructBLIP-7B (0.83), and Qwen-VL (1.00) exhibit substantial to perfect agreement.
In contrast, OtterHD (0.50) had fair agreement, and Fuyu (-0.09) shows less than chance agreement, suggesting significant discrepancies.
Overall, on all models except for Fuyu, we found above moderate agreement between Llama-2 and human judgment, while on six models, this agreement is substantial.

\begin{table}[t]
\centering
\caption{Evaluating a subset of responses for comparing Llama-2 with human judgments: Llama-2 and humans exhibit moderate to substantial agreement on all models except for Fuyu.}
\label{tab:judge_llm_vs_human_kappa}
\begin{tabular}{@{}lrrr@{}}
\toprule
\multicolumn{1}{c}{\textbf{Model}} & \textbf{Llama-2} & \textbf{Human} & \textbf{$\kappa$} \\ \midrule
GPT-4V                             & 60.0               & 50.0                & 0.80                  \\
LLaVA-1.5-13B                      & 25.0               & 20.0                & 0.57                  \\
LLaVA-1.5-7B                       & 35.0               & 15.0                & 0.49                  \\
Long-SPHINX                        & 25.0               & 35.0                & 0.53                  \\
SPHINX                             & 30.0               & 25.0                & 0.63                  \\
InstructBLIP-13B                   & 20.0               & 10.0                & 0.62                  \\
InstructBLIP-7B                    & 20.0               & 15.0                & 0.83                  \\
MiniGPT-v2                         & 10.0               & 5.0                 & 0.64                  \\
Qwen-VL                            & 20.0               & 20.0                & 1.00                  \\
OtterHD                            & 25.0               & 10.0                & 0.50                  \\
Fuyu                               & 20.0               & 5.0                 & -0.09                 \\ \midrule
\multicolumn{1}{c}{\textbf{$\mu \pm \sigma $}} & \multicolumn{2}{c}{26.4 $\pm$ 12.8,\xspace\xspace 19.1 $\pm$ 13.5} & 0.64 \\ \bottomrule
\end{tabular}
\end{table}

\begin{figure}[t]
    \centering
    \includegraphics[width=1\columnwidth]{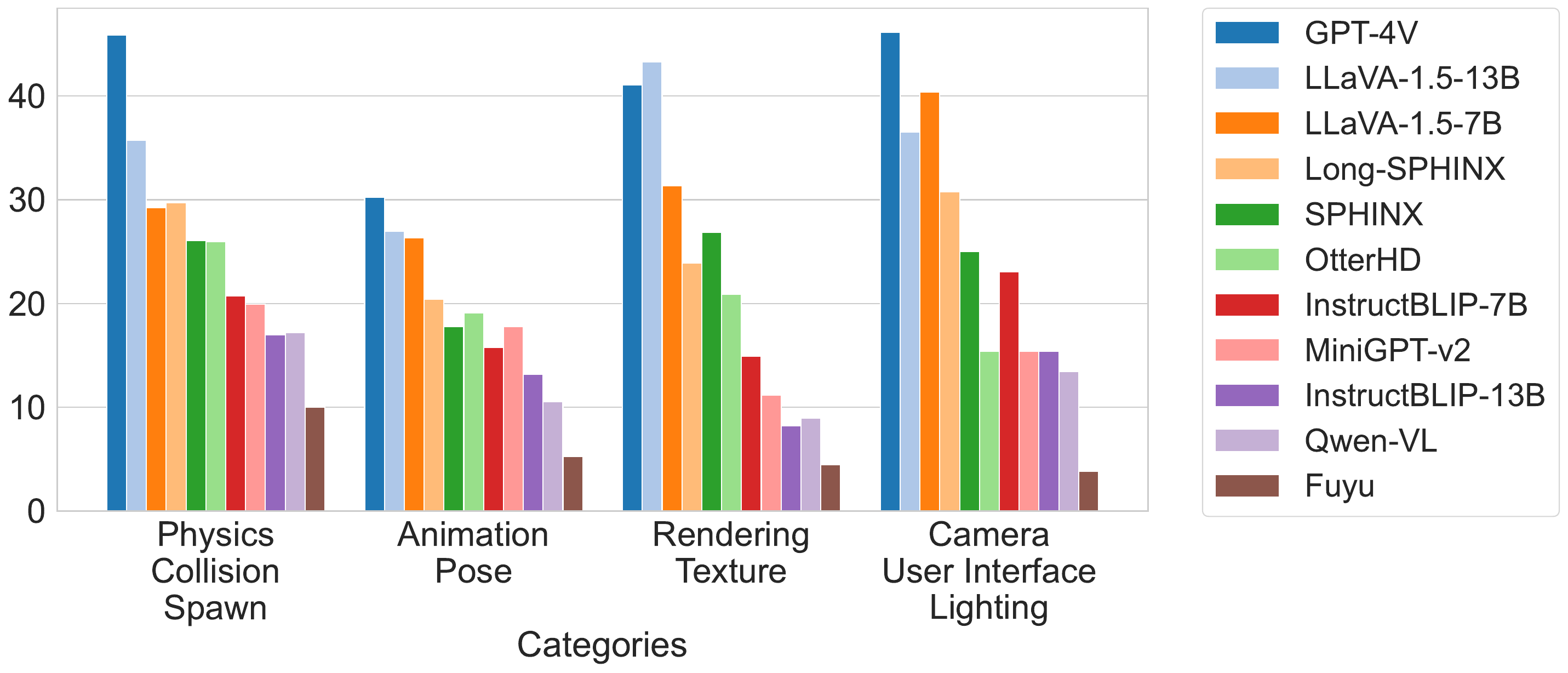}
    \caption{The performance of all tested models on different categories of images in \datasetname.}
    \label{fig:radar_plot_accuracy}
\end{figure}

\paragraph{Accuracy breakdown by category of glitches:} 
\cref{fig:radar_plot_accuracy} shows the breakdown of the performance of all tested models across the four studied glitch categories. GPT-4V is the best-performing model across all categories, with the exception of the \textit{Rendering and Texture} category, where LLaVA-1.5-13B slightly outperforms it by \increasenoparent{2.3} (41.0\% vs. 43.3\%). Overall, the \textit{Animation and Pose} category consistently proves to be the most challenging. This category contains images of characters in unusual poses, distorted body joints, or twisted bodies (see an example in \cref{fig:gpt4fails_basketball}).

\subsection{Qualitative Observations and Analysis}

\paragraph{Failing to reason about unusual aspects of the image:}
We observed that in several cases, particularly in open-source models, the model reports phrases such as ``\textit{the problem with this image is that it is computer-generated}'' or ``\textit{this is not an actual scene but a scene from a video game}'', along with similar phrases conveying the same meaning. These phrases suggest that, despite the model's ability to \textit{see} the content of the image, the language component of the model completely fails to reason about the content of the image.

Another observation is that InstructBLIP-13B often responds with ``\textit{nothing}'' or similar phrases and completely fails to reason about the image. This is the reason why the smaller InstructBLIP-7B can achieve higher accuracy on \datasetname. (See~\cref{sec:supp-FailedReasoning} for samples.)

\begin{figure}[t] 
\begin{ebox}{GPT-4V fails to recognize the absence of facial skin}
\centering
\includegraphics[width=\textwidth]{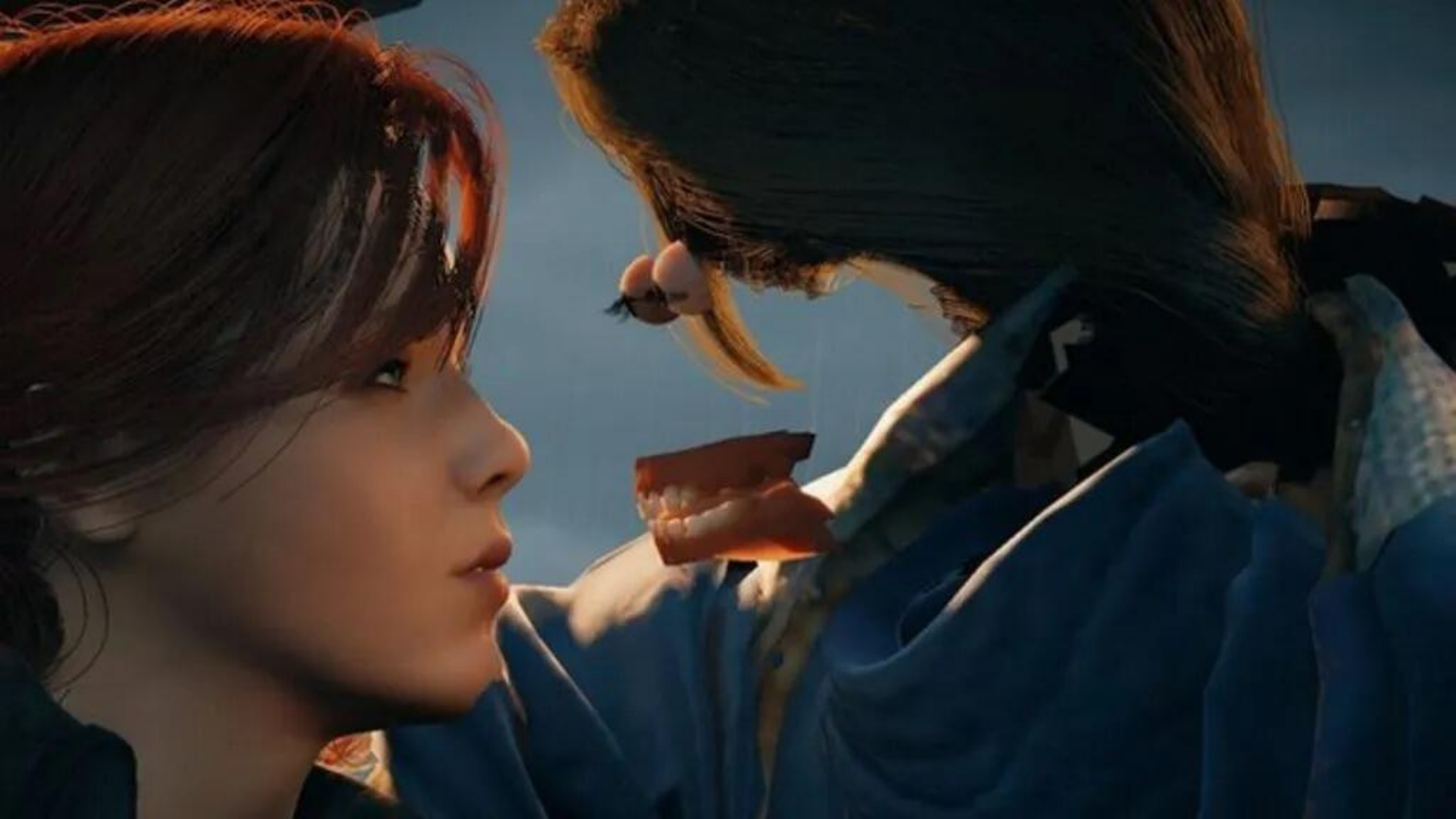}
\raggedright
\textbf{\textcolor{darkblue}{Question:}} What is wrong with this image?
\\
\textbf{\textcolor{darkgreen}{Ground truth:}} The character's face mesh is not rendered correctly.
\\
\textbf{\textcolor{darkred}{GPT-4V:}} 
The image appears to be from a video game or some form of CGI (computer-generated imagery) with very high-quality rendering. The characters depicted are very lifelike, suggesting a significant level of detail and investment in graphics. However, there's nothing inherently ``wrong" with the image in terms of content; it seems to portray a close-up moment between two characters.
\\
\textbf{\textcolor{darkyellow}{Llama-2 (Judge):}}
No, the generated description does not accurately convey the key information and themes of the ground truth. While it acknowledges the high quality of the graphics and the lifelike nature of the characters, it does not address the issue of the character's face mesh not being rendered correctly, which is the main focus of the ground truth.
\end{ebox}
\caption{One of the several cases in which GPT-4V fails to detect a problem with facial features.}
\label{fig:facelesskiss}
\end{figure}

\paragraph{GPT-4V struggles with faces:} 
GPT-4V is the best-performing model, yet it struggles with characters' faces, as shown in~\cref{fig:facelesskiss}.
We found several issues when processing glitches related to faces, and in the majority of cases, GPT-4V fails to detect the glitch and sometimes hallucinates about characters wearing costumes (\cref{suppfig:facelesskiss2}), where there are basically no discernible facial features.
On the other hand, smaller open-source models can sometimes detect glitches where GPT-4V fails, but they cannot describe the glitch clearly.
We hypothesize that this might be due to the privacy features of GPT-4V, preventing it from seeing the face clearly
(see~\cref{sec:supp-fail_to_detect_face_glitches} for more samples).

\begin{figure}[t] 
\begin{ebox}{GPT-4V fails to detect unusual body configuration}
\centering
\includegraphics[width=1\textwidth]{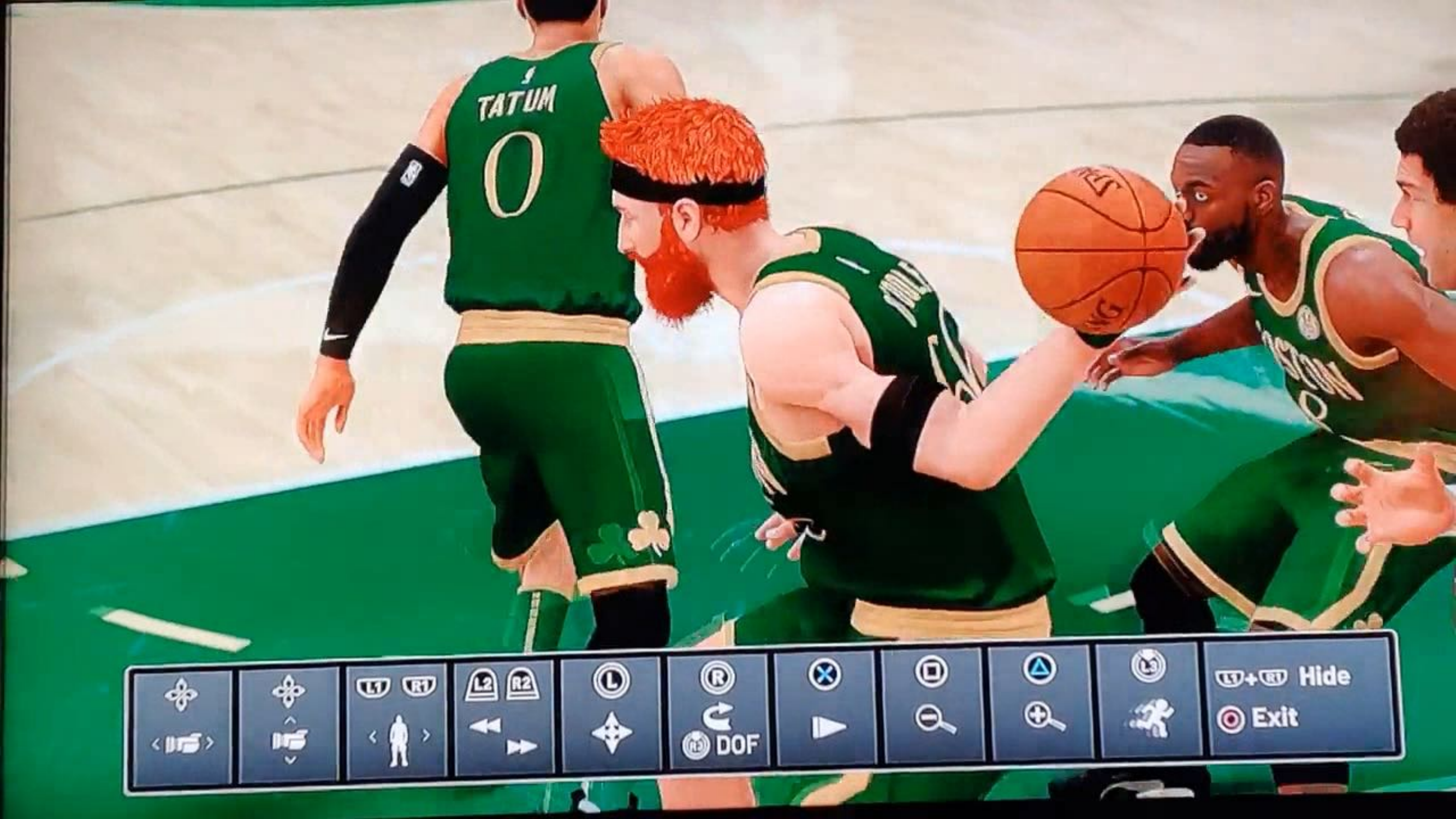}
\raggedright
\textbf{\textcolor{darkblue}{Question:}} What is unusual about this image?
\\
\textbf{\textcolor{darkgreen}{Ground truth:}} The left hand of the basketball player is in an unusual position and is stretched backward.
\\
\textbf{\textcolor{darkred}{GPT-4V:}} 
The image appears to be a screenshot from a basketball video game, possibly from the NBA 2K series. The unusual aspects of this image include: One of the players has a bright orange beard that stands out from the rest. Another player seems to be kissing or blowing at the basketball, which is not a typical basketball move or gesture. The characters and graphics look computer-generated, which indicates it's from a video game rather than a real-life game.
\\
The interface at the bottom of the image shows various controls and buttons, which further suggests that it's from a video game.
\\
\textbf{\textcolor{darkyellow}{Llama-2 (Judge):}}
No, the generated description does not accurately convey the key information and themes of the ground truth. The ground truth focuses on the unusual position of the basketball player's left hand, while the generated description highlights three unusual aspects of the image, none of which relate to the player's hand position.
\end{ebox}
\caption{The image shows a basketball player with an unnatural, impossible elbow pose. 
GPT-4V fails to focus on small details such as body configuration and is unable to report this issue.}
\label{fig:gpt4fails_basketball}
\end{figure}

\paragraph{Poor performance in the \textit{Animation and Pose} category:}
\cref{fig:radar_plot_accuracy} shows that \textit{Animation and Pose} is the hardest category of glitches.
During manual analysis, we found that LMMs struggle to detect unnatural body and limb configurations and incorrect animations being displayed.
For instance,~\cref{fig:gpt4fails_basketball} shows an example of an unnatural arm position for a basketball player that GPT-4V cannot detect.
This category can be further divided into three subcategories: (1) Heavily distorted body, \eg, when a character's body is overstretched or expanded; (2) Nuances in body configuration, such as contorted or twisted limbs and hands (\cref{fig:gpt4fails_basketball}); and (3) Characters playing animations with invisible props (\eg, a missing gun in the hand, as shown in~\cref{fig:missing_gun}).

\vspace{-3mm}

\paragraph{Prevalent hallucination in open-source models:}
\textit{Hallucination} typically refers to situations in which the model's generated text contains information not present in the image~\cite{rohrbach2018object, zhao2021evaluation, zhou2023analyzing, cui2023holistic}.
We noticed that open-source models often hallucinate extra objects or content in the image, e.g., we found that Fuyu's responses almost always contain hallucinations (see~\cref{sec:supp-Multimodal_Hallucination}).
The hallucination can contain additional characters or entirely new objects. These extra elements sometimes mislead Llama-2 into accepting an incorrect response as correct (see~\cref{sec:supp-judge_fails}).

\paragraph{Detecting some glitches requires paying attention to small details:}
Different glitches in our dataset require varying levels of visual attention. For example, when a car is flying in the air (\cref{suppxfig:multimodal_hallucination_3}), it usually occupies a large portion of the pixels on the screen, and models can easily pick up on such issues. This type of glitch is easier to catch, and GPT-4V, in particular, is very accurate at detecting it.

Some types of glitches require close attention to detail, such as clipping issues with clothing, where parts of the clothes intersect with the face or body of a character.
While GPT-4V is generally the best model for detecting clipping, it is still not perfect. In some cases, GPT-4V misses the clipping, and in other cases, it hallucinates about clipping (see \cref{sec:supp-clipping_issues_and_details} for samples).

\section{Discussion and Limitation}
\label{sec:discussions}

\paragraph{Comparing \datasetname with other benchmarks:}
The performance of various models across different benchmarks is presented in~\cref{tab:comparing_benchmarking_datasets}. It becomes evident that GPT-4V shows different performance against open-source models compared to \datasetname.
For instance, on VQAv2, LLaVA-1.5 and QWEN-VL score \increasenoparent{5.8} (80.0\% vs 74.2\%) and \increasenoparent{5.3} pp (79.5\% vs 74.2\%) higher than GPT-4V, respectively. However, on \datasetname, they lag behind by \decreasenoparent{9.9} (33.4\% vs. 43.5\%) and \decreasenoparent{28} pp (15.4\% vs. 43.4\%). The most notable gap is seen in Fuyu's performance against GPT-4V: while Fuyu exceeds on both OKVQA and AI2D, it significantly lags behind on \datasetname with only 8.5\% compared to GPT-4V's 43.4\%.

In sum, across multiple existing LMM benchmarks, open-source models can perform on par with or even surpass GPT-4V. However, their performance on \datasetname, which is derived from a real-world task in game quality assurance, falls significantly short of GPT-4V. In other words, the performance of models in real-world settings does not correlate well with existing benchmarks. This discrepancy partly comes from the design choices typical of LMM benchmarks, as they often opt for Yes/No or multiple-choice formats~\cite{li2023seed, liu2023mmbench, fu2023mme}. These formats allow models to find shortcuts for scoring high without necessarily generalizing well to other tasks.


\paragraph{Limitation:} 
We constructed our dataset by randomly sampling videos and observed a prevalence of video games with an open-world genre on the Reddit website. Consequently, during our sampling process, video games from this genre, characterized by their distinct mechanics, were more frequently represented compared to other types.

\begin{table}[t]
\centering
\caption{Comparing \datasetname with other visual benchmarks --- the bold numbers show the best model per benchmark (\%)}
\label{tab:comparing_benchmarking_datasets}
\resizebox{0.5\textwidth}{!}{%
\begin{tabular}{lrrrrrrr}
\toprule
\textbf{Model/Task} & \multicolumn{1}{c}{\textbf{\texttt{Glitch}}} & \multicolumn{1}{c}{\textbf{VQAv2}} & \multicolumn{1}{c}{\textbf{OKVQA}} & \multicolumn{1}{c}{\textbf{AI2D}} & \multicolumn{1}{c}{\textbf{SEED}}  & \multicolumn{1}{c}{\textbf{POPE}} & \multicolumn{1}{c}{\textbf{MMB}} \\ 
                    & \multicolumn{1}{c}{(Ours)}                      & \multicolumn{1}{c}{\cite{goyal2017making}} & \multicolumn{1}{c}{\cite{schwenk2022okvqa}} & \multicolumn{1}{c}{\cite{kembhavi2016diagram}} & \multicolumn{1}{c}{\cite{li2023seed}} & \multicolumn{1}{c}{\cite{li2023evaluating}} & \multicolumn{1}{c}{\cite{liu2023mmbench}} \\ 
\midrule
\textbf{GPT-4V} & \textbf{43.4} & 74.2 & 60.6 & 64.5 & - & - & - \\ 
\textbf{LLaVA} & 33.5 & \textbf{80.0} & - & - & 70.7 & - & \textbf{67.7} \\
\textbf{SPHINX} & 27.9 & - & - & - & 71.6 & \textbf{90.8} & 67.1 \\ 
\textbf{InstructBLIP}  & 19.6 & 62.1 & - & - & - & 78.9 & 36.0 \\ 
\textbf{MiniGPT} & 18.5 & - & 57.0 & - & - & -& - \\
\textbf{QWEN-VL} & 15.4 & 79.5 & 58.6 & 62.3 & 58.2 & - & 60.6 \\
\textbf{OtterHD} & 15.2 & - & - & - & - & 86.1 & 58.5 \\ 
\textbf{Fuyu} & 8.5 & 77.4 & \textbf{63.1} & \textbf{73.7} & - & - & - \\ 
\bottomrule
\end{tabular}%
}
\end{table}

\section{Conclusion}
\label{sec:conclusion}

We introduce \datasetname, a new challenging benchmark for evaluating multimodal models on the video game glitch detection task. Detecting glitches requires various levels of reasoning skills, such as an understanding of the laws of physics and commonsense, making it well-suited for testing the generalization capabilities of large multimodal models.
Comparing models' performance on various multimodal benchmarks and \datasetname reveals a disparity: High performance on prior benchmarks does not guarantee high performance on real-world tasks that demand extensive reasoning abilities.
We show that \datasetname, derived from real-world video game quality assurance, presents a new challenge for the AI community and is a valuable addition to existing multimodal benchmarks.

\clearpage
\section*{Acknowledgement}
AN is supported by the NaphCare Foundation, Adobe Research gifts, and NSF grant no. 2145767.

{
    \small
    \bibliographystyle{ieeenat_fullname}
    \bibliography{references}

\begin{thebibliography}{85}
\providecommand{\natexlab}[1]{#1}
\providecommand{\url}[1]{\texttt{#1}}
\expandafter\ifx\csname urlstyle\endcsname\relax
  \providecommand{\doi}[1]{doi: #1}\else
  \providecommand{\doi}{doi: \begingroup \urlstyle{rm}\Url}\fi

\bibitem[gam()]{gamers_epic}
Report: Epic games business breakdown \& founding story.
\newblock \url{https://research.contrary.com/reports/epic-games}.
\newblock (Accessed on 11/15/2023).

\bibitem[gpt(2023)]{gpt4v}
{OpenAI's GPT-4V(ision)}, 2023.

\bibitem[Agrawal et~al.(2019)Agrawal, Desai, Wang, Chen, Jain, Johnson, Batra, Parikh, Lee, and Anderson]{agrawal2019nocaps}
Harsh Agrawal, Karan Desai, Yufei Wang, Xinlei Chen, Rishabh Jain, Mark Johnson, Dhruv Batra, Devi Parikh, Stefan Lee, and Peter Anderson.
\newblock Nocaps: Novel object captioning at scale.
\newblock In \emph{Proceedings of the IEEE/CVF international conference on computer vision}, pages 8948--8957, 2019.

\bibitem[Alayrac et~al.(2022)Alayrac, Donahue, Luc, Miech, Barr, Hasson, Lenc, Mensch, Millican, Reynolds, et~al.]{alayrac2022flamingo}
Jean-Baptiste Alayrac, Jeff Donahue, Pauline Luc, Antoine Miech, Iain Barr, Yana Hasson, Karel Lenc, Arthur Mensch, Katherine Millican, Malcolm Reynolds, et~al.
\newblock Flamingo: a visual language model for few-shot learning.
\newblock \emph{Advances in Neural Information Processing Systems}, 35:\penalty0 23716--23736, 2022.

\bibitem[Awadalla et~al.(2023)Awadalla, Gao, Gardner, Hessel, Hanafy, Zhu, Marathe, Bitton, Gadre, Sagawa, et~al.]{awadalla2023openflamingo}
Anas Awadalla, Irena Gao, Josh Gardner, Jack Hessel, Yusuf Hanafy, Wanrong Zhu, Kalyani Marathe, Yonatan Bitton, Samir Gadre, Shiori Sagawa, et~al.
\newblock Openflamingo: An open-source framework for training large autoregressive vision-language models.
\newblock \emph{arXiv preprint arXiv:2308.01390}, 2023.

\bibitem[Bai et~al.(2023)Bai, Bai, Yang, Wang, Tan, Wang, Lin, Zhou, and Zhou]{bai2023qwen}
Jinze Bai, Shuai Bai, Shusheng Yang, Shijie Wang, Sinan Tan, Peng Wang, Junyang Lin, Chang Zhou, and Jingren Zhou.
\newblock Qwen-vl: A frontier large vision-language model with versatile abilities.
\newblock \emph{arXiv preprint arXiv:2308.12966}, 2023.

\bibitem[Bavishi et~al.(2023)Bavishi, Elsen, Hawthorne, Nye, Odena, Somani, and Ta\c{s}\i{}rlar]{fuyu-8b}
Rohan Bavishi, Erich Elsen, Curtis Hawthorne, Maxwell Nye, Augustus Odena, Arushi Somani, and Sa\u{g}nak Ta\c{s}\i{}rlar.
\newblock Introducing our multimodal models, 2023.

\bibitem[Bitton et~al.(2023)Bitton, Bansal, Hessel, Shao, Zhu, Awadalla, Gardner, Taori, and Schimdt]{bitton2023visit}
Yonatan Bitton, Hritik Bansal, Jack Hessel, Rulin Shao, Wanrong Zhu, Anas Awadalla, Josh Gardner, Rohan Taori, and Ludwig Schimdt.
\newblock Visit-bench: A benchmark for vision-language instruction following inspired by real-world use.
\newblock \emph{arXiv preprint arXiv:2308.06595}, 2023.

\bibitem[Bitton-Guetta et~al.(2023)Bitton-Guetta, Bitton, Hessel, Schmidt, Elovici, Stanovsky, and Schwartz]{bitton2023breaking}
Nitzan Bitton-Guetta, Yonatan Bitton, Jack Hessel, Ludwig Schmidt, Yuval Elovici, Gabriel Stanovsky, and Roy Schwartz.
\newblock Breaking common sense: Whoops! a vision-and-language benchmark of synthetic and compositional images.
\newblock In \emph{Proceedings of the IEEE/CVF International Conference on Computer Vision}, pages 2616--2627, 2023.

\bibitem[Bubeck et~al.(2023)Bubeck, Chandrasekaran, Eldan, Gehrke, Horvitz, Kamar, Lee, Lee, Li, Lundberg, et~al.]{bubeck2023sparks}
S{\'e}bastien Bubeck, Varun Chandrasekaran, Ronen Eldan, Johannes Gehrke, Eric Horvitz, Ece Kamar, Peter Lee, Yin~Tat Lee, Yuanzhi Li, Scott Lundberg, et~al.
\newblock Sparks of artificial general intelligence: Early experiments with gpt-4.
\newblock \emph{arXiv preprint arXiv:2303.12712}, 2023.

\bibitem[Chen et~al.(2023)Chen, Li, Zhang, Xiong, and Elhoseiny]{chen2023minigpt}
Jun Chen, Deyao Zhu1 Xiaoqian Shen1~Xiang Li, Zechun Liu2~Pengchuan Zhang, Raghuraman Krishnamoorthi2 Vikas Chandra2~Yunyang Xiong, and Mohamed Elhoseiny.
\newblock Minigpt-v2: Large language model as a unified interface for vision-language multi-task learning.
\newblock \emph{arXiv preprint arXiv:2310.09478}, 2023.

\bibitem[Chen et~al.(2021)Chen, Li, Chen, Fan, Hu, and Yang]{chen2021glib}
Ke Chen, Yufei Li, Yingfeng Chen, Changjie Fan, Zhipeng Hu, and Wei Yang.
\newblock Glib: towards automated test oracle for graphically-rich applications.
\newblock In \emph{Proceedings of the 29th ACM Joint Meeting on European Software Engineering Conference and Symposium on the Foundations of Software Engineering}, pages 1093--1104, 2021.

\bibitem[Chen et~al.(2015)Chen, Fang, Lin, Vedantam, Gupta, Doll{\'a}r, and Zitnick]{chen2015microsoft}
Xinlei Chen, Hao Fang, Tsung-Yi Lin, Ramakrishna Vedantam, Saurabh Gupta, Piotr Doll{\'a}r, and C~Lawrence Zitnick.
\newblock Microsoft coco captions: Data collection and evaluation server.
\newblock \emph{arXiv preprint arXiv:1504.00325}, 2015.

\bibitem[Choi et~al.(2012)Choi, Torralba, and Willsky]{CHOI2012853}
Myung~Jin Choi, Antonio Torralba, and Alan~S. Willsky.
\newblock Context models and out-of-context objects.
\newblock \emph{Pattern Recognition Letters}, 33\penalty0 (7):\penalty0 853--862, 2012.
\newblock Special Issue on Awards from ICPR 2010.

\bibitem[Cohen(1960)]{cohen1960coefficient}
Jacob Cohen.
\newblock A coefficient of agreement for nominal scales.
\newblock \emph{Educational and Psychological Measurement}, 20\penalty0 (1):\penalty0 37, 1960.

\bibitem[Cui et~al.(2023)Cui, Zhou, Yang, Wu, Zhang, Zou, and Yao]{cui2023holistic}
Chenhang Cui, Yiyang Zhou, Xinyu Yang, Shirley Wu, Linjun Zhang, James Zou, and Huaxiu Yao.
\newblock Holistic analysis of hallucination in gpt-4v (ision): Bias and interference challenges.
\newblock \emph{arXiv preprint arXiv:2311.03287}, 2023.

\bibitem[Dai et~al.(2023)Dai, Li, Li, Tiong, Zhao, Wang, Li, Fung, and Hoi]{dai2023instructblip}
Wenliang Dai, Junnan Li, Dongxu Li, Anthony Meng~Huat Tiong, Junqi Zhao, Weisheng Wang, Boyang Li, Pascale Fung, and Steven Hoi.
\newblock Instructblip: Towards general-purpose vision-language models with instruction tuning, 2023.

\bibitem[Frank et~al.(2021)Frank, Bugliarello, and Elliott]{frank2021vision}
Stella Frank, Emanuele Bugliarello, and Desmond Elliott.
\newblock Vision-and-language or vision-for-language? on cross-modal influence in multimodal transformers.
\newblock \emph{arXiv preprint arXiv:2109.04448}, 2021.

\bibitem[Fu et~al.(2023)Fu, Chen, Shen, Qin, Zhang, Lin, Qiu, Lin, Yang, Zheng, et~al.]{fu2023mme}
Chaoyou Fu, Peixian Chen, Yunhang Shen, Yulei Qin, Mengdan Zhang, Xu Lin, Zhenyu Qiu, Wei Lin, Jinrui Yang, Xiawu Zheng, et~al.
\newblock Mme: A comprehensive evaluation benchmark for multimodal large language models.
\newblock \emph{arXiv preprint arXiv:2306.13394}, 2023.

\bibitem[Gokhale et~al.(2022)Gokhale, Palangi, Nushi, Vineet, Horvitz, Kamar, Baral, and Yang]{gokhale2022benchmarking}
Tejas Gokhale, Hamid Palangi, Besmira Nushi, Vibhav Vineet, Eric Horvitz, Ece Kamar, Chitta Baral, and Yezhou Yang.
\newblock Benchmarking spatial relationships in text-to-image generation.
\newblock \emph{arXiv preprint arXiv:2212.10015}, 2022.

\bibitem[Google(2023)]{google2023bard}
Google.
\newblock Bard, 2023.

\bibitem[Goyal et~al.(2017)Goyal, Khot, Summers-Stay, Batra, and Parikh]{goyal2017making}
Yash Goyal, Tejas Khot, Douglas Summers-Stay, Dhruv Batra, and Devi Parikh.
\newblock Making the v in vqa matter: Elevating the role of image understanding in visual question answering.
\newblock In \emph{Proceedings of the IEEE conference on computer vision and pattern recognition}, pages 6904--6913, 2017.

\bibitem[Gurari et~al.(2018)Gurari, Li, Stangl, Guo, Lin, Grauman, Luo, and Bigham]{gurari2018vizwiz}
Danna Gurari, Qing Li, Abigale~J Stangl, Anhong Guo, Chi Lin, Kristen Grauman, Jiebo Luo, and Jeffrey~P Bigham.
\newblock Vizwiz grand challenge: Answering visual questions from blind people.
\newblock In \emph{Proceedings of the IEEE conference on computer vision and pattern recognition}, pages 3608--3617, 2018.

\bibitem[Hendrycks and Dietterich(2019)]{hendrycks2019benchmarking}
Dan Hendrycks and Thomas Dietterich.
\newblock Benchmarking neural network robustness to common corruptions and perturbations.
\newblock \emph{arXiv preprint arXiv:1903.12261}, 2019.

\bibitem[Hendrycks et~al.(2021)Hendrycks, Basart, Mu, Kadavath, Wang, Dorundo, Desai, Zhu, Parajuli, Guo, et~al.]{hendrycks2021many}
Dan Hendrycks, Steven Basart, Norman Mu, Saurav Kadavath, Frank Wang, Evan Dorundo, Rahul Desai, Tyler Zhu, Samyak Parajuli, Mike Guo, et~al.
\newblock The many faces of robustness: A critical analysis of out-of-distribution generalization.
\newblock In \emph{Proceedings of the IEEE/CVF International Conference on Computer Vision}, pages 8340--8349, 2021.

\bibitem[Hiippala et~al.(2021)Hiippala, Alikhani, Haverinen, Kalliokoski, Logacheva, Orekhova, Tuomainen, Stone, and Bateman]{hiippala2021ai2d}
Tuomo Hiippala, Malihe Alikhani, Jonas Haverinen, Timo Kalliokoski, Evanfiya Logacheva, Serafina Orekhova, Aino Tuomainen, Matthew Stone, and John~A Bateman.
\newblock Ai2d-rst: A multimodal corpus of 1000 primary school science diagrams.
\newblock \emph{Language Resources and Evaluation}, 55:\penalty0 661--688, 2021.

\bibitem[Hudson and Manning(2019)]{hudson2019gqa}
Drew~A Hudson and Christopher~D Manning.
\newblock Gqa: A new dataset for real-world visual reasoning and compositional question answering.
\newblock In \emph{Proceedings of the IEEE/CVF conference on computer vision and pattern recognition}, pages 6700--6709, 2019.

\bibitem[Jiang et~al.(2022)Jiang, Ma, Nie, Yu, Zhu, and Anandkumar]{jiang2022bongard}
Huaizu Jiang, Xiaojian Ma, Weili Nie, Zhiding Yu, Yuke Zhu, and Anima Anandkumar.
\newblock Bongard-hoi: Benchmarking few-shot visual reasoning for human-object interactions.
\newblock In \emph{Proceedings of the IEEE/CVF Conference on Computer Vision and Pattern Recognition}, pages 19056--19065, 2022.

\bibitem[Kamath et~al.(2023)Kamath, Hessel, and Chang]{kamath2023s}
Amita Kamath, Jack Hessel, and Kai-Wei Chang.
\newblock What's" up" with vision-language models? investigating their struggle with spatial reasoning.
\newblock \emph{arXiv preprint arXiv:2310.19785}, 2023.

\bibitem[Kembhavi et~al.(2016)Kembhavi, Salvato, Kolve, Seo, Hajishirzi, and Farhadi]{kembhavi2016diagram}
Aniruddha Kembhavi, Mike Salvato, Eric Kolve, Minjoon Seo, Hannaneh Hajishirzi, and Ali Farhadi.
\newblock A diagram is worth a dozen images.
\newblock In \emph{Computer Vision--ECCV 2016: 14th European Conference, Amsterdam, The Netherlands, October 11--14, 2016, Proceedings, Part IV 14}, pages 235--251. Springer, 2016.

\bibitem[Kojima et~al.(2022)Kojima, Gu, Reid, Matsuo, and Iwasawa]{kojima2022large}
Takeshi Kojima, Shixiang~Shane Gu, Machel Reid, Yutaka Matsuo, and Yusuke Iwasawa.
\newblock Large language models are zero-shot reasoners.
\newblock \emph{Advances in neural information processing systems}, 35:\penalty0 22199--22213, 2022.

\bibitem[Lewis et~al.(2010)Lewis, Whitehead, and Wardrip-Fruin]{lewis2010went}
Chris Lewis, Jim Whitehead, and Noah Wardrip-Fruin.
\newblock What went wrong: a taxonomy of video game bugs.
\newblock In \emph{Proceedings of the fifth international conference on the foundations of digital games}, pages 108--115, 2010.

\bibitem[Li et~al.(2023{\natexlab{a}})Li, Wang, Wang, Ge, Ge, and Shan]{li2023seed}
Bohao Li, Rui Wang, Guangzhi Wang, Yuying Ge, Yixiao Ge, and Ying Shan.
\newblock Seed-bench: Benchmarking multimodal llms with generative comprehension.
\newblock \emph{arXiv preprint arXiv:2307.16125}, 2023{\natexlab{a}}.

\bibitem[Li et~al.(2023{\natexlab{b}})Li, Zhang, Yang, Zhang, Pu, and Liu]{li2023otterhd}
Bo Li, Peiyuan Zhang, Jingkang Yang, Yuanhan Zhang, Fanyi Pu, and Ziwei Liu.
\newblock Otterhd: A high-resolution multi-modality model.
\newblock 2023{\natexlab{b}}.

\bibitem[Li(2023)]{li2023large}
Chunyuan Li.
\newblock Large multimodal models: Notes on cvpr 2023 tutorial.
\newblock \emph{arXiv preprint arXiv:2306.14895}, 2023.

\bibitem[Li et~al.(2022)Li, Li, Xiong, and Hoi]{li2022blip}
Junnan Li, Dongxu Li, Caiming Xiong, and Steven Hoi.
\newblock Blip: Bootstrapping language-image pre-training for unified vision-language understanding and generation.
\newblock In \emph{International Conference on Machine Learning}, pages 12888--12900. PMLR, 2022.

\bibitem[Li et~al.(2023{\natexlab{c}})Li, Du, Zhou, Wang, Zhao, and Wen]{li2023evaluating}
Yifan Li, Yifan Du, Kun Zhou, Jinpeng Wang, Wayne~Xin Zhao, and Ji-Rong Wen.
\newblock Evaluating object hallucination in large vision-language models.
\newblock \emph{arXiv preprint arXiv:2305.10355}, 2023{\natexlab{c}}.

\bibitem[Lin et~al.(2023)Lin, Liu, Zhang, Gao, Qiu, Xiao, Qiu, Lin, Shao, Chen, Han, Huang, Zhang, He, Li, and Qiao]{lin2023sphinx}
Ziyi Lin, Chris Liu, Renrui Zhang, Peng Gao, Longtian Qiu, Han Xiao, Han Qiu, Chen Lin, Wenqi Shao, Keqin Chen, Jiaming Han, Siyuan Huang, Yichi Zhang, Xuming He, Hongsheng Li, and Yu Qiao.
\newblock Sphinx: The joint mixing of weights, tasks, and visual embeddings for multi-modal large language models, 2023.

\bibitem[Ling et~al.(2020)Ling, Tollmar, and Gissl{\'e}n]{ling2020using}
Carlos Ling, Konrad Tollmar, and Linus Gissl{\'e}n.
\newblock Using deep convolutional neural networks to detect rendered glitches in video games.
\newblock In \emph{Proceedings of the AAAI Conference on Artificial Intelligence and Interactive Digital Entertainment}, pages 66--73, 2020.

\bibitem[Liu et~al.(2023{\natexlab{a}})Liu, Guan, Li, Chen, Yacoob, Manocha, and Zhou]{liu2023hallusionbench}
Fuxiao Liu, Tianrui Guan, Zongxia Li, Lichang Chen, Yaser Yacoob, Dinesh Manocha, and Tianyi Zhou.
\newblock Hallusionbench: You see what you think? or you think what you see? an image-context reasoning benchmark challenging for gpt-4v (ision), llava-1.5, and other multi-modality models.
\newblock \emph{arXiv preprint arXiv:2310.14566}, 2023{\natexlab{a}}.

\bibitem[Liu et~al.(2023{\natexlab{b}})Liu, Lin, Li, Wang, Yacoob, and Wang]{liu2023aligning}
Fuxiao Liu, Kevin Lin, Linjie Li, Jianfeng Wang, Yaser Yacoob, and Lijuan Wang.
\newblock Aligning large multi-modal model with robust instruction tuning.
\newblock \emph{arXiv preprint arXiv:2306.14565}, 2023{\natexlab{b}}.

\bibitem[Liu et~al.(2023{\natexlab{c}})Liu, Li, Li, and Lee]{liu2023improved}
Haotian Liu, Chunyuan Li, Yuheng Li, and Yong~Jae Lee.
\newblock Improved baselines with visual instruction tuning.
\newblock \emph{arXiv preprint arXiv:2310.03744}, 2023{\natexlab{c}}.

\bibitem[Liu et~al.(2023{\natexlab{d}})Liu, Li, Wu, and Lee]{liu2023visual}
Haotian Liu, Chunyuan Li, Qingyang Wu, and Yong~Jae Lee.
\newblock Visual instruction tuning.
\newblock \emph{arXiv preprint arXiv:2304.08485}, 2023{\natexlab{d}}.

\bibitem[Liu et~al.(2023{\natexlab{e}})Liu, Cheng, Liu, Zhang, Li, Ren, Zou, Yang, Su, Zhu, et~al.]{liu2023llava}
Shilong Liu, Hao Cheng, Haotian Liu, Hao Zhang, Feng Li, Tianhe Ren, Xueyan Zou, Jianwei Yang, Hang Su, Jun Zhu, et~al.
\newblock Llava-plus: Learning to use tools for creating multimodal agents.
\newblock \emph{arXiv preprint arXiv:2311.05437}, 2023{\natexlab{e}}.

\bibitem[Liu et~al.(2023{\natexlab{f}})Liu, Duan, Zhang, Li, Zhang, Zhao, Yuan, Wang, He, Liu, et~al.]{liu2023mmbench}
Yuan Liu, Haodong Duan, Yuanhan Zhang, Bo Li, Songyang Zhang, Wangbo Zhao, Yike Yuan, Jiaqi Wang, Conghui He, Ziwei Liu, et~al.
\newblock Mmbench: Is your multi-modal model an all-around player?
\newblock \emph{arXiv preprint arXiv:2307.06281}, 2023{\natexlab{f}}.

\bibitem[Lu et~al.(2021)Lu, Qiu, Chen, Xia, Zhao, Zhang, Yu, Liang, and Zhu]{lu2021iconqa}
Pan Lu, Liang Qiu, Jiaqi Chen, Tony Xia, Yizhou Zhao, Wei Zhang, Zhou Yu, Xiaodan Liang, and Song-Chun Zhu.
\newblock Iconqa: A new benchmark for abstract diagram understanding and visual language reasoning.
\newblock \emph{arXiv preprint arXiv:2110.13214}, 2021.

\bibitem[Lu et~al.(2022)Lu, Mishra, Xia, Qiu, Chang, Zhu, Tafjord, Clark, and Kalyan]{lu2022learn}
Pan Lu, Swaroop Mishra, Tanglin Xia, Liang Qiu, Kai-Wei Chang, Song-Chun Zhu, Oyvind Tafjord, Peter Clark, and Ashwin Kalyan.
\newblock Learn to explain: Multimodal reasoning via thought chains for science question answering.
\newblock \emph{Advances in Neural Information Processing Systems}, 35:\penalty0 2507--2521, 2022.

\bibitem[Lu et~al.(2023)Lu, Bansal, Xia, Liu, Li, Hajishirzi, Cheng, Chang, Galley, and Gao]{lu2023mathvista}
Pan Lu, Hritik Bansal, Tony Xia, Jiacheng Liu, Chunyuan Li, Hannaneh Hajishirzi, Hao Cheng, Kai-Wei Chang, Michel Galley, and Jianfeng Gao.
\newblock Mathvista: Evaluating mathematical reasoning of foundation models in visual contexts.
\newblock \emph{arXiv preprint arXiv:2310.02255}, 2023.

\bibitem[Marino et~al.(2019)Marino, Rastegari, Farhadi, and Mottaghi]{marino2019ok}
Kenneth Marino, Mohammad Rastegari, Ali Farhadi, and Roozbeh Mottaghi.
\newblock Ok-vqa: A visual question answering benchmark requiring external knowledge.
\newblock In \emph{Proceedings of the IEEE/cvf conference on computer vision and pattern recognition}, pages 3195--3204, 2019.

\bibitem[Mishra et~al.(2019)Mishra, Shekhar, Singh, and Chakraborty]{mishra2019ocr}
Anand Mishra, Shashank Shekhar, Ajeet~Kumar Singh, and Anirban Chakraborty.
\newblock Ocr-vqa: Visual question answering by reading text in images.
\newblock In \emph{2019 international conference on document analysis and recognition (ICDAR)}, pages 947--952. IEEE, 2019.

\bibitem[Nantes et~al.(2008)Nantes, Brown, and Maire]{nantes2008framework}
Alfredo Nantes, Ross Brown, and Frederic Maire.
\newblock A framework for the semi-automatic testing of video games.
\newblock In \emph{Proceedings of the AAAI Conference on Artificial Intelligence and Interactive Digital Entertainment}, pages 197--202, 2008.

\bibitem[Peng et~al.(2023)Peng, Wang, Dong, Hao, Huang, Ma, and Wei]{peng2023kosmos}
Zhiliang Peng, Wenhui Wang, Li Dong, Yaru Hao, Shaohan Huang, Shuming Ma, and Furu Wei.
\newblock Kosmos-2: Grounding multimodal large language models to the world.
\newblock \emph{arXiv preprint arXiv:2306.14824}, 2023.

\bibitem[Plummer et~al.(2015)Plummer, Wang, Cervantes, Caicedo, Hockenmaier, and Lazebnik]{plummer2015flickr30k}
Bryan~A Plummer, Liwei Wang, Chris~M Cervantes, Juan~C Caicedo, Julia Hockenmaier, and Svetlana Lazebnik.
\newblock Flickr30k entities: Collecting region-to-phrase correspondences for richer image-to-sentence models.
\newblock In \emph{Proceedings of the IEEE international conference on computer vision}, pages 2641--2649, 2015.

\bibitem[Qin et~al.(2023)Qin, Ji, Khan, Fan, Khan, and Gool]{Qin_2023}
Haotong Qin, Ge-Peng Ji, Salman Khan, Deng-Ping Fan, Fahad~Shahbaz Khan, and Luc~Van Gool.
\newblock How good is google bard's visual understanding? an empirical study on open challenges.
\newblock \emph{Machine Intelligence Research}, 20\penalty0 (5):\penalty0 605--613, 2023.

\bibitem[Rahman(2023)]{rahman2023weak}
Farrukh Rahman.
\newblock Weak supervision for label efficient visual bug detection.
\newblock \emph{arXiv preprint arXiv:2309.11077}, 2023.

\bibitem[Rani et~al.(2023)Rani, Pandey, Wagde, and Dhaka]{rani2023deep}
Geeta Rani, Upasana Pandey, Aniket~Anil Wagde, and Vijaypal~Singh Dhaka.
\newblock A deep reinforcement learning technique for bug detection in video games.
\newblock \emph{International Journal of Information Technology}, 15\penalty0 (1):\penalty0 355--367, 2023.

\bibitem[Research(2023)]{GrandViewResearch2023}
Grand~View Research.
\newblock Video game market size, share and growth report, 2030, 2023.

\bibitem[Rohrbach et~al.(2018)Rohrbach, Hendricks, Burns, Darrell, and Saenko]{rohrbach2018object}
Anna Rohrbach, Lisa~Anne Hendricks, Kaylee Burns, Trevor Darrell, and Kate Saenko.
\newblock Object hallucination in image captioning.
\newblock \emph{arXiv preprint arXiv:1809.02156}, 2018.

\bibitem[Russakovsky et~al.(2015)Russakovsky, Deng, Su, Krause, Satheesh, Ma, Huang, Karpathy, Khosla, Bernstein, et~al.]{russakovsky2015imagenet}
Olga Russakovsky, Jia Deng, Hao Su, Jonathan Krause, Sanjeev Satheesh, Sean Ma, Zhiheng Huang, Andrej Karpathy, Aditya Khosla, Michael Bernstein, et~al.
\newblock Imagenet large scale visual recognition challenge.
\newblock \emph{International journal of computer vision}, 115:\penalty0 211--252, 2015.

\bibitem[Schwenk et~al.(2022)Schwenk, Khandelwal, Clark, Marino, and Mottaghi]{schwenk2022okvqa}
Dustin Schwenk, Apoorv Khandelwal, Christopher Clark, Kenneth Marino, and Roozbeh Mottaghi.
\newblock A-okvqa: A benchmark for visual question answering using world knowledge.
\newblock In \emph{European Conference on Computer Vision}, pages 146--162. Springer, 2022.

\bibitem[Shinn et~al.(2023)Shinn, Labash, and Gopinath]{shinn2023reflexion}
Noah Shinn, Beck Labash, and Ashwin Gopinath.
\newblock Reflexion: an autonomous agent with dynamic memory and self-reflection.
\newblock \emph{arXiv preprint arXiv:2303.11366}, 2023.

\bibitem[Sidorov et~al.(2020)Sidorov, Hu, Rohrbach, and Singh]{sidorov2020textcaps}
Oleksii Sidorov, Ronghang Hu, Marcus Rohrbach, and Amanpreet Singh.
\newblock Textcaps: a dataset for image captioning with reading comprehension.
\newblock In \emph{Computer Vision--ECCV 2020: 16th European Conference, Glasgow, UK, August 23--28, 2020, Proceedings, Part II 16}, pages 742--758. Springer, 2020.

\bibitem[Singh et~al.(2019)Singh, Natarajan, Shah, Jiang, Chen, Batra, Parikh, and Rohrbach]{singh2019towards}
Amanpreet Singh, Vivek Natarajan, Meet Shah, Yu Jiang, Xinlei Chen, Dhruv Batra, Devi Parikh, and Marcus Rohrbach.
\newblock Towards vqa models that can read.
\newblock In \emph{Proceedings of the IEEE/CVF conference on computer vision and pattern recognition}, pages 8317--8326, 2019.

\bibitem[Taesiri et~al.()Taesiri, Nguyen, Habchi, Bezemer, and Nguyen]{taesiriimagenet}
Mohammad~Reza Taesiri, Giang Nguyen, Sarra Habchi, Cor-Paul Bezemer, and Anh Nguyen.
\newblock Imagenet-hard: The hardest images remaining from a study of the power of zoom and spatial biases in image classification.

\bibitem[Taesiri et~al.(2020)Taesiri, Habibi, and Fazli]{taesiri2020video}
Mohammad~Reza Taesiri, Moslem Habibi, and Mohammad~Amin Fazli.
\newblock A video game testing method utilizing deep learning.
\newblock \emph{Iran Journal of Computer Science}, 17\penalty0 (2), 2020.

\bibitem[Taesiri et~al.(2022)Taesiri, Macklon, and Bezemer]{taesiri2022clip}
Mohammad~Reza Taesiri, Finlay Macklon, and Cor-Paul Bezemer.
\newblock Clip meets gamephysics: Towards bug identification in gameplay videos using zero-shot transfer learning.
\newblock In \emph{Proceedings of the 19th International Conference on Mining Software Repositories}, pages 270--281, 2022.

\bibitem[Thrush et~al.(2022)Thrush, Jiang, Bartolo, Singh, Williams, Kiela, and Ross]{thrush2022winoground}
Tristan Thrush, Ryan Jiang, Max Bartolo, Amanpreet Singh, Adina Williams, Douwe Kiela, and Candace Ross.
\newblock Winoground: Probing vision and language models for visio-linguistic compositionality.
\newblock In \emph{Proceedings of the IEEE/CVF Conference on Computer Vision and Pattern Recognition}, pages 5238--5248, 2022.

\bibitem[Touvron et~al.(2023)Touvron, Martin, Stone, Albert, Almahairi, Babaei, Bashlykov, Batra, Bhargava, Bhosale, et~al.]{touvron2023llama}
Hugo Touvron, Louis Martin, Kevin Stone, Peter Albert, Amjad Almahairi, Yasmine Babaei, Nikolay Bashlykov, Soumya Batra, Prajjwal Bhargava, Shruti Bhosale, et~al.
\newblock Llama 2: Open foundation and fine-tuned chat models.
\newblock \emph{arXiv preprint arXiv:2307.09288}, 2023.

\bibitem[Truelove et~al.(2021)Truelove, de~Almeida, and Ahmed]{truelove2021we}
Andrew Truelove, Eduardo~Santana de Almeida, and Iftekhar Ahmed.
\newblock We'll fix it in post: what do bug fixes in video game update notes tell us?
\newblock In \emph{2021 IEEE/ACM 43rd International Conference on Software Engineering (ICSE)}, pages 736--747. IEEE, 2021.

\bibitem[Wei et~al.(2022)Wei, Wang, Schuurmans, Bosma, Xia, Chi, Le, Zhou, et~al.]{wei2022chain}
Jason Wei, Xuezhi Wang, Dale Schuurmans, Maarten Bosma, Fei Xia, Ed Chi, Quoc~V Le, Denny Zhou, et~al.
\newblock Chain-of-thought prompting elicits reasoning in large language models.
\newblock \emph{Advances in Neural Information Processing Systems}, 35:\penalty0 24824--24837, 2022.

\bibitem[Wen et~al.(2023)Wen, Yang, Fu, Wang, Cai, Li, Ma, Li, Xu, Shang, Zhu, Sun, Bai, Cai, Dou, Hu, and Shi]{wen2023road}
Licheng Wen, Xuemeng Yang, Daocheng Fu, Xiaofeng Wang, Pinlong Cai, Xin Li, Tao Ma, Yingxuan Li, Linran Xu, Dengke Shang, Zheng Zhu, Shaoyan Sun, Yeqi Bai, Xinyu Cai, Min Dou, Shuanglu Hu, and Botian Shi.
\newblock On the road with gpt-4v(ision): Early explorations of visual-language model on autonomous driving, 2023.

\bibitem[Wilkins and Stathis(2022)]{wilkins2022learning}
Benedict Wilkins and Kostas Stathis.
\newblock Learning to identify perceptual bugs in 3d video games.
\newblock \emph{arXiv preprint arXiv:2202.12884}, 2022.

\bibitem[Wu et~al.(2023)Wu, Wang, Yang, Zheng, Zhang, Zhao, and Qin]{wu2023early}
Yang Wu, Shilong Wang, Hao Yang, Tian Zheng, Hongbo Zhang, Yanyan Zhao, and Bing Qin.
\newblock An early evaluation of gpt-4v (ision).
\newblock \emph{arXiv preprint arXiv:2310.16534}, 2023.

\bibitem[Yan et~al.(2023)Yan, Zhang, Zhou, He, Li, and Sun]{yan2023multimodal}
Zhiling Yan, Kai Zhang, Rong Zhou, Lifang He, Xiang Li, and Lichao Sun.
\newblock Multimodal chatgpt for medical applications: an experimental study of gpt-4v, 2023.

\bibitem[Yang et~al.(2023{\natexlab{a}})Yang, Zhang, Li, Zou, Li, and Gao]{yang2023set}
Jianwei Yang, Hao Zhang, Feng Li, Xueyan Zou, Chunyuan Li, and Jianfeng Gao.
\newblock Set-of-mark prompting unleashes extraordinary visual grounding in gpt-4v.
\newblock \emph{arXiv preprint arXiv:2310.11441}, 2023{\natexlab{a}}.

\bibitem[Yang et~al.(2023{\natexlab{b}})Yang, Ping, Liu, Korthikanti, Nie, Huang, Fan, Yu, Lan, Li, et~al.]{yang2023re}
Zhuolin Yang, Wei Ping, Zihan Liu, Vijay Korthikanti, Weili Nie, De-An Huang, Linxi Fan, Zhiding Yu, Shiyi Lan, Bo Li, et~al.
\newblock Re-vilm: Retrieval-augmented visual language model for zero and few-shot image captioning.
\newblock \emph{arXiv preprint arXiv:2302.04858}, 2023{\natexlab{b}}.

\bibitem[Ye et~al.(2023)Ye, Xu, Xu, Ye, Yan, Zhou, Wang, Hu, Shi, Shi, et~al.]{ye2023mplug}
Qinghao Ye, Haiyang Xu, Guohai Xu, Jiabo Ye, Ming Yan, Yiyang Zhou, Junyang Wang, Anwen Hu, Pengcheng Shi, Yaya Shi, et~al.
\newblock mplug-owl: Modularization empowers large language models with multimodality.
\newblock \emph{arXiv preprint arXiv:2304.14178}, 2023.

\bibitem[Yin et~al.(2023)Yin, Wang, Cao, Shi, Liu, Li, Sheng, Bai, Huang, Wang, et~al.]{yin2023lamm}
Zhenfei Yin, Jiong Wang, Jianjian Cao, Zhelun Shi, Dingning Liu, Mukai Li, Lu Sheng, Lei Bai, Xiaoshui Huang, Zhiyong Wang, et~al.
\newblock Lamm: Language-assisted multi-modal instruction-tuning dataset, framework, and benchmark.
\newblock \emph{arXiv preprint arXiv:2306.06687}, 2023.

\bibitem[Yu et~al.(2023)Yu, Yang, Li, Wang, Lin, Liu, Wang, and Wang]{yu2023mm}
Weihao Yu, Zhengyuan Yang, Linjie Li, Jianfeng Wang, Kevin Lin, Zicheng Liu, Xinchao Wang, and Lijuan Wang.
\newblock Mm-vet: Evaluating large multimodal models for integrated capabilities.
\newblock \emph{arXiv preprint arXiv:2308.02490}, 2023.

\bibitem[Zellers et~al.(2019)Zellers, Bisk, Farhadi, and Choi]{zellers2019recognition}
Rowan Zellers, Yonatan Bisk, Ali Farhadi, and Yejin Choi.
\newblock From recognition to cognition: Visual commonsense reasoning.
\newblock In \emph{Proceedings of the IEEE/CVF conference on computer vision and pattern recognition}, pages 6720--6731, 2019.

\bibitem[Zhao et~al.(2021)Zhao, Anderson, Jain, Wang, Ku, Baldridge, and Ie]{zhao2021evaluation}
Ming Zhao, Peter Anderson, Vihan Jain, Su Wang, Alexander Ku, Jason Baldridge, and Eugene Ie.
\newblock On the evaluation of vision-and-language navigation instructions.
\newblock \emph{arXiv preprint arXiv:2101.10504}, 2021.

\bibitem[Zheng et~al.(2023)Zheng, Chiang, Sheng, Zhuang, Wu, Zhuang, Lin, Li, Li, Xing, Zhang, Gonzalez, and Stoica]{zheng2023judging}
Lianmin Zheng, Wei-Lin Chiang, Ying Sheng, Siyuan Zhuang, Zhanghao Wu, Yonghao Zhuang, Zi Lin, Zhuohan Li, Dacheng Li, Eric Xing, Hao Zhang, Joseph~E. Gonzalez, and Ion Stoica.
\newblock Judging {LLM}-as-a-judge with {MT}-bench and chatbot arena.
\newblock In \emph{Thirty-seventh Conference on Neural Information Processing Systems Datasets and Benchmarks Track}, 2023.

\bibitem[Zheng et~al.(2019)Zheng, Xie, Su, Ma, Hao, Meng, Liu, Shen, Chen, and Fan]{zheng2019wuji}
Yan Zheng, Xiaofei Xie, Ting Su, Lei Ma, Jianye Hao, Zhaopeng Meng, Yang Liu, Ruimin Shen, Yingfeng Chen, and Changjie Fan.
\newblock Wuji: Automatic online combat game testing using evolutionary deep reinforcement learning.
\newblock In \emph{2019 34th IEEE/ACM International Conference on Automated Software Engineering (ASE)}, pages 772--784. IEEE, 2019.

\bibitem[Zhou et~al.(2023{\natexlab{a}})Zhou, Lai, Yeong, Mouratidis, and Jiang]{zhou2023rome}
Kankan Zhou, Eason Lai, Wei Bin~Au Yeong, Kyriakos Mouratidis, and Jing Jiang.
\newblock Rome: Evaluating pre-trained vision-language models on reasoning beyond visual common sense.
\newblock \emph{arXiv preprint arXiv:2310.19301}, 2023{\natexlab{a}}.

\bibitem[Zhou et~al.(2023{\natexlab{b}})Zhou, Cui, Yoon, Zhang, Deng, Finn, Bansal, and Yao]{zhou2023analyzing}
Yiyang Zhou, Chenhang Cui, Jaehong Yoon, Linjun Zhang, Zhun Deng, Chelsea Finn, Mohit Bansal, and Huaxiu Yao.
\newblock Analyzing and mitigating object hallucination in large vision-language models.
\newblock \emph{arXiv preprint arXiv:2310.00754}, 2023{\natexlab{b}}.

\end{thebibliography}
}


\clearpage
\setcounter{page}{1}
\maketitlesupplementary


\newcommand{\beginsupplementary}{%
    \setcounter{table}{0}
    \renewcommand{\thetable}{A\arabic{table}}%
    
    \setcounter{figure}{0}
    \renewcommand{\thefigure}{A\arabic{figure}}%
    
    \setcounter{section}{0}
    \renewcommand{\thesection}{A\arabic{section}}
    \renewcommand{\thesubsection}{\thesection.\arabic{subsection}}
}
\beginsupplementary%
\setcounter{figure}{0}
\renewcommand{\thefigure}{A\arabic{figure}}%
\renewcommand{\theHfigure}{SuppFigureA\arabic{figure}} 

\setcounter{table}{0}
\renewcommand{\thetable}{A\arabic{table}}%
\renewcommand{\theHtable}{SuppTableA\arabic{table}} 

\setcounter{section}{0}
\renewcommand{\thesection}{A\arabic{section}}
\renewcommand{\thesubsection}{\thesection.\arabic{subsection}}
\renewcommand{\theHsection}{SuppSectionA\arabic{section}} 

\section{Implementation Details}
\label{sec:supp-immplementation_details}

\subsection{Details about model inference}

For each open source model, we used the provided sample code and demo from their respective repositories.
Minor modifications were made to enable automatic processing of all images with designated prompts.
The results were then stored in individual CSV files for each model.
For OtterHD, which offers an API, we used the API to submit each image along with the appropriate prompt and recorded the responses.
Our experiment was done prior to the official release of the GPT-4V API, and we used the ChatGPT web version for the benchmark, using a Chrome extension to assist in the process.

We kept the temperature and other parameters of each model unchanged. The only modification involved increasing the \texttt{max\_token} limit, ensuring that the model's response length was not restricted.

\subsection{Details about the judge}

In our experiment, the Llama-2-70B model served as the judge. We utilized the API from \href{https://www.perplexity.ai/}{perplexity.ai}, which is compatible with OpenAI's Python package.
Additionally, we employed a custom system message, as detailed below:

\begin{mdframed}
    \emph{Your task is to compare a model-generated text with a ground truth reference, assessing whether the key information and themes are similarly conveyed, even if worded differently. Focus on semantic content, thematic alignment, and intent, rather than exact phrasing or word usage. Recognize synonyms, paraphrases, and different stylistic expressions as valid, provided they faithfully represent the ground truth's meaning. Offer feedback on the correlation between the texts and suggest improvements for alignment, while appreciating creative or varied linguistic expression that maintains the essence of the ground truth.
    \\
    \\
    First analyze, then report the final answer in either of Yes or No}
\end{mdframed}

\clearpage
\section{Additional Results}
\label{sec:supp-additional_results}

\subsection{Breakdown of Performance by Various Glitch Types}

\begin{table}[h]
\centering
\caption{Breakdown of Performance for Different LMMs by Various Glitch Types (\%)}
\label{tab:perf_categorized_by_typs}
\resizebox{0.8\textwidth}{!}{%
\begin{tabular}{@{}lrrrrr@{}}
\toprule
 &
  \multicolumn{1}{c}{\textbf{\begin{tabular}[c]{@{}c@{}}Camera,\\ User Interface,\\ Lighting\end{tabular}}} &
  \multicolumn{1}{c}{\textbf{\begin{tabular}[c]{@{}c@{}}Animation,\\ Pose\end{tabular}}} &
  \multicolumn{1}{c}{\textbf{\begin{tabular}[c]{@{}c@{}}Physics,\\ Collision,\\ Spawn\end{tabular}}} &
  \multicolumn{1}{c}{\textbf{\begin{tabular}[c]{@{}c@{}}Rendering,\\ Texture\end{tabular}}} &
  \multicolumn{1}{c}{\textbf{\begin{tabular}[c]{@{}c@{}}Average\\ Performance\end{tabular}}} \\ \midrule
GPT-4V           & 46.2 & 30.3 & 45.9 & \multicolumn{1}{r|}{41.0} & 40.8 \\
LLaVA-1.5-13B    & 36.5 & 27.0 & 35.7 & \multicolumn{1}{r|}{43.3} & 35.6 \\
LLaVA-1.5-7B     & 40.4 & 26.3 & 29.2 & \multicolumn{1}{r|}{31.3} & 31.8 \\
Long-SPHINX      & 30.8 & 20.4 & 29.7 & \multicolumn{1}{r|}{23.9} & 26.2 \\
SPHINX           & 25.0 & 17.8 & 26.1 & \multicolumn{1}{r|}{26.9} & 23.9 \\
OtterHD          & 15.4 & 19.1 & 25.9 & \multicolumn{1}{r|}{20.9} & 20.3 \\
InstructBLIP-7B  & 23.1 & 15.8 & 20.8 & \multicolumn{1}{r|}{14.9} & 18.6 \\
MiniGPT-v2       & 15.4 & 17.8 & 19.9 & \multicolumn{1}{r|}{11.2} & 16.1 \\
InstructBLIP-13B & 15.4 & 13.2 & 17.0 & \multicolumn{1}{r|}{8.2}  & 13.4 \\
Qwen-VL          & 13.5 & 10.5 & 17.2 & \multicolumn{1}{r|}{9.0}  & 12.5 \\
Fuyu             & 3.8  & 5.3  & 10.0 & \multicolumn{1}{r|}{4.5}  & 5.9  \\ \bottomrule
\end{tabular}%
}
\end{table}

\subsection{Using multiple frames}


As explained in \cref{sec:method}, we went through our data to ensure that the glitches could be detected without using temporal information.
To investigate whether current LMMs can effectively utilize temporal information between several input images, we conducted an experiment with a random sample of 150 videos. We extracted frames one second before and after the glitch from each video. We included the three frames and added `\texttt{Given the sequence of images}' to the original GPT-4V prompt. The accuracy dropped to 36\% (from 39\%) for Q1 and to 28\% (from 35\%) for Q2. This suggests that current LMM performance cannot be improved by providing multiple frames.

\clearpage
\section{Glossary of Model Failures}
\label{sec:supp-Glossary_of_Model_Failures}

In this section, we offer a summary of our qualitative analysis, categorizing failures in multiple dimensions. These include instances where LMMs either fail to detect glitches or to note unusual aspects of images, as well as instances where the judge incorrectly labels the model's responses as correct.
Additionally, we address issues such as models producing hallucinations and other shortcomings.


\changelinkcolor{blue}


\begin{itemize}[leftmargin=1cm]  
  \item[\faArrowCircleRight \color{gray}] \textcolor{blue}{\small \hyperref[sec:supp-FailedReasoning]{Failures Related to Reasoning About the Content of the Image}}
  \item[\faArrowCircleRight \color{gray}] \textcolor{blue}{\small \hyperref[sec:supp-fail_to_detect_face_glitches]{Failures Related to Facial Features}}
  \item[\faArrowCircleRight \color{gray}] \textcolor{blue}{\small \hyperref[sec:supp-fail_to_detect_unnatrual_body_pose]{Failures Related to Unnatural Body Positions}}
  \item[\faArrowCircleRight \color{gray}] \textcolor{blue}{\small \hyperref[sec:supp-Multimodal_Hallucination]{Failures Related to Multimodal Hallucination}}
  \item[\faArrowCircleRight \color{gray}] \textcolor{blue}{\small \hyperref[sec:supp-judge_fails]{Failures Related to the Judge Accepting the Wrong Answer}}
  \item[\faArrowCircleRight \color{gray}] \textcolor{blue}{\small \hyperref[sec:supp-clipping_issues_and_details]{Failures Related to Clipping Issues}}
  \item[\faArrowCircleRight \color{gray}] \textcolor{blue}{\small \hyperref[sec:supp-unusual_setups]{Failures Related to Unusual Circumstances}}
\end{itemize}

\changelinkcolor{black}


\clearpage
\subsection{Failures Related to Reasoning About the Content of the Image}
\label{sec:supp-FailedReasoning}

\begin{figure*}[h]
\begin{ebox}{{InstructBLIP-7B fails to reason about the image and outputs \textit{nothing}.}}
\centering
\includegraphics[width=0.75\textwidth]{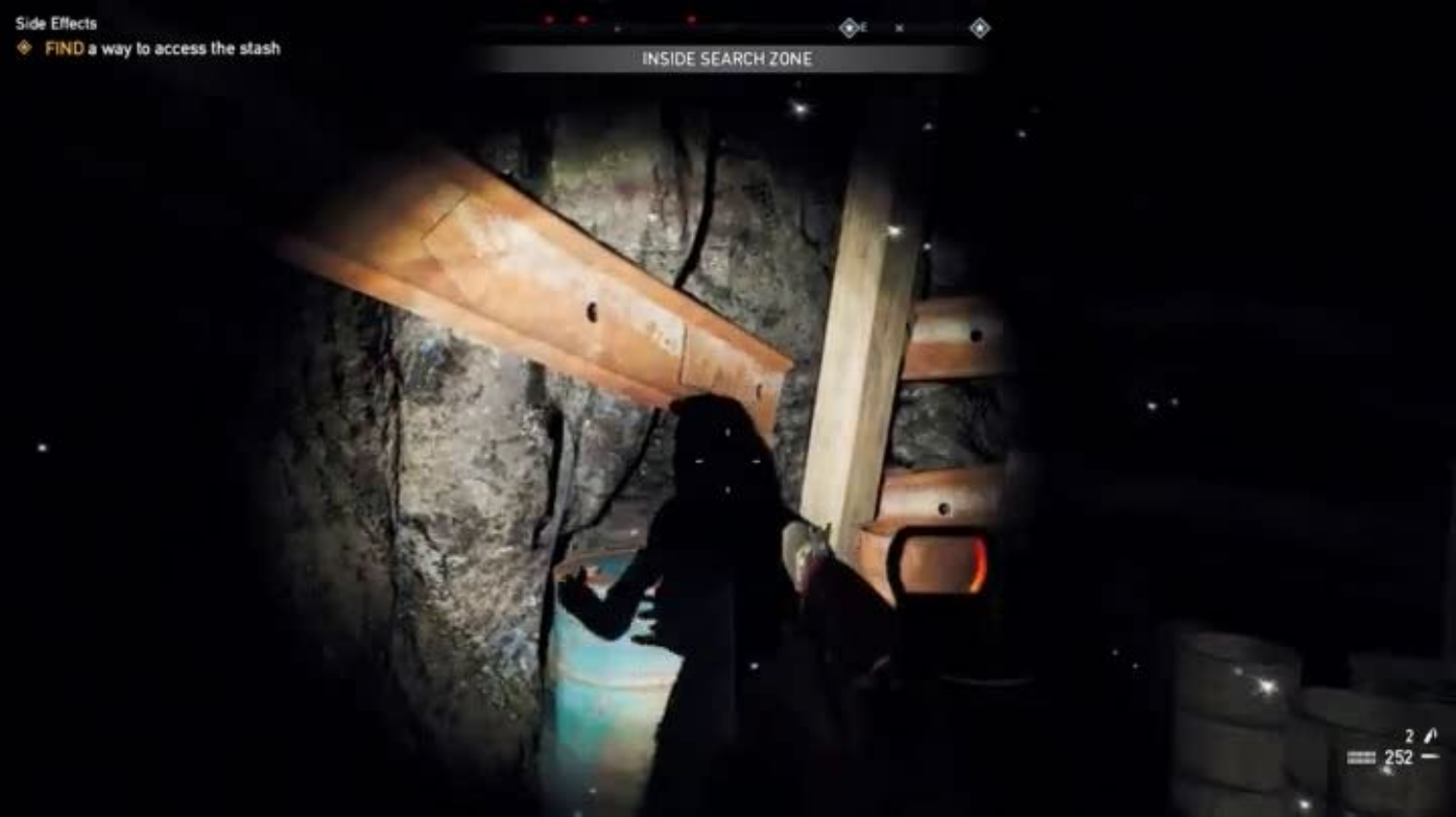}
\\
\raggedright
\textbf{\textcolor{darkblue}{Question:}} What is wrong with this image?
\\
\textbf{\textcolor{darkgreen}{Ground truth:}} A person is entirely black despite being in front of the flashlight.
\\
\textbf{\textcolor{darkred}{InstructBLIP-7B:}}
nothing
\end{ebox}

\begin{ebox}{{InstructBLIP-13B fails to reason about the image.}}
\centering
\includegraphics[width=0.75\textwidth]{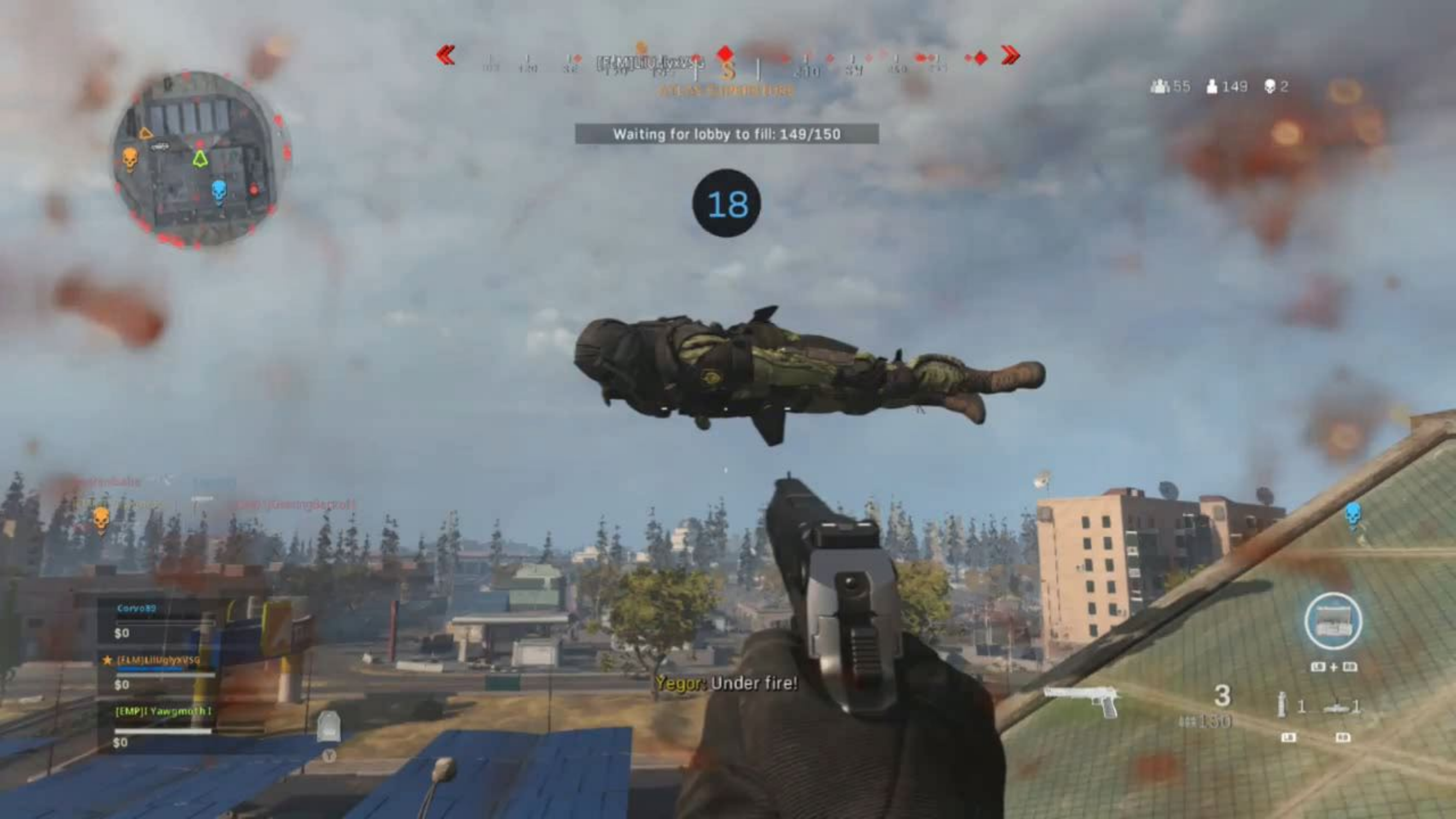}
\\
\raggedright
\textbf{\textcolor{darkblue}{Question:}} What is wrong with this image?
\\
\textbf{\textcolor{darkgreen}{Ground truth:}} A person is floating in the air in a horizontal position.
\\
\textbf{\textcolor{darkred}{InstructBLIP-13B:}}
There is nothing wrong with this image.
\end{ebox}
\caption{
Samples for which models failed to reason about the content of the image and instead responded with \textit{nothing} and similar phrases.
}
\label{suppxfig:fail_to_reason2}
\end{figure*}


\clearpage
\subsection{Failures Related to Facial Features}
\label{sec:supp-fail_to_detect_face_glitches}

\begin{figure*}[ht]
\begin{ebox}{GPT-4V fails to recognize the absence of the facial skin}
\centering
\includegraphics[width=\textwidth]{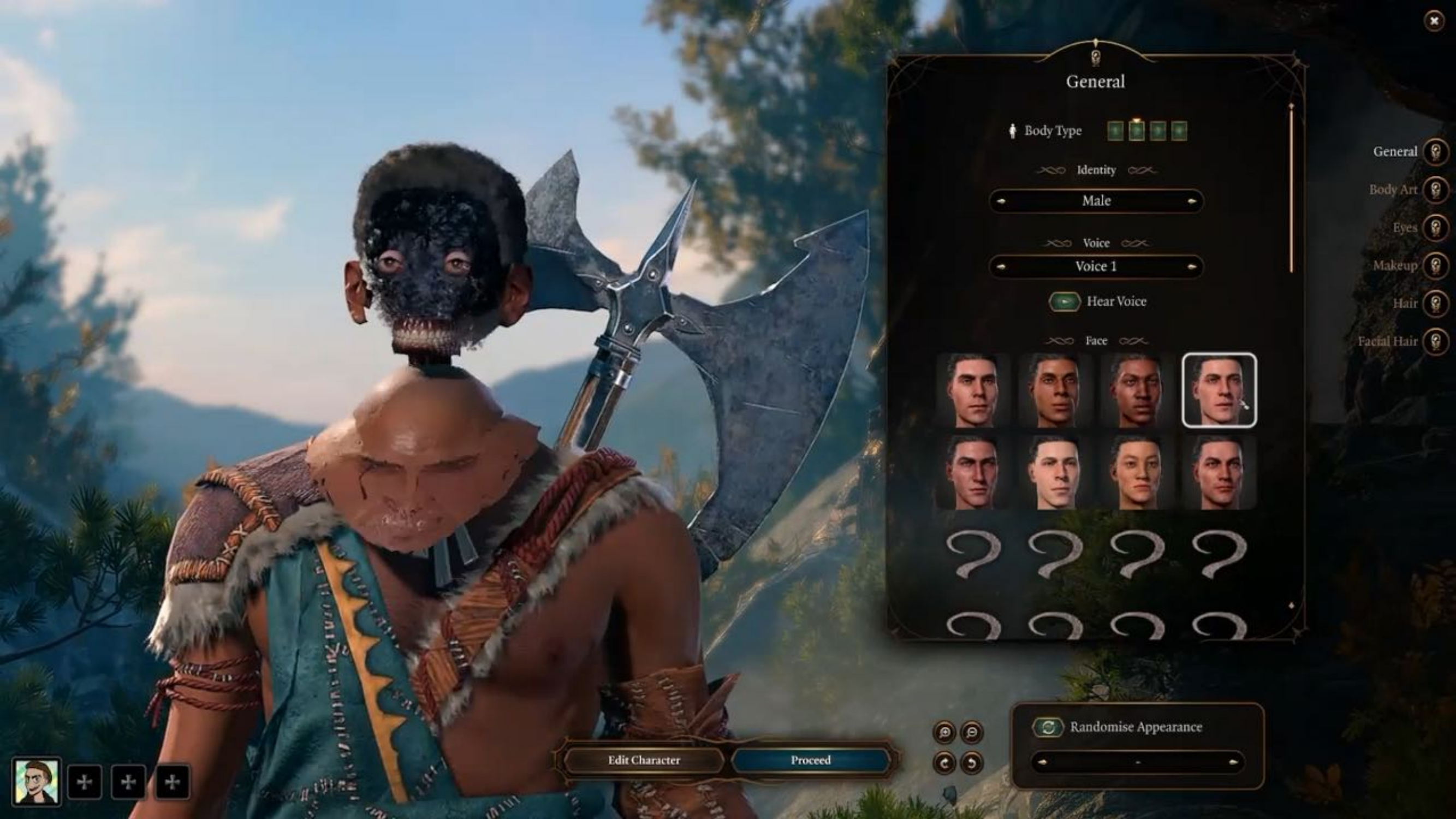}
\raggedright
\textbf{\textcolor{darkblue}{Question:}} What is wrong with this image?
\\
\textbf{\textcolor{darkgreen}{Ground truth:}} The person is missing facial skin.
\\
\textbf{\textcolor{darkred}{GPT-4V:}} 
This image appears to be from a character creation screen of a video game, where players can customize the appearance of their character. What's wrong or unusual is that the character's head has been replaced with that of a chimpanzee. This is likely a humorous or whimsical modification either due to a mod, a glitch, or a special feature within the game that allows such customization. Normally, in character customization screens, the options for head shapes and features are human, so this stands out as intentionally odd or comically absurd.
\\
\textbf{\textcolor{darkyellow}{Llama-2 (Judge):}}
No, the generated description does not accurately convey the key information and themes of the ground truth. The ground truth states that a person is missing facial skin, which is a serious and potentially disturbing condition. In contrast, the generated description describes a humorous or whimsical modification to a video game character's appearance, which is a completely different theme and tone. The generated description does not convey the same level of seriousness or medical nature as the ground truth.
\\
Feedback: The generated description could benefit from a more serious and realistic tone, as well as a focus on the medical aspects of the condition. It would be more accurate to describe the image as showing a person with a serious medical condition, rather than a humorous or whimsical modification to a video game character's appearance. Additionally, the description could benefit from more precise language, such as using terms like ``skin absence'' or ``facial skin deficiency'' to describe the condition, rather than the vague term ``missing facial skin.''
\end{ebox}
\caption{GPT-4V not only fails to recognize the absence of the facial skin, but it also hallucinates that the character's head is replaced with that of a chimpanzee.}
\label{suppfig:facelesskiss2}
\end{figure*}

\begin{figure*}[ht]
\begin{ebox}{GPT-4V fails to recognize the absence of the head.}
\centering
\includegraphics[width=0.9\textwidth]{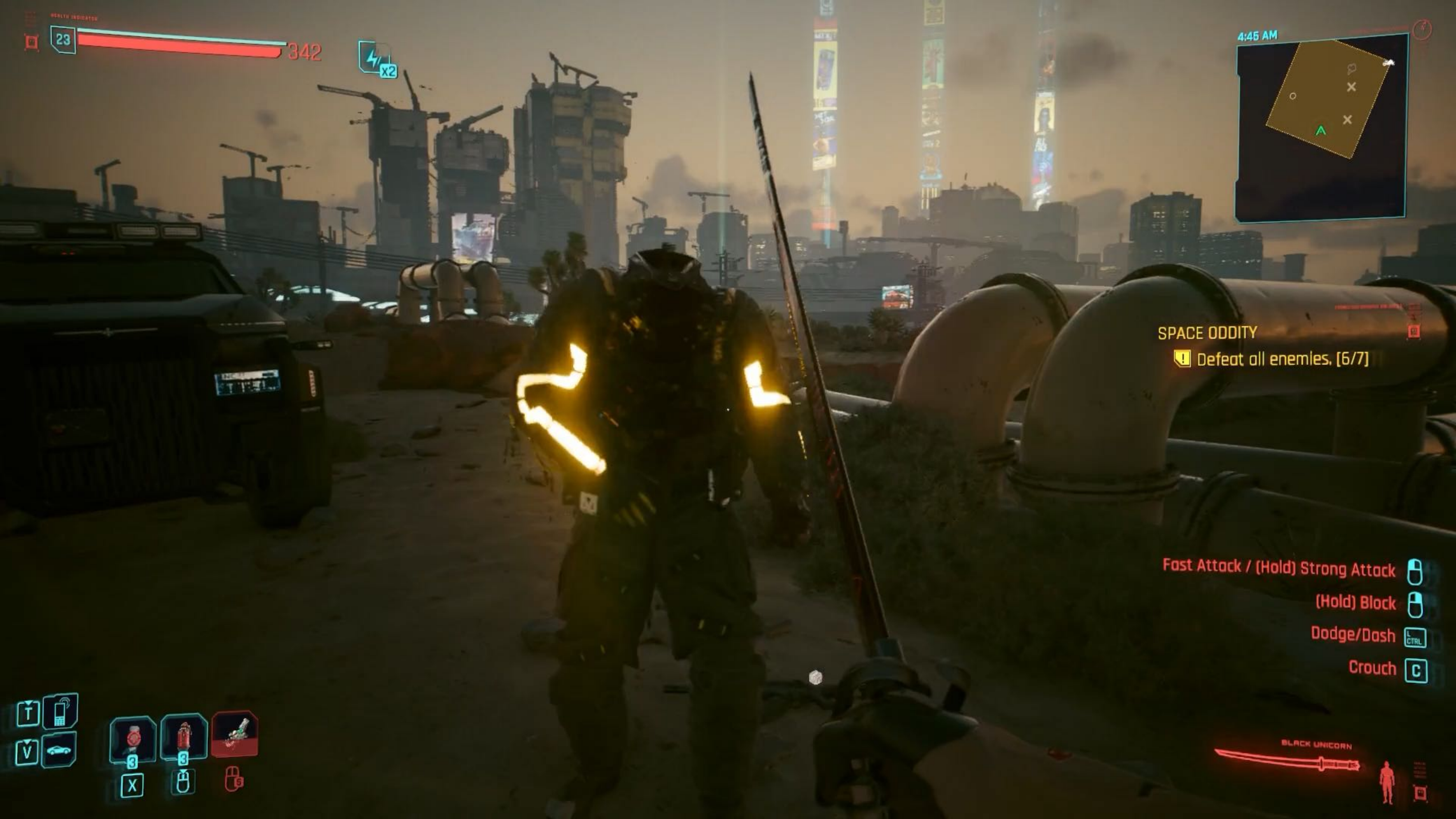}
\\
\raggedright
\textbf{\textcolor{darkblue}{Question:}} What is wrong with this image?
\\
\textbf{\textcolor{darkgreen}{Ground truth:}} The person is missing their head.
\\
\textbf{\textcolor{darkred}{GPT-4V:}} 
This image appears to be a screenshot from a video game, most likely ``Cyberpunk 2077,'' given the distinctive futuristic urban environment, user interface, and character designs that match the style of the game. There seems to be nothing inherently ``wrong'' with the image in terms of gameplay or the environment depicted, as it looks like a typical scene from within the game where the player is engaged in combat.
\\
However, if you're looking for discrepancies or errors:

\begin{enumerate}
    \item The ``Defeat all enemies'' counter shows (6/7), suggesting one enemy is left, but we can't see any enemies besides the character in the center. This might imply that the last enemy is not in the player's view or that there is a glitch where the enemy isn't appearing.
    \item The mission titled ``SPACE ODDITY'' in the top right corner suggests a specific quest or objective, and the player has a sword named ``BLACK UNICORN'' equipped, as indicated by the weapon's name at the bottom of the screen. Both are elements of the game and don't indicate anything incorrect unless they are not supposed to appear together in the context of the game, which cannot be determined from this image alone.
    
    \item The image shows a first-person perspective where the player is holding a sword. However, it could be considered unusual or suboptimal to use a melee weapon like a sword when the enemy appears to be wearing heavy armor with illuminated sections, possibly indicating a high level of threat where firearms or other strategies might be more effective.
\end{enumerate}
Unless there's a specific context or element not aligning with the game's intended design or a technical issue you're referring to, the image looks like a standard gameplay moment from ``Cyberpunk 2077.''

\textbf{\textcolor{darkyellow}{Llama-2 (Judge):}}
No, the generated description does not accurately convey the key information and themes of the ground truth.
\\
The ground truth statement ``The person is missing their head'' is a straightforward description of a disturbing and abnormal situation. The generated description, on the other hand, is a detailed analysis of a screenshot from a video game, discussing various elements of the image such as the game's environment, character designs, weaponry, and possible glitches or discrepancies. While it provides a thorough examination of the image, it completely misses the mark in terms of conveying the actual content and meaning of the ground truth. ...
\end{ebox}
\caption{GPT-4V provides a detailed description of the image, yet it fails to notice that the head is missing. The judge's response truncated to save space.}
\label{suppfig:facelesskiss1}
\end{figure*}


\clearpage
\subsection{Failures Related to Unnatural Body Positions}
\label{sec:supp-fail_to_detect_unnatrual_body_pose}

\begin{figure*}[ht]
\begin{ebox}{GPT-4V fails to detect unnaturally stretched hands.}
\centering
\includegraphics[width=\textwidth]{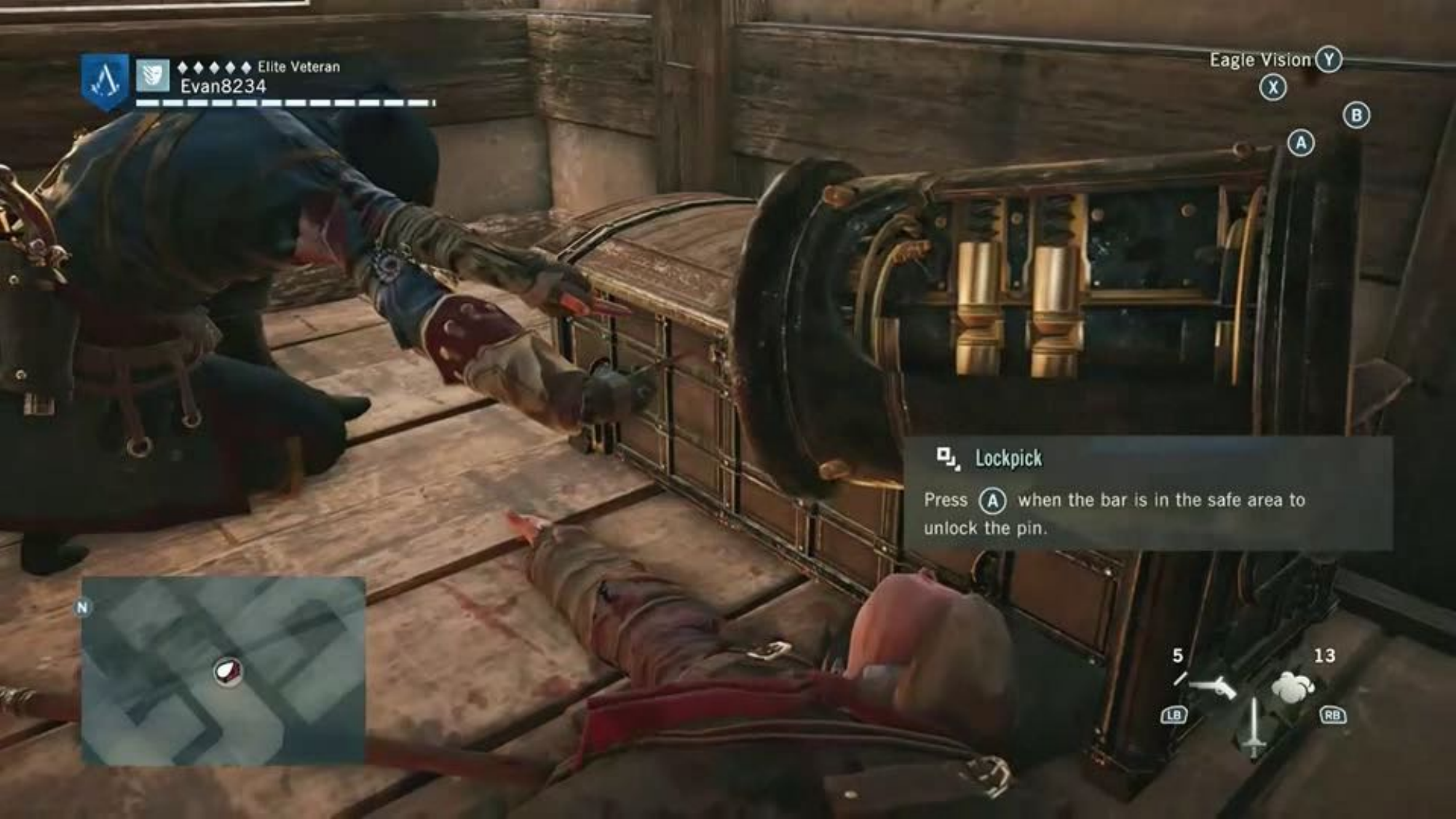}
\raggedright

\textbf{\textcolor{darkblue}{Question:}} What is unusual about this image?
\\
\textbf{\textcolor{darkgreen}{Ground truth:}} A person with unusually stretched hands.
\\
\textbf{\textcolor{darkred}{GPT-4V:}} 
The image depicts a scene from what appears to be a video game, likely from the ``Assassin's Creed'' series given the interface and character design. A character is interacting with a lock on a chest, while a downed enemy is nearby. Here are a few points that might be considered unusual, depending on context: The character is lockpicking the chest in broad daylight and seemingly in the open, which might be risky in a game focused on stealth.The nearby downed enemy suggests there was a recent confrontation, yet the character seems undisturbed while focusing on the lock.The interface shows various buttons, indicating the game is being played on a console. The button prompts suggest specific actions the player can take.The player's status, such as ``Elite Veteran'' and username ``Elvan8234'', is displayed on the screen.
\\
Without further context, it's hard to pinpoint exactly what might be unusual, as different viewers might interpret the scene in various ways based on their familiarity with the game or the specific situation presented.
\\
\textbf{\textcolor{darkyellow}{Llama-2 (Judge):}}
No, the generated description does not accurately convey the key information and themes of the ground truth.
\\
The ground truth describes a person with unusually stretched hands, which is not mentioned in the generated description. Instead, the generated description focuses on the details of a video game scene, such as the character's interaction with a lock on a chest, the nearby downed enemy, and the game's interface. While the generated description provides some context and details, it does not align with the main theme of the ground truth, which is the person's unusual physical characteristic.
\\
\end{ebox}
\caption{GPT-4V can detect some details from the image and the franchises of video games, yet it cannot recognize that the hands of the character shown in the image are unnaturally stretched to reach the box.}
\label{suppfig:unnatural_body_pose_1}
\end{figure*}

\begin{figure*}[ht]
\begin{ebox}{{LLaVA-1.5 fails to notice the leg position of the character.}}
\centering
\includegraphics[width=\textwidth]{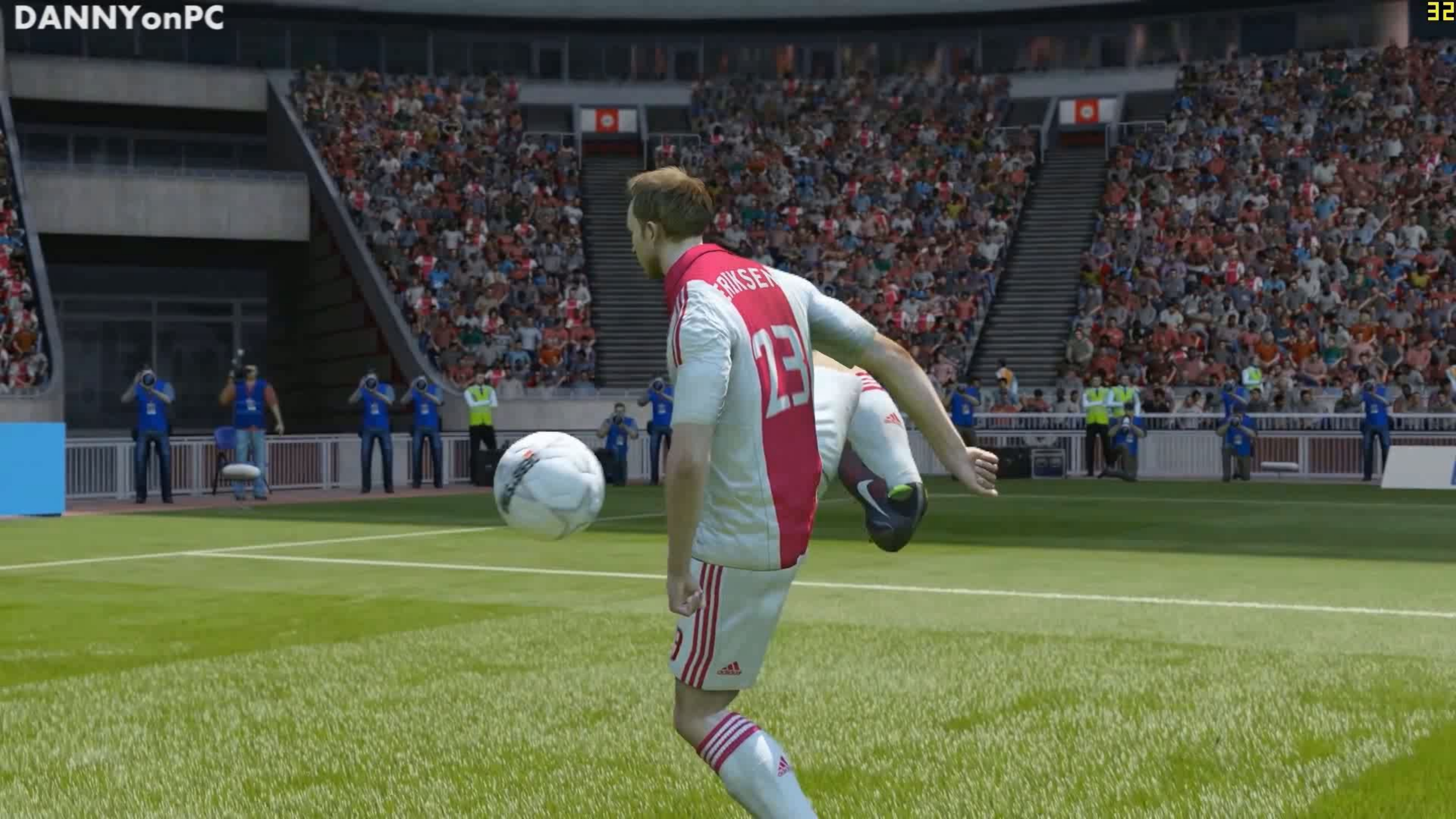}
\raggedright

\textbf{\textcolor{darkblue}{Question:}} What is unusual about this image?
\\
\textbf{\textcolor{darkgreen}{Ground truth:}} An unnatural leg position of a football player.
\\
\textbf{\textcolor{darkred}{LLaVA-1.5-13B:}} 
The unusual aspect of this image is that it is a computer-generated (CGI) image of a soccer player kicking a soccer ball. The player is wearing a red and white uniform, and the scene appears to be a soccer match. The CGI nature of the image suggests that it is not a real-life photograph but rather a digital representation of a soccer player in action. This type of image is typically used in video games, movies, or other digital media where realistic depictions of sports are required.
\\
\textbf{\textcolor{darkyellow}{Llama-2 (Judge):}}
No, the generated description does not accurately convey the key information and themes of the ground truth.
\\
The ground truth states that the image depicts an unnatural leg position of a football player, which is not explicitly mentioned in the generated description. The generated description focuses on the fact that the image is a CGI representation of a soccer player kicking a soccer ball, rather than the unusual leg position of the player.
\\
While the generated description provides some relevant information about the image, such as the player's uniform and the possibility that it may be used in digital media, it does not accurately convey the essence of the ground truth. To improve the alignment, the generated description should focus more on the unusual leg position of the player and the fact that it is an unnatural position for a football player.
\end{ebox}
\caption{LLaVA-1.5 can describe the image content and details, such as the character's clothing, but it fails to notice that the character's leg is bent in an unnatural way.}
\label{suppfig:unnatural_body_pose_2}
\end{figure*}

\begin{figure*}[ht]
\begin{ebox}{{Qwen-VL fails to notice the unnatural positions of the hands.}}
\centering
\includegraphics[width=\textwidth]{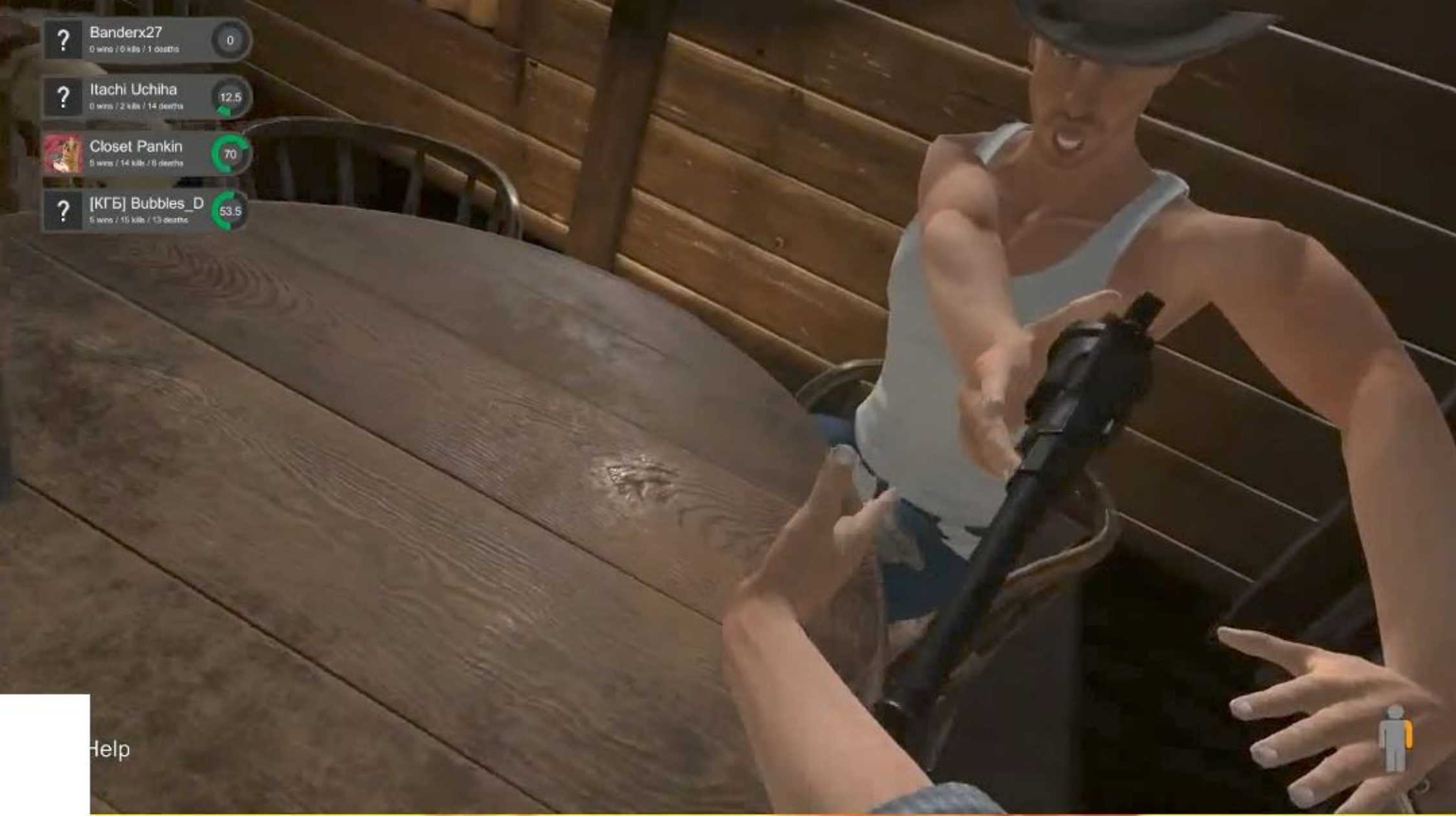}
\raggedright

\textbf{\textcolor{darkblue}{Question:}} What is unusual about this image?
\\
\textbf{\textcolor{darkgreen}{Ground truth:}} A person with unnatural hand positions.
\\
\textbf{\textcolor{darkred}{Qwen-VL:}} 
The unusual aspect of this image is that it is a screenshot taken from a video game, featuring a character pointing a gun at another character. This is not something one would typically encounter in real life.
\\
\textbf{\textcolor{darkyellow}{Llama-2 (Judge):}}
No.
\\
The generated description does not accurately convey the key information and themes of the ground truth. The ground truth focuses on the unusual hand positions of a person, while the generated description describes a scene from a video game featuring a character with a gun. The themes of violence and gaming are present in the generated description, but they are not related to the theme of unusual hand positions.
\\
Even though the generated description uses different words and phrasing, it does not faithfully represent the meaning of the ground truth. The semantic content and thematic alignment are not present, and the intent of the generated description is different from the ground truth.
\\
To improve the alignment, the generated description could focus on the unusual hand positions and exclude the elements of violence and gaming. For example, ``A person with their hands in unnatural positions, such as fingers spread widely or palms facing upwards.'' This would maintain the essence of the ground truth while allowing for creative or varied linguistic expression.
\end{ebox}
\caption{Qwen-VL can describe the image correctly, but it fails to understand the character's body configuration and the unnatural hand positions.}
\label{suppfig:unnatural_body_pose_3}
\end{figure*}


\clearpage
\subsection{Failures Related to Multimodal Hallucination}
\label{sec:supp-Multimodal_Hallucination}

\begin{figure*}[ht]
\begin{ebox}{{GPT-4V hallucinates about a person pushing a photocopier.}}
\centering
\includegraphics[width=1\textwidth]{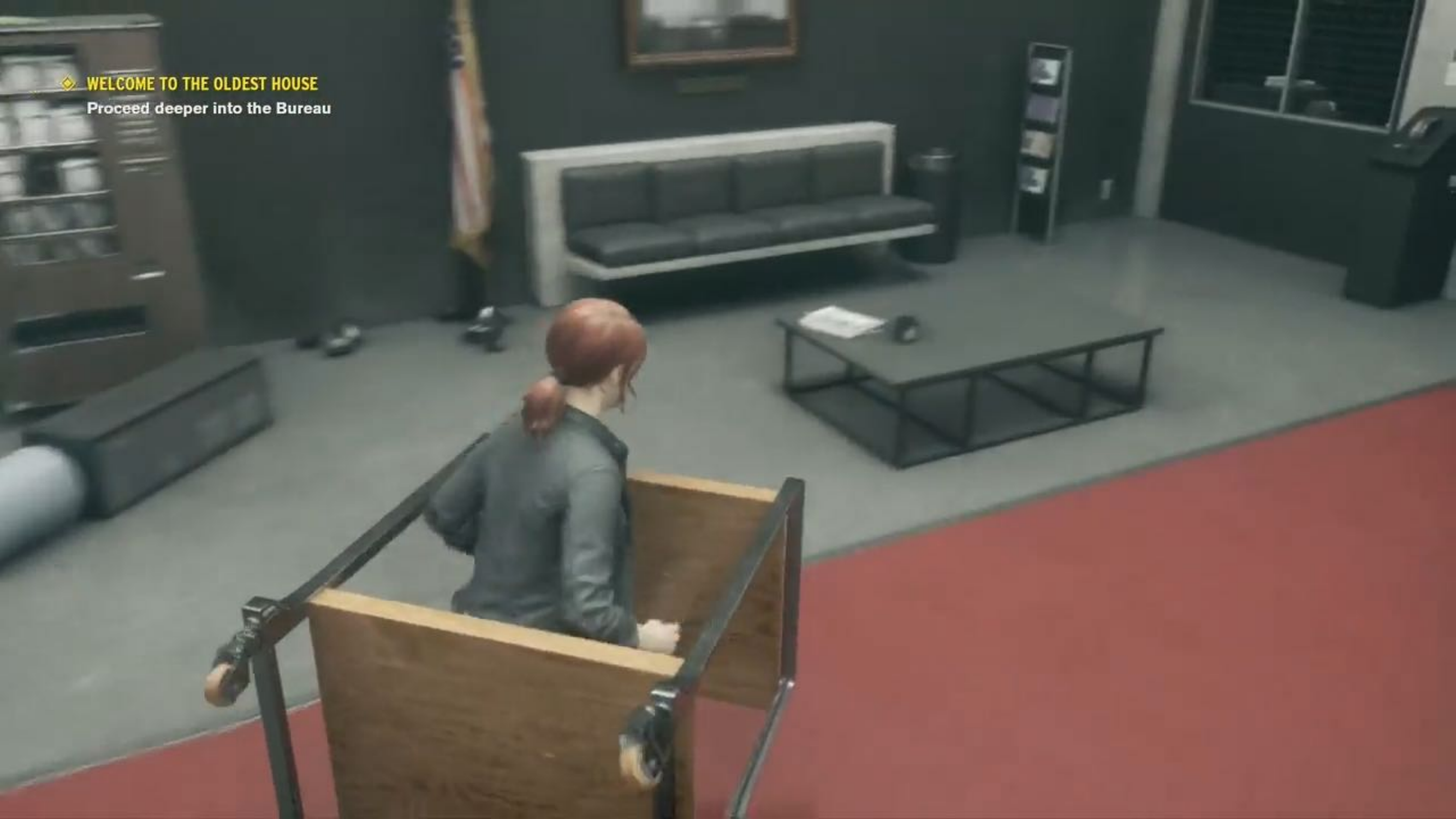}
\raggedright
\textbf{\textcolor{darkblue}{Question:}} What is unusual about this image?
\\
\textbf{\textcolor{darkgreen}{Ground truth:}} A person stuck in a piece of furniture.
\\
\textbf{\textcolor{darkred}{GPT-4V:}} 
The image appears to be from a video game, which can be inferred from the visual style and the overlaid text that reads ``WELCOME TO THE OLDEST HOUSE - Proceed deeper into the Bureau``. The unusual aspect is that it depicts a third-person view of a character pushing a photocopier inside an office environment. This is an uncommon action within the context of a video game, as photocopiers are typically static objects in such settings. The scenario suggests a gameplay element or a task within the game that involves moving office equipment, which is not a conventional video game objective.
\\
\textbf{\textcolor{darkyellow}{Llama-2 (Judge):}}
No, the generated description does not accurately convey the key information and themes of the ground truth.
\\
The ground truth describes a person stuck in a piece of furniture, which is not mentioned in the generated description. Instead, the description focuses on the video game aspect and the character's actions, which are not related to the ground truth. The themes of being stuck and struggling to get out are not present in the generated description, and the intent of the image is not accurately conveyed.
\\
While the generated description does contain some creative and varied linguistic expressions, such as ``WELCOME TO THE OLDEST HOUSE'' and ``Proceed deeper into the Bureau,'' they do not align with the meaning of the ground truth. To improve the correlation between the texts, the generated description could focus more on the person's predicament and their attempts to escape, rather than the video game context and actions.
\end{ebox}
\caption{GPT-4V starts by describing the image as a video game screenshot and then begins to read the text on the image. Regarding the actual content of the image, it fails to understand the content and hallucinates about a person pushing a photocopier.}
\label{suppfig:multimodal_hallucination_person_stuck_in_a_desk}
\end{figure*}

\begin{figure*}[ht]
\begin{ebox}{{GPT-4V hallucinates about a mirror.}}
\centering
\includegraphics[width=1\textwidth]{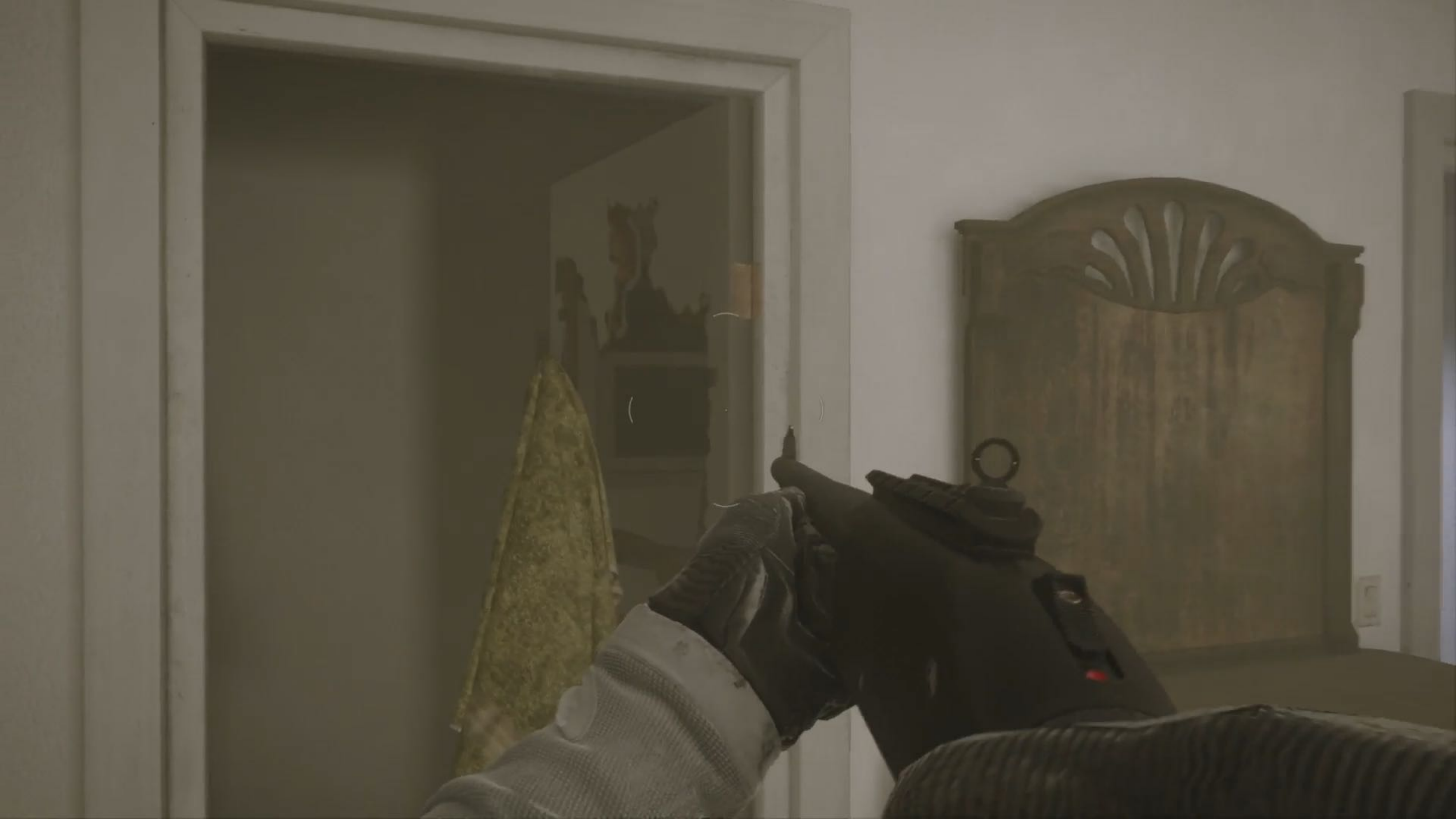}
\raggedright
\textbf{\textcolor{darkblue}{Question:}} What is unusual about this image?
\\
\textbf{\textcolor{darkgreen}{Ground truth:}} A towel is floating in the air.
\\
\textbf{\textcolor{darkred}{GPT-4V:}} 
The image appears to be a screenshot from a video game, featuring a first-person perspective where the player's hand and a gun are visible, aiming towards a mirror. What is unusual is that the mirror does not reflect the player's hand or the gun; instead, it shows a room with furniture, which suggests a graphical error or a deliberate design choice within the game. Mirrors in video games are often challenging to render correctly due to the computational demand of real-time reflections.
\\
\textbf{\textcolor{darkyellow}{Llama-2 (Judge):}}
No.
\\
The generated description does not accurately convey the key information and themes of the ground truth. The ground truth simply states that a towel is floating in the air, whereas the generated description describes a scene from a video game with a first-person perspective, a mirror, and furniture. The themes of a towel floating and a video game are unrelated, and the generated description does not capture the essence of the ground truth.
\\
While the generated description recognizes synonyms and paraphrases, such as ``a first-person perspective'' instead of ``a person holding a gun,'' it does not maintain the essence of the ground truth. The creative or varied linguistic expression in the generated description does not align with the ground truth, and the correlation between the two texts is weak.
\\
To improve the alignment, the generated description could focus on the visual elements in the image, such as the floating towel, and avoid introducing unrelated elements like a video game or furniture. A revised description that accurately conveys the key information and themes of the ground truth could be: ``A white towel is seen floating in mid-air, with no visible means of support.'' This description maintains the essence of the ground truth while using different words and phrasing.
\end{ebox}
\caption{While GPT-4V detects that the image is a screenshot of a first-person video game with a person holding a weapon, it fails to detect the floating towel; instead, it hallucinates about a mirror.}
\label{suppfig:multimodal_hallucination_towel}
\end{figure*}

\begin{figure*}[h] 
\begin{ebox}{{InstructBLIP hallucinations about a horse.}}
\centering
\includegraphics[width=0.6\textwidth]{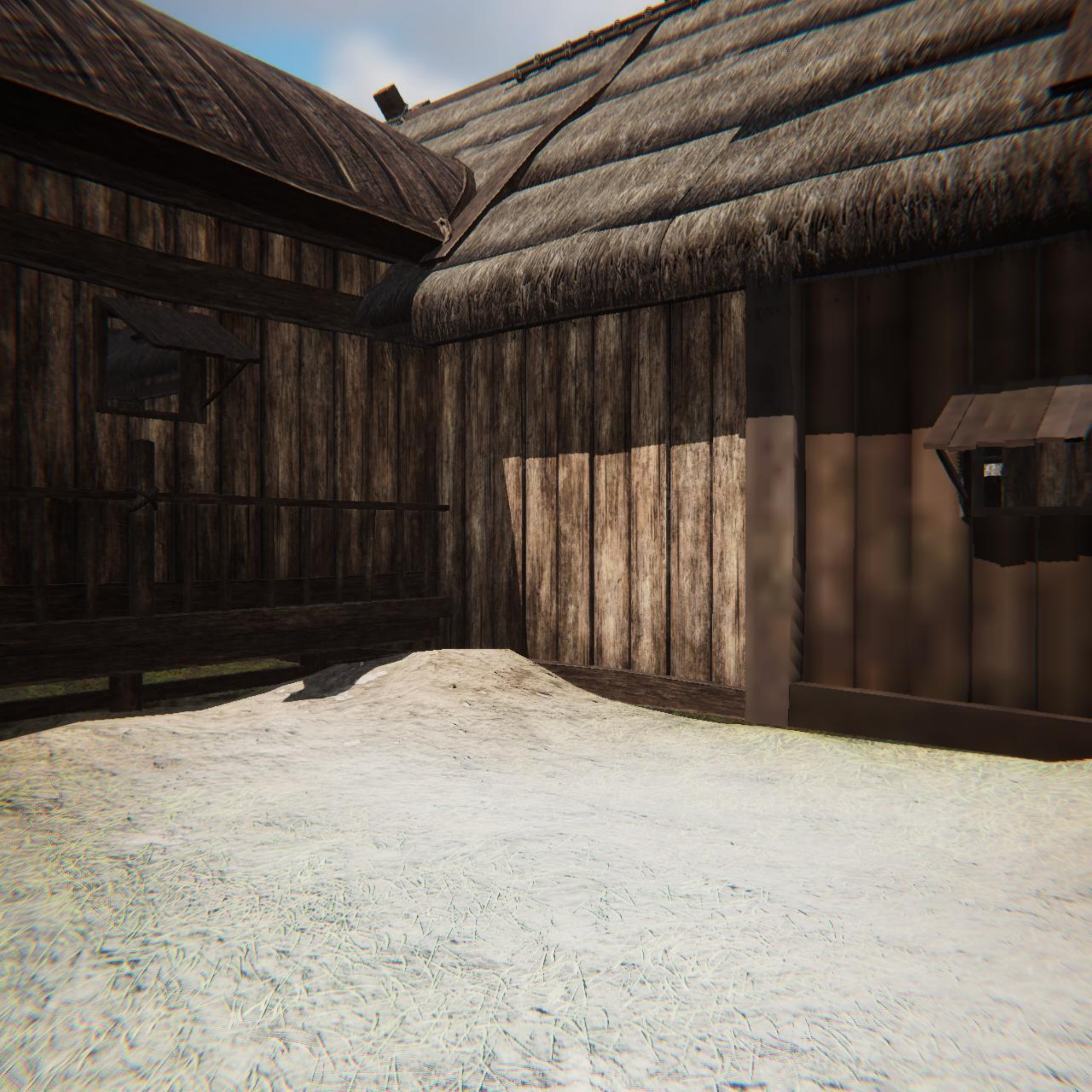}
\\
\raggedright
\textbf{\textcolor{darkblue}{Question:}} What is wrong with this image?
\\
\textbf{\textcolor{darkgreen}{Ground truth:}} Part of the house's external wall has a low-resolution texture.
\\
\textbf{\textcolor{darkred}{InstructBLIP-7B:}} 
The image is unusual because it features a horse standing in front of a wooden house with a thatched roof. Horses are typically associated with outdoor environments, such as fields or stables, rather than urban settings like the one depicted in this image. Additionally, the presence of a clock and a dog in the scene adds to the unconventional nature of the image, as these elements are not typically found in close proximity to a horse and a wooden house with a thatched roof.
\\
\textbf{\textcolor{darkyellow}{Llama-2 (Judge):}}
No, the generated description does not accurately convey the key information and themes of the ground truth. The ground truth statement simply describes a physical characteristic of a house, whereas the generated description describes a scene with a horse, clock, and dog, and focuses on the unusual nature of the image. The generated description does not mention the low-resolution texture of the house's external wall, which is the main focus of the ground truth.
\\
To improve the alignment between the generated description and the ground truth, the model could be fine-tuned to focus more on the physical characteristics of the house, rather than the objects in the scene. Additionally, the model could be trained to recognize and use more specific vocabulary related to architecture and building materials, such as ``low-resolution texture'' instead of ``unusual.''
\\
However, it's worth noting that the generated description does show some creativity and varied linguistic expression, and it could be considered a valid description of the image from a different perspective.
\end{ebox}
\caption{Sample hallucination made by InstructBLIP-7B. The model mentions the presence of a horse in front of a wooden horse.}
\label{suppxfig:multimodal_hallucination_1}
\end{figure*}

\begin{figure*}[h] 
\begin{ebox}{{LLaVA-1.5-7B hallucination about person holding a book.}}
\centering
\includegraphics[width=1\textwidth]{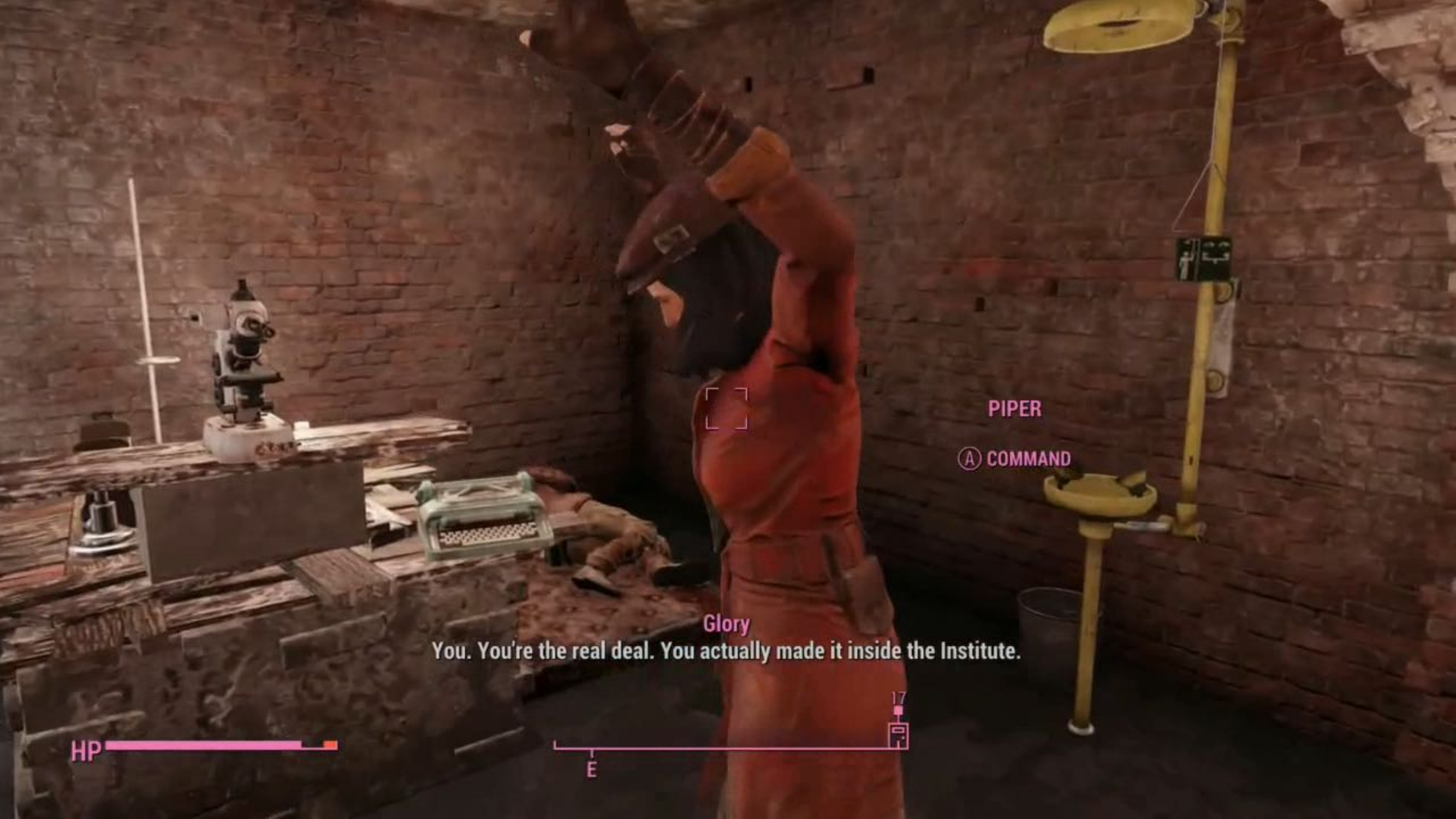}
\\
\raggedright
\textbf{\textcolor{darkblue}{Question:}} What is wrong with this image?
\\
\textbf{\textcolor{darkgreen}{Ground truth:}} A person with their hand stretched up.
\\
\textbf{\textcolor{darkred}{LLaVA-1.5-7B:}} 
The unusual aspect of this image is the presence of a person wearing a red coat and a hat, who is standing in a room with a typewriter and a person lying on the floor. The person in the red coat is holding a book, and there is a computer mouse nearby. The scene appears to be a mix of old-fashioned and modern elements, as the person is dressed in a red coat and hat, while the presence of a computer mouse and a typewriter suggests a more traditional setting. The combination of these elements creates an intriguing and unexpected visual experience.
\textbf{\textcolor{darkyellow}{Llama-2 (Judge):}}
No, the generated description does not accurately convey the key information and themes of the ground truth.
\\
The ground truth image contains a person with their hand stretched up, which is not mentioned in the generated description. The description focuses on the person's clothing, the presence of a typewriter and computer mouse, and the mixing of old-fashioned and modern elements, which are not the main aspects of the ground truth image.
\\
To improve the alignment between the generated description and the ground truth, the model should focus on the person's gesture of stretching their hand up and the overall simplicity of the image, rather than the details of the person's clothing and the surrounding objects. A possible revised description could be:
\\
``A person with their hand stretched up, surrounded by a minimalist background, creating a sense of isolation and focus.''
\\
This revised description maintains the essence of the ground truth image while using different wording and emphasizing the key elements of the image.
\end{ebox}
\caption{While LLaVA-1.5-7B provides some details about the image, such as a person lying on the ground, it hallucinates the presence of a computer mouse and a person holding a book.}
\label{suppxfig:multimodal_hallucination_2}
\end{figure*}

\begin{figure*}[h] 
\begin{ebox}{{Fuyu-8B hallucinates all the details about the image.}}
\centering
\includegraphics[width=1\textwidth]{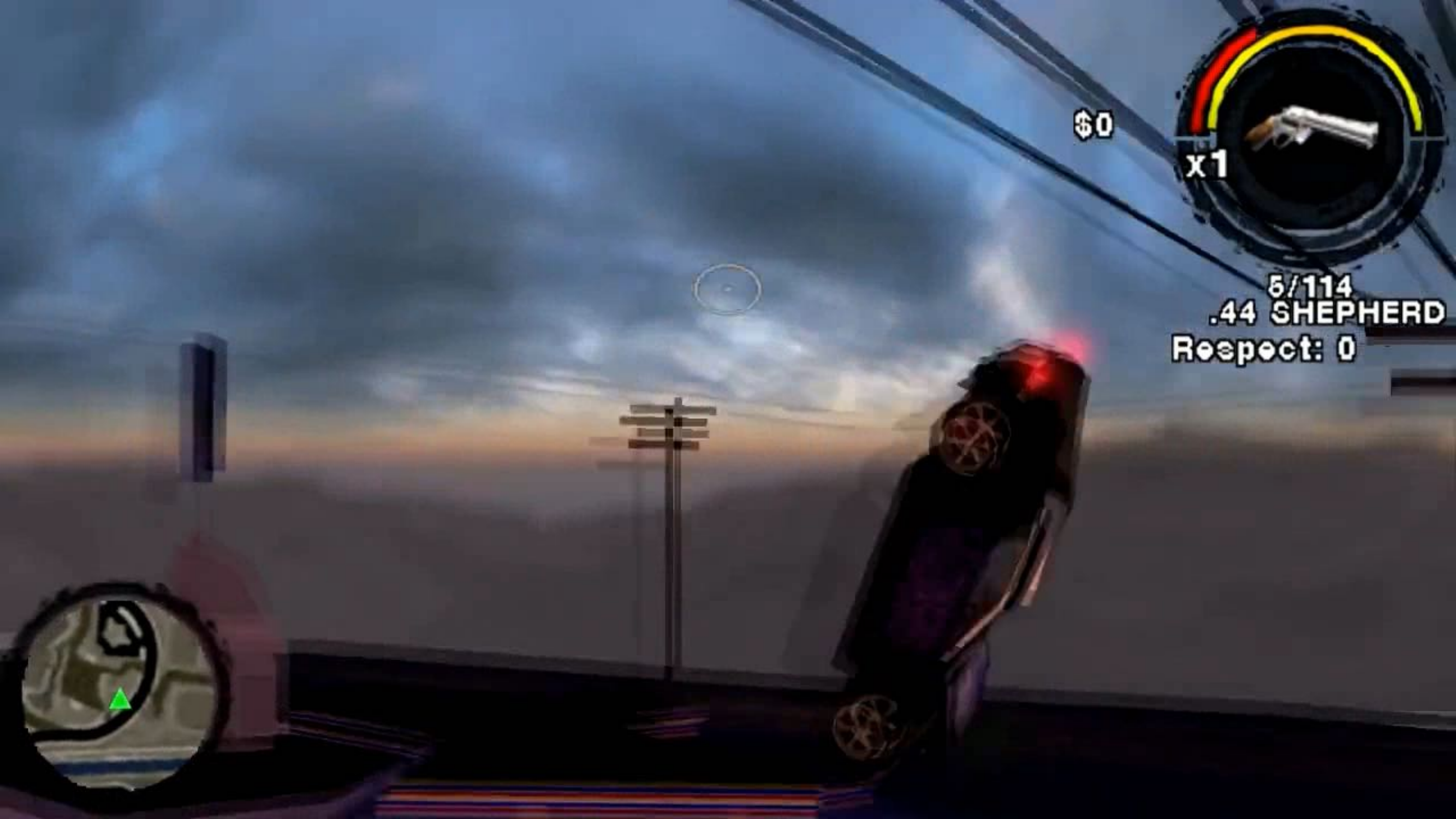}
\\
\raggedright
\textbf{\textcolor{darkblue}{Question:}} What is wrong with this image?
\\
\textbf{\textcolor{darkgreen}{Ground truth:}} A car is flying in the air.
\\
\textbf{\textcolor{darkred}{Fuyu-8B:}} 
In the image, there is a person standing on a street corner, wearing a mask and holding a weapon. The scene appears to be altered, possibly to make it appear as if the person is a zombie or a masked person. The weapon, a revolver, is visible, and the scene appears to be set in a surreal or futuristic setting. The altered nature of the image, combined with the altered perspective, creates an interesting and unusual visual effect.
\\
\textbf{\textcolor{darkyellow}{Llama-2 (Judge):}}
No, the generated description does not accurately convey the key information and themes of the ground truth. The ground truth describes a car flying in the air, while the generated description describes a person standing on a street corner with a weapon, in a possibly surreal or futuristic setting. The themes of transportation and movement are not present in the generated description, and the focus on a person and a weapon suggests a different theme altogether.
\\
While the generated description does use creative and varied linguistic expressions, such as ``altered perspective'' and ``surreal or futuristic setting,'' the overall meaning and intent of the description does not align with the ground truth. To improve the alignment, the generated description could focus more on the movement and transportation themes present in the ground truth, and strive to convey the idea of a car flying in the air in a creative and visually engaging way.
\end{ebox}
\caption{The screenshot displays a car floating in the air, but Fuyu-8B provides incorrect details and generates numerous hallucinations. All details are incorrect.}
\label{suppxfig:multimodal_hallucination_3}
\end{figure*}

\begin{figure*}[h] 
\begin{ebox}{{MiniGPT-v2 hallucinates about everything, including a red fire truck.}}
\centering
\includegraphics[width=1\textwidth]{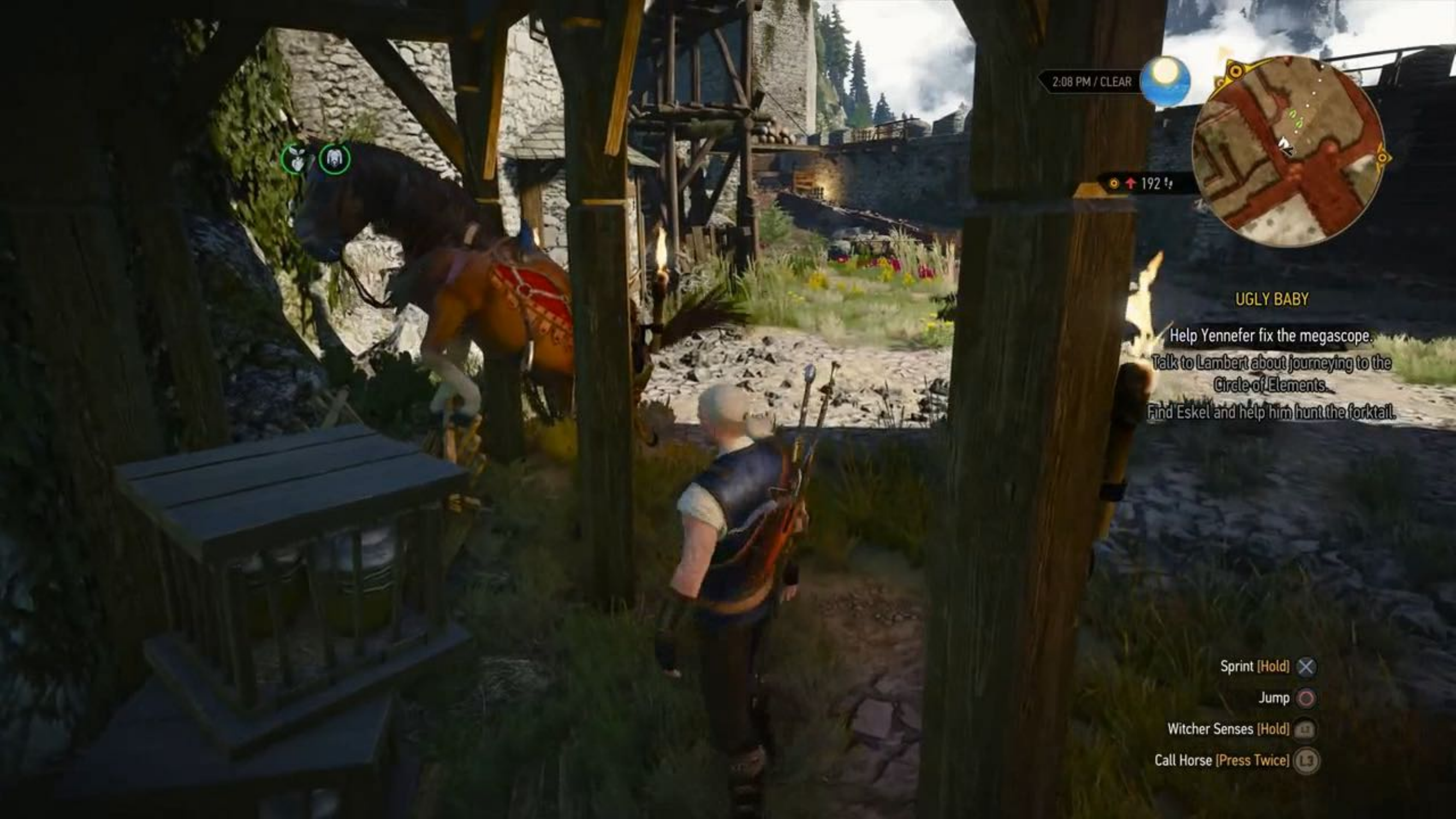}
\\
\raggedright
\textbf{\textcolor{darkblue}{Question:}} What is wrong with this image?
\\
\textbf{\textcolor{darkgreen}{Ground truth:}} A horse is in an unusual position with its front legs raised and resting on a ledge of a wooden feeding trough.
\\
\textbf{\textcolor{darkred}{MiniGPT-v2:}} 
In the image, there are two men standing near an old-looking building with wood planks. One man is pointing a bow and arrow at the other man who is holding a large, long-handled spoon that looks like it could be used for a bowling ball. There's also a wooden table nearby, but the scene appears to be quite peculiar as there are a few horses in the vicinity, as well as two dogs, one of which is sitting on the ground.
\\
There is also a large red fire truck that appears to be partially buried in the ground nearby. The combination of people and animals along with the unconventional weapons and objects in this setting may imply some sort of unusual or fictional activity, possibly a video game scenario.
\\
\textbf{\textcolor{darkyellow}{Llama-2 (Judge):}}
No, the generated description does not accurately convey the key information and themes of the ground truth. The ground truth describes a horse in an unusual position, resting its front legs on a ledge, while the generated description focuses on the presence of men, horses, dogs, and a fire truck in a setting that appears to be a video game scenario. The generated description does not mention the horse's unusual position or the wooden feeding trough, and instead introduces unrelated elements such as the bow and arrow, the large spoon, and the partially buried fire truck.
\\
The lacunarity in the generated description suggests that the model has not fully captured the essence of the ground truth, and may have been influenced by other factors such as the presence of people and animals in the image, or the suggestion of a fictional or video game-like scenario. To improve the alignment between the generated description and the ground truth, the model could benefit from additional training on images that depict unusual or unexpected scenarios, as well as reinforcement of the importance of focusing on the main subject and its unique features, rather than peripheral elements.
\\
That being said, the generated description does display some creative and varied linguistic expression, using phrases such as ``unusual or fictional activity'' and ``large, long-handled spoon that looks like it could be used for a bowling ball.'' These expressions suggest that the model has some ability to generate engaging and descriptive language, which could be leveraged to improve the accuracy and fidelity of its descriptions.
\end{ebox}
\caption{The screenshot shows a scene from The Witcher 3 game in which a horse has an unusual posture. MiniGPT-v2 provides incorrect details and generates numerous hallucinations. Almost all details are incorrect.}
\label{suppxfig:multimodal_hallucination_4}
\end{figure*}

\begin{figure*}[h] 
\begin{ebox}{{SPHINX hallucinates about ``a dog is standing on its hind legs next to a car''.}}
\centering
\includegraphics[width=1\textwidth]{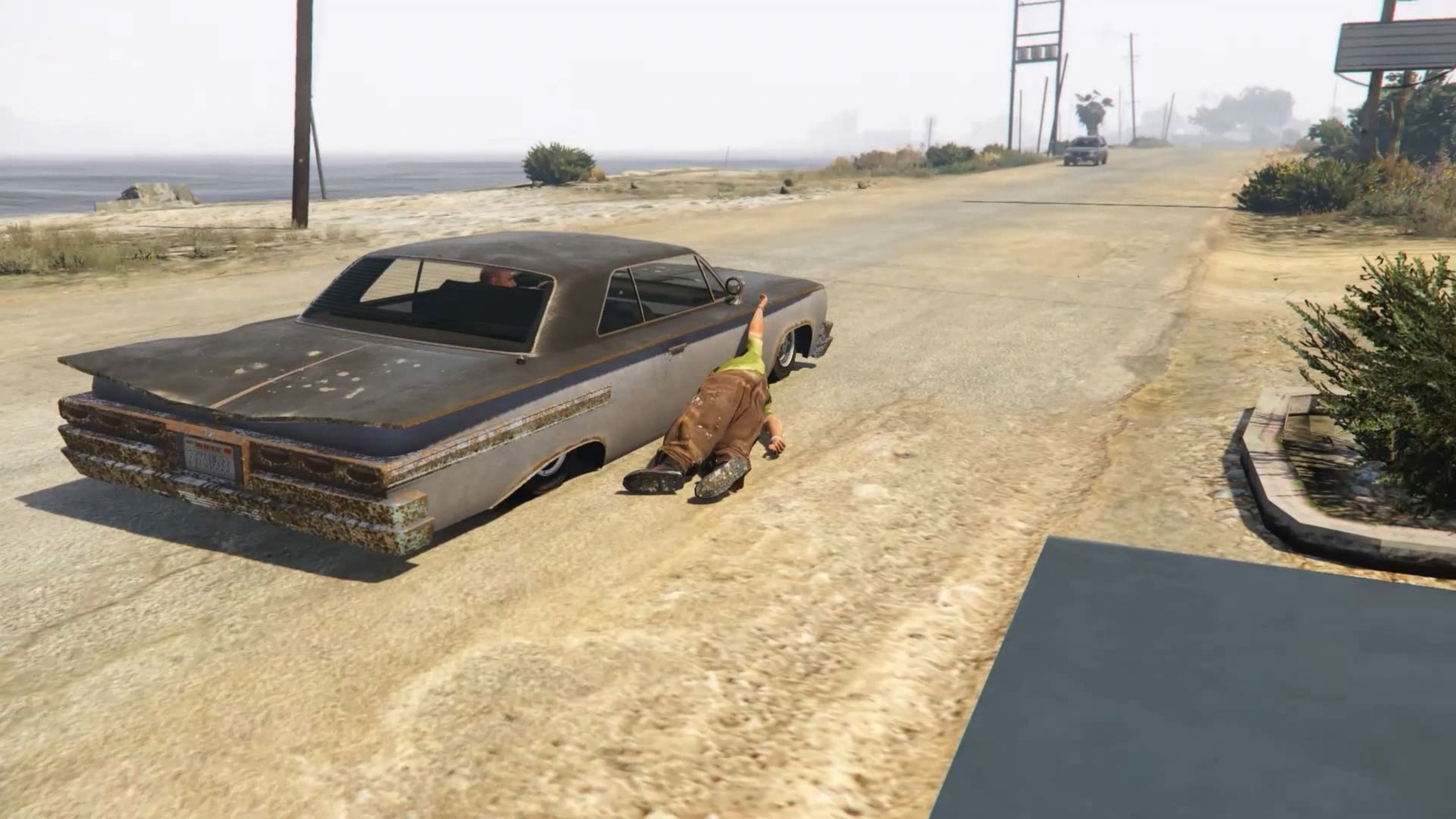}
\\
\raggedright
\textbf{\textcolor{darkblue}{Question:}} What is wrong with this image?
\\
\textbf{\textcolor{darkgreen}{Ground truth:}} A person is attached to a car's door.
\\
\textbf{\textcolor{darkred}{SPHINX:}} 
The unusual aspect of this image is that a dog is standing on its hind legs next to a car, which is not a typical behavior for a dog. This scene is likely a staged or edited photograph, as it is not common for a dog to be able to stand on its hind legs in such a manner. The dog's position and the car's location on the road create an interesting and unexpected visual effect.
\\
\textbf{\textcolor{darkyellow}{Llama-2 (Judge):}}
No, the generated description does not accurately convey the key information and themes of the ground truth.
\\
The ground truth states that a person is attached to a car's door, which suggests a potentially dangerous or emergency situation. In contrast, the generated description focuses on the unusual behavior of a dog standing on its hind legs next to a car, which is not the main focus of the ground truth. Additionally, the generated description implies that the scene is staged or edited, which is not mentioned in the ground truth.
\\
To improve the alignment between the generated description and the ground truth, the model should focus more on the person's attachment to the car's door and the potential danger or urgency of the situation. The description could be rephrased to emphasize the unexpected nature of the person's attachment and the car's location, while still acknowledging the dog's unusual behavior. For example: ``An unexpected scene unfolds as a person is seen attached to the door of a car, while a dog stands on its hind legs nearby. The person's attachment to the car creates a sense of urgency, and the dog's unusual behavior adds to the overall unexpected nature of the scene.''
\end{ebox}
\caption{Sample image in which SPHINX hallucinates about a dog. However, some details about the image, such as the presence of a car, are correct.}
\label{suppxfig:multimodal_hallucination_5}
\end{figure*}


\clearpage
\subsection{Failures Related to the Judge Accepting the Wrong Answer}
\label{sec:supp-judge_fails}

\begin{figure*}[h] 
\begin{ebox}{{Llama-2 aligns ``desert-like environment'' with the  ``pixelated character floating in the air''.}}
\centering
\includegraphics[width=0.85\textwidth]{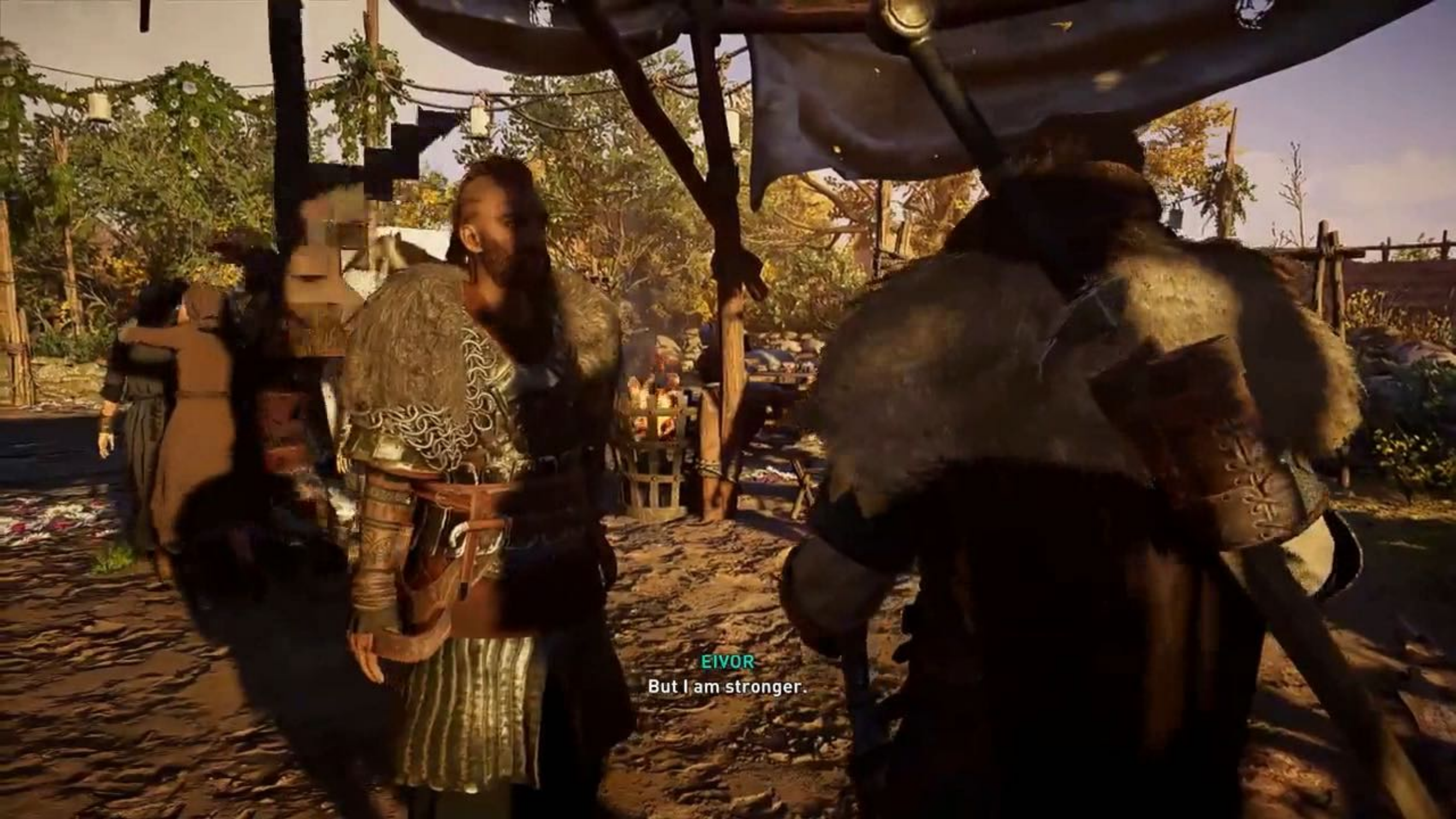}
\\
\raggedright
\textbf{\textcolor{darkblue}{Question:}} What is wrong with this image?
\\
\textbf{\textcolor{darkgreen}{Ground truth:}} A pixelated character is floating in the air.
\\
\textbf{\textcolor{darkred}{LLaVA-1.5-13B:}} 
In the image, there is a man standing in front of a tent, wearing a long coat and holding a sword. The man appears to be a character from a video game, possibly a Viking or a warrior. The scene seems to be set in a desert-like environment.
\\
There are a few minor issues in the image. The man's sword is positioned in a way that it appears to be floating in the air, rather than being held properly. Additionally, there is a small glitch in the image where the man's head appears to be cut off, making it difficult to see his facial features. These issues suggest that the image might be a screenshot from a video game, and the game's developers might need to address these bugs and glitches to improve the overall gaming experience.
\\
\textbf{\textcolor{darkyellow}{Llama-2 (Judge):}}
Yes.
\\
The generated description successfully conveys the key information and themes of the ground truth, despite some differences in wording and phrasing. The reference to a ``pixelated character'' in the ground truth is interpreted by the model as a ``man standing in front of a tent'' who appears to be a video game character, possibly a Viking or a warrior. The mention of a ``desert-like environment'' aligns with the pixelated character floating in the air.
\\
The model's description also creatively includes details not present in the ground truth, such as the character's long coat and the sword being held in a way that appears to be floating. These additional elements maintain the essence of the ground truth while offering a unique expression.
\\
However, there are minor issues in the image that the model has identified, such as the glitch in the man's head and the sword appearing to be floating. These observations suggest that the image might be a screenshot from a video game, and the game's developers might need to address these bugs and glitches.
\\
Overall, the generated description effectively conveys the key information and themes of the ground truth and offers some creative variations while maintaining the essence of the original meaning. The model's output demonstrates a good understanding of the semantic content and thematic alignment, and its feedback on the minor issues in the image is appreciated.
\end{ebox}
\caption{Llama-2 is confused and accepts the wrong response. The response generated by LLaVA-1.5-13B hallucinates a floating sword, which is incorrect. However, Llama-2 matches ``desert-like environment'' with the ground truth: ``A pixelated character is floating in the air.''}
\label{suppxfig:judge_fails_1}
\end{figure*}

\begin{figure*}[h] 
\begin{ebox}{{Llama-2 aligns ``unique and exciting match-up'' with ``clipping into each other''.}}
\centering
\includegraphics[width=0.95\textwidth]{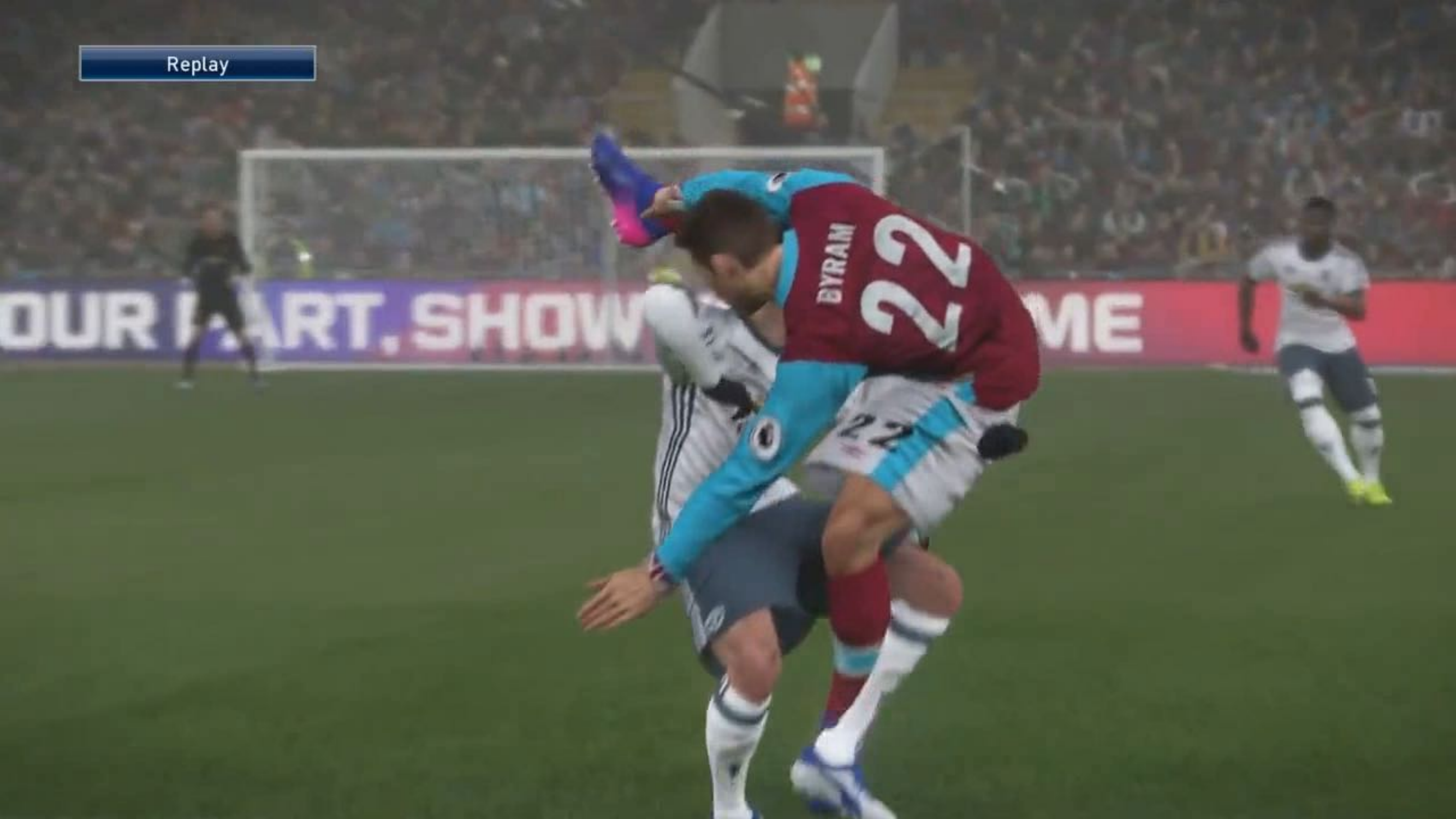}
\\
\raggedright
\textbf{\textcolor{darkblue}{Question:}} What is wrong with this image?
\\
\textbf{\textcolor{darkgreen}{Ground truth:}} Two characters are clipping into each other.
\\
\textbf{\textcolor{darkred}{MiniGPT-v2:}} 
In this image, two soccer players are engaged in a unique and exciting match-up. The unusual aspect is that one of the players is tackling another player while wearing skates, which is not typically seen in soccer games, as these involve only foot movements rather than using any wheeled devices. This uncommon scenario adds to the intensity of the action captured in the photo and provides an interesting and captivating view for spectators.
\\
\textbf{\textcolor{darkyellow}{Llama-2 (Judge):}}
Yes, the generated description conveys the key information and themes of the ground truth. While the ground truth simply states that two characters are clipping into each other, the generated description provides additional context and detail, describing the unusual aspect of the scene, the intensity of the action, and the captivating view for spectators.
\\
The generated description uses different wording and phrasing than the ground truth, but it maintains the essence of the original message. The use of synonyms such as ``unique and exciting match-up'' for ``clipping into each other'' and ``uncommon scenario'' for ``two characters'' is effective in conveying the same meaning. Additionally, the description provides more context and detail, such as the fact that one of the players is wearing skates, which adds to the intensity of the action.
\\
Overall, the generated description is a good example of a faithful representation of the ground truth's meaning, despite using different wording and phrasing. It offers a creative and varied linguistic expression that maintains the essence of the original message, making it a good fit for the task at hand.
\end{ebox}
\caption{MiniGPT-v2 produces a partial description of the image with some hallucination, leading Llama-2 to accept the wrong answer as a correct match with the ground truth.}
\label{suppxfig:judge_fails_2}
\end{figure*}

\begin{figure*}[h] 
\begin{ebox}{Llama-2 incorrectly focuses on potential themes related to the ground truth and ignores glitches.}
\centering
\includegraphics[width=0.95\textwidth]{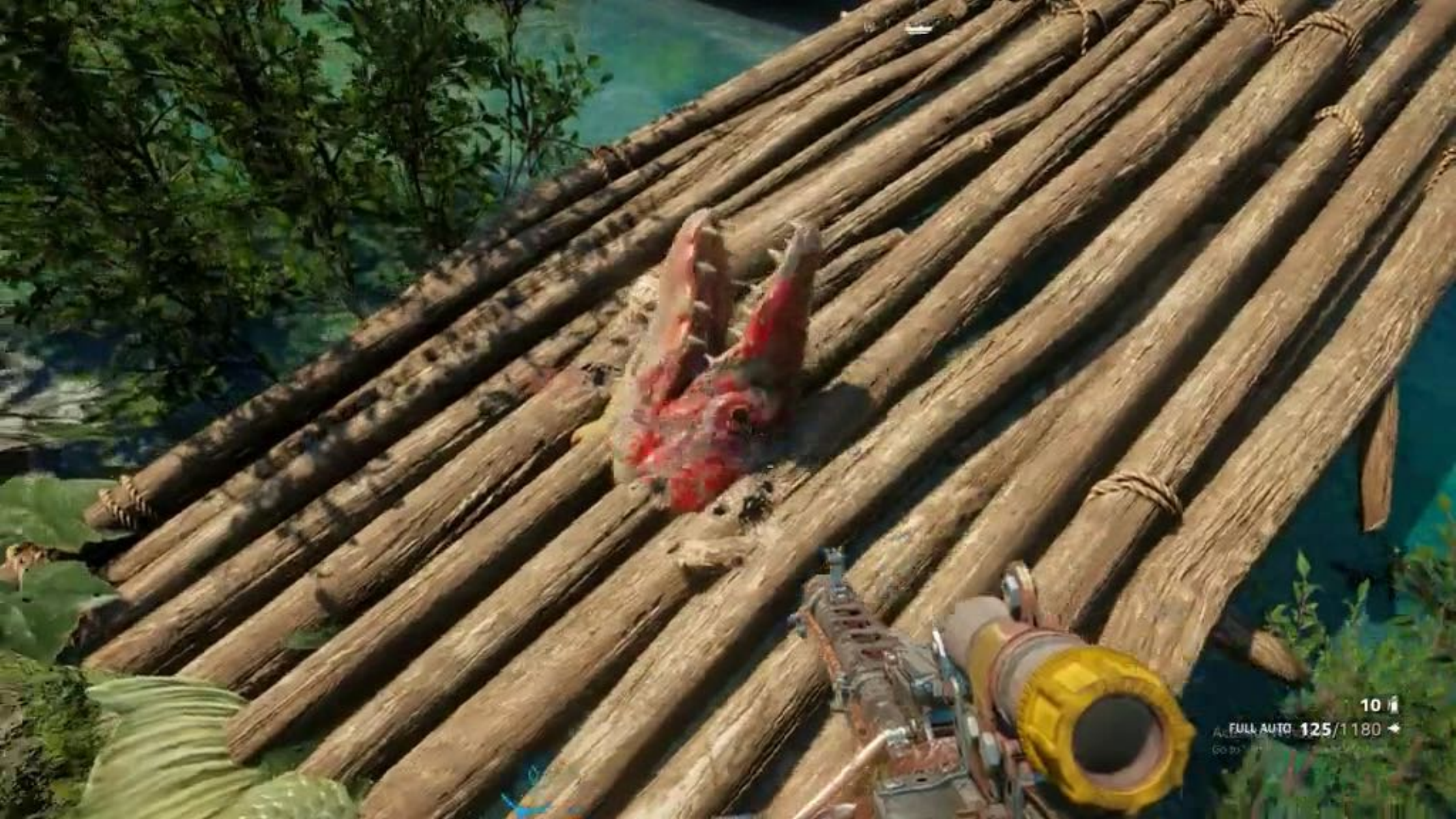}
\\
\raggedright
\textbf{\textcolor{darkblue}{Question:}} What is unusual about this image?
\\
\textbf{\textcolor{darkgreen}{Ground truth:}} A creature is clipping inside a wooden pathway, with its head visible.
\\
\textbf{\textcolor{darkred}{InstructBLIP-13B:}} 
In this image, there is an unusual presence of a crocodile on a wooden bridge. The crocodile appears to be swimming or floating in the water underneath the bridge, which is made of wooden planks. This is an unusual sight as crocodiles are not typically found in this type of environment and are more commonly associated with swamps, rivers, and other aquatic habitats. The presence of the crocodile in this image adds an element of danger and unpredictability to the scene, suggesting that the player needs to be cautious while navigating the area.
\\
\textbf{\textcolor{darkyellow}{Llama-2 (Judge):}}
Yes, the generated description conveys the key information and themes of the ground truth, even if worded differently. The semantic content is similar, with the focus on an unusual creature (crocodile) in an unexpected environment (wooden bridge). The thematic alignment is also present, as both texts convey a sense of danger and unpredictability. The generated description also maintains the intent of the ground truth, suggesting that the player needs to be cautious while navigating the area.
\\
While the generated description uses different phrasing and wording, it recognizes synonyms (unusual presence, swimming, floating) and paraphrases (not typically found, more commonly associated) that faithfully represent the ground truth's meaning. Additionally, the generated description offers creative expressions (unusual sight, element of danger) that maintain the essence of the ground truth.
\\
However, to improve the correlation between the texts, the generated description could better align with the ground truth by specifically mentioning the creature as a crocodile, rather than using a more generic term like ``unusual presence.'' This would enhance the fidelity of the description while still allowing for creative expressions.
\\
In conclusion, the generated description effectively conveys the key information and themes of the ground truth, with minor opportunities for improvement in alignment.
\end{ebox}
\caption{While the InstructBLIP's response contains some key elements such as crocodiles and wooden planks, it does not mention the clipping issue. InstructBLIP also hallucinates about crocodiles swimming in the water. However, Llama-2, acting as the judge, aligns with the theme of the ground truth in the InstructBLIP's response and accepts the answer.}
\label{suppxfig:judge_fails_3}
\end{figure*}

\begin{figure*}[h] 
\begin{ebox}{{Llama-2 does not focus on the meaning of the generated response.}}
\centering
\includegraphics[width=0.95\textwidth]{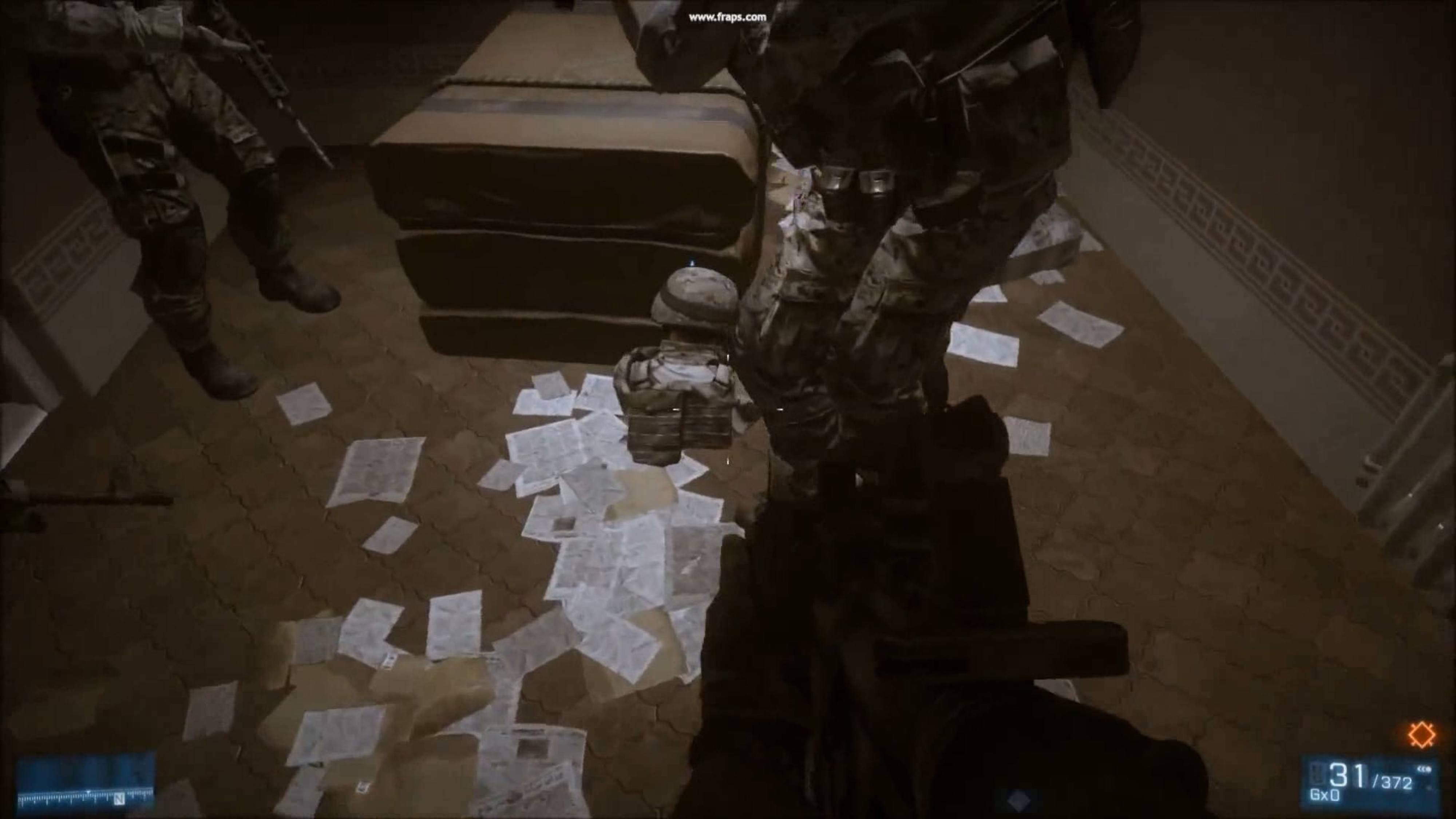}
\\
\raggedright
\textbf{\textcolor{darkblue}{Question:}} What is unusual about this image?
\\
\textbf{\textcolor{darkgreen}{Ground truth:}} A person is clipped through the ground and only upper body and head is visible
\\
\textbf{\textcolor{darkred}{OtterHD:}} 
In this image, a soldier is standing on top of a bed and pointing a gun at a fellow soldier who is kneeling on the ground. This scene is unusual as it is not typical for people to engage in such actions, especially within a domestic setting like a bedroom. Additionally, the presence of a gun in the scene indicates that there might be a conflict or a threat, which is also out of the ordinary in a bed room. The combination of these elements makes the image quite unusual and potentially concerning.
\\
\textbf{\textcolor{darkyellow}{Llama-2 (Judge):}}
Yes, the generated description accurately conveys the key information and themes of the ground truth.
\\
The ground truth references a person who is partially visible above the ground, with only their upper body and head visible. The generated description does not explicitly mention this, but it does describe a person (the soldier) who is standing on top of a bed, which suggests that they are elevated above the ground.
\\
Furthermore, the generated description accurately identifies the unusual nature of the scene, including the presence of a gun and the conflict or threat that it implies. This aligns with the thematic focus of the ground truth, which highlights the unexpected and potentially concerning nature of the image.
\\
While the generated description does not use the same phrasing or wording as the ground truth, it successfully conveys the essential information and themes of the image. Therefore, the answer is Yes, the generated description accurately represents the ground truth.
\end{ebox}
\caption{OtterHD thinks that the presence of soldiers and guns in the bedroom is an unusual aspect of the image. However, it fails to notice the clipping issue and mistakes it for a soldier kneeling on the ground. Llama-2, as the judge, acknowledges that the generated text and the ground truth have different wording, yet it completely fails to evaluate the meaning of the response and incorrectly accepts it.}
\label{suppxfig:judge_fails_4}
\end{figure*}

\begin{figure*}[h] 
\begin{ebox}{{Llama-2 matches the inaccurate description of the scene with the ground truth.}}
\centering
\includegraphics[width=0.95\textwidth]{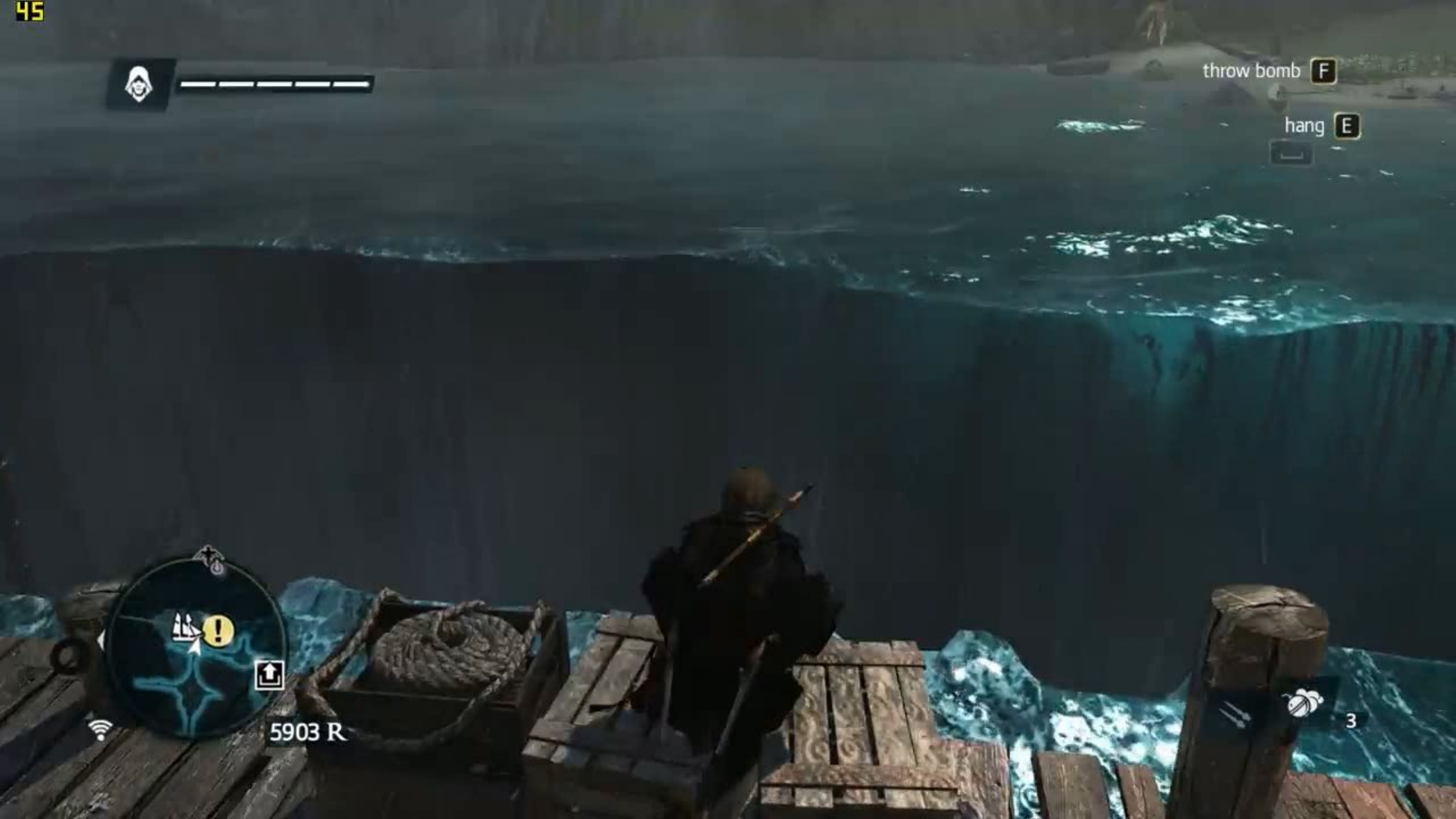}
\\
\raggedright
\textbf{\textcolor{darkblue}{Question:}} What is wrong with this image?
\\
\textbf{\textcolor{darkgreen}{Ground truth:}} The water surface is not rendered correctly and seems to have a rectangular hole in it.
\\
\textbf{\textcolor{darkred}{SPHINX:}} 
The image is a computer game scene, featuring a man sitting on a dock overlooking a large body of water. The man is holding a fishing pole, and there is a boat visible in the water. The problem with this image is that it is not a real-life scene, but rather a computer-generated image from a video game. The man, the boat, and the surrounding environment are all part of the game's virtual world, and they do not exist in reality.
\\
\textbf{\textcolor{darkyellow}{Llama-2 (Judge):}}
Yes, the generated description accurately conveys the key information and theme of the ground truth reference.
\\
Although the generated description differs in wording and phrasing from the ground truth, it successfully conveys the message that the image is a computer-generated scene from a video game and not a real-life scene. It highlights the same problem as the ground truth, i.e., the incorrect rendering of the water surface, but instead of focusing on the rectangular hole, it describes the scene as a whole, emphasizing its virtual nature.
\\
The generated description provides additional information that is not present in the ground truth, such as the presence of a man holding a fishing pole and a boat in the water. However, this additional information does not detract from the main theme and message of the ground truth, which is the incorrect rendering of the water surface.
\\
Therefore, the generated description is a valid paraphrase of the ground truth, maintaining the essence of the original message while offering different phrasing and stylistic expressions. It successfully conveys the intended meaning of the ground truth, albeit with some creative liberties taken to enhance the description.
\end{ebox}
\caption{Despite the fact that SPHINX's response does not mention any problem with the water, Llama-2 matches SPHINX's response with the ground truth and accepts it.}
\label{suppxfig:judge_fails_5}
\end{figure*}


\clearpage
\subsection{Failures Related to Clipping Issues}
\label{sec:supp-clipping_issues_and_details}

\begin{figure*}[h] 
\begin{ebox}{{GPT-4V fails to detect two cars of the same model clipping on top of each other.}}
\centering
\includegraphics[width=1\textwidth]{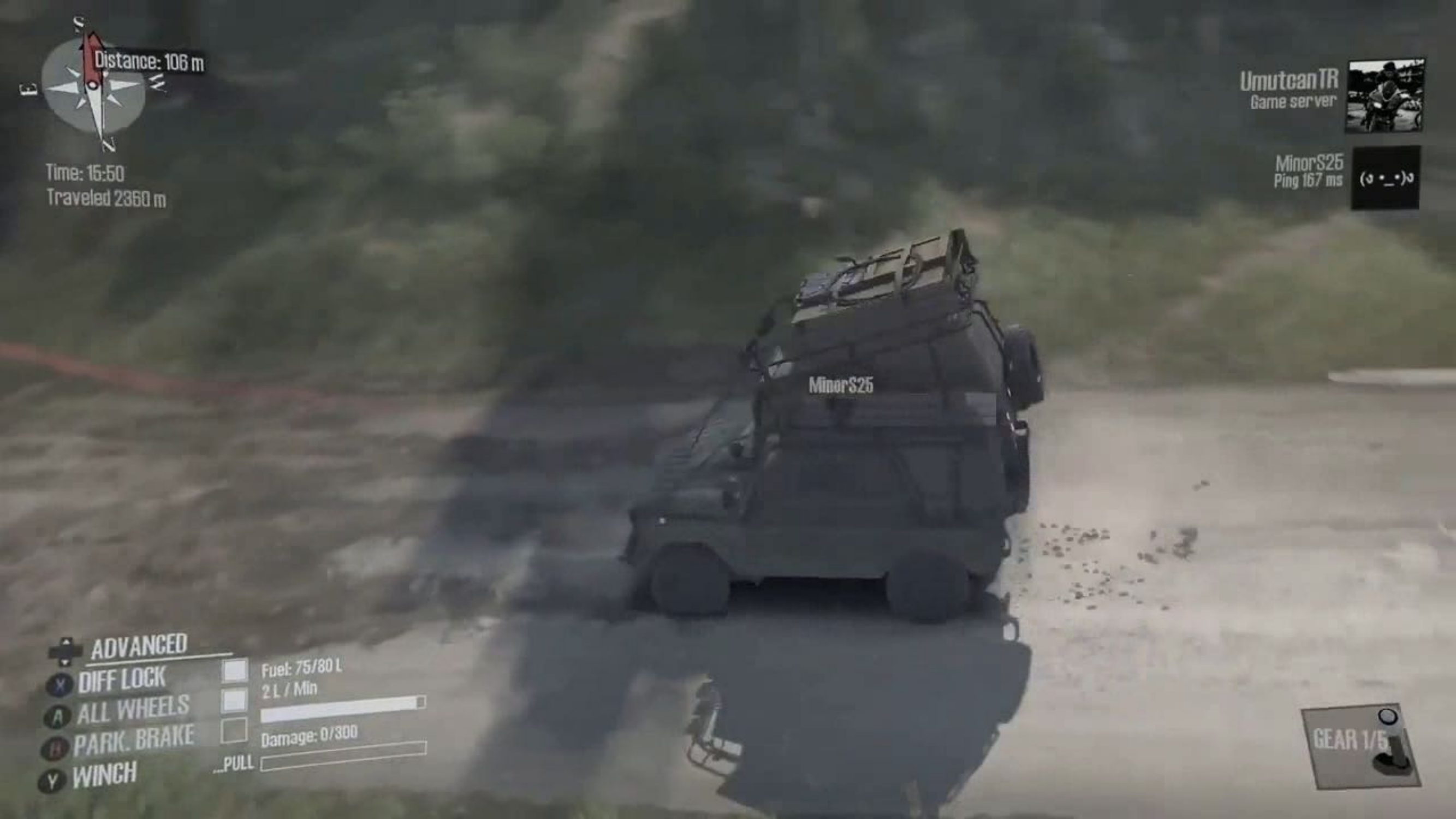}
\\
\raggedright
\textbf{\textcolor{darkblue}{Question:}} What is unusual about this image?
\\
\textbf{\textcolor{darkgreen}{Ground truth:}} Two cars are intersecting and clipping through each other.
\\
\textbf{\textcolor{darkred}{GPT-4V:}} 
The image appears to be a screenshot from a video game, possibly a driving or off-road simulation game, given the various HUD (heads-up display) elements visible, such as the ``Distance,'' ``Time,'' ``Traveled,'' and vehicle status indicators like ``Fuel,'' ``Diff Lock,'' ``All Wheels,'' and ``Winch.''
\\
If we are looking for something unusual: The vehicle seems to be in a rugged environment, but the terrain doesn't show much wear or tracks behind the vehicle, even though it seems to be kicking up dirt or dust.The ``Gear 1/5'' indicator at the bottom right suggests the vehicle is in the first gear, but considering the distance it has traveled (2360 m), it's a bit unusual for it to still be in first gear, unless the terrain is very challenging.
\\
There's a ``MinorS25'' tag on the vehicle, which may be a player's username, but without context, it's hard to say if this is unusual or not.
\\
It's important to note that ``unusual'' is subjective and could vary depending on the context or the specific game mechanics. If you had something specific in mind, please provide more context or clarify your question.
\\
\textbf{\textcolor{darkyellow}{Llama-2 (Judge):}}
No, the generated description does not accurately convey the key information and themes of the ground truth. The ground truth states that two cars are intersecting and clipping through each other, but the generated description does not mention this at all. Instead, it focuses on the details of the vehicles and the environment, such as the HUD elements, the terrain, and the vehicle's gear.
\\
While the generated description does contain some creative and varied linguistic expressions, such as ``kicking up dirt or dust'' and ``rugged environment,'' it does not faithfully represent the meaning of the ground truth. ...
\end{ebox}
\caption{GPT-4V fails to detect a clipping glitch in which two cars of the same model and visually identical are placed on top of each other on a dirt road. The judge's response is truncated to save space.}
\label{suppxfig:clipping_issues_1}
\end{figure*}

\begin{figure*}[ht]
\begin{ebox}{{GPT-4V fails to detect clipping glitch between character and the door.}}
\centering
\includegraphics[width=1\textwidth]{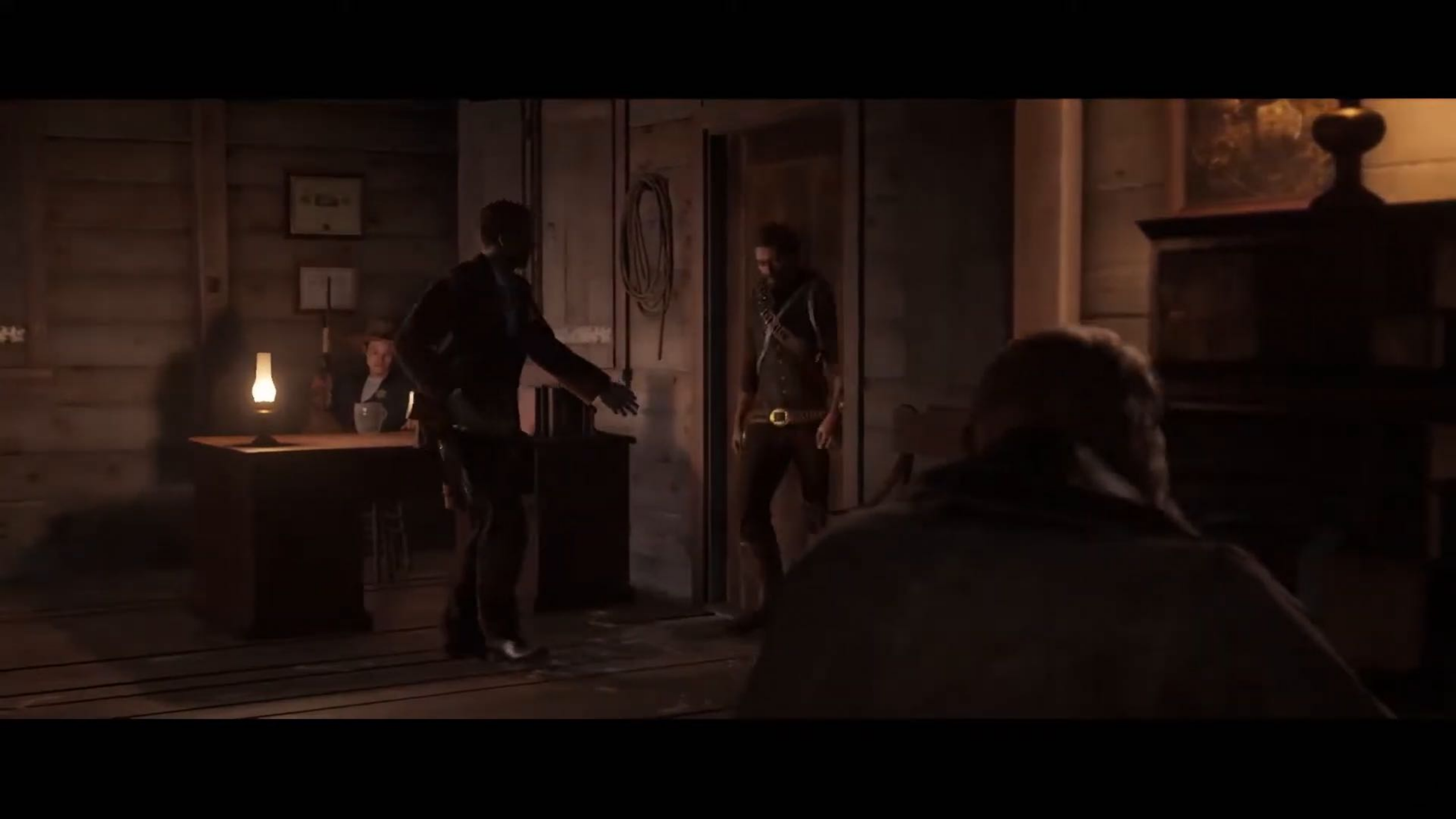}
\raggedright
\textbf{\textcolor{darkblue}{Question:}} What is wrong with this image?
\\
\textbf{\textcolor{darkgreen}{Ground truth:}} A man is clipping and intersecting with a closed door.
\\
\textbf{\textcolor{darkred}{GPT-4V:}} 
The image appears to be a screenshot from a video game that resembles the style and characters of ``Red Dead Redemption 2,'' given the Western theme and period-appropriate clothing and setting. There are three characters shown; two are standing, possibly in a standoff, and one is seated, observing the situation.

As for what might be ``wrong'' with the image, that could depend on context. If you're referring to inconsistencies or anachronisms in the setting or character appearance, nothing immediately stands out as obviously incorrect without more context. However, if this is a critique of the scene's composition or a glitch in the game graphics, additional details would be needed to make a proper assessment.

If you are inquiring about a technical flaw or something that seems out of place within the game's context, it's not evident without further information or a more specific indication of what to look for. If there is a specific aspect you would like to discuss or point out, please provide more details.
\\
\textbf{\textcolor{darkyellow}{Llama-2 (Judge):}}
The generated description does not accurately convey the key information and themes of the ground truth. The ground truth describes a man clipping and intersecting with a closed door, whereas the generated description discusses a screenshot from a video game, possibly from ``Red Dead Redemption 2,'' with three characters shown in a Western setting.
\\
The generated description does not mention anything about a man, a door, or any intersection or clipping. Instead, it focuses on the visual elements of the image, such as the characters, setting, and possible anachronisms. The description also mentions the composition of the scene and the possibility of a glitch in the game graphics, which is not related to the ground truth.
\\
Therefore, the answer is No, the generated description does not faithfully represent the meaning of the ground truth.
\\
To improve the alignment between the generated description and the ground truth, the model could focus more on the physical interactions and objects depicted in the image, rather than the setting and characters. The model could also benefit from additional context or information about the image, such as the purpose of the man's actions or the significance of the closed door.
\end{ebox}
\caption{The screenshot shows a glitch in which a character is standing in a doorway frame while the door is closed, resulting in clipping with the door. GPT-4V fails to notice that the door is closed.}
\label{suppxfig:clipping_issues_2}
\end{figure*}

\begin{figure*}[h] 
\begin{ebox}{{LLaVA-1.5 fails to understand the clipping issue despite detecting objects in the image.}}
\centering
\includegraphics[width=1\textwidth]{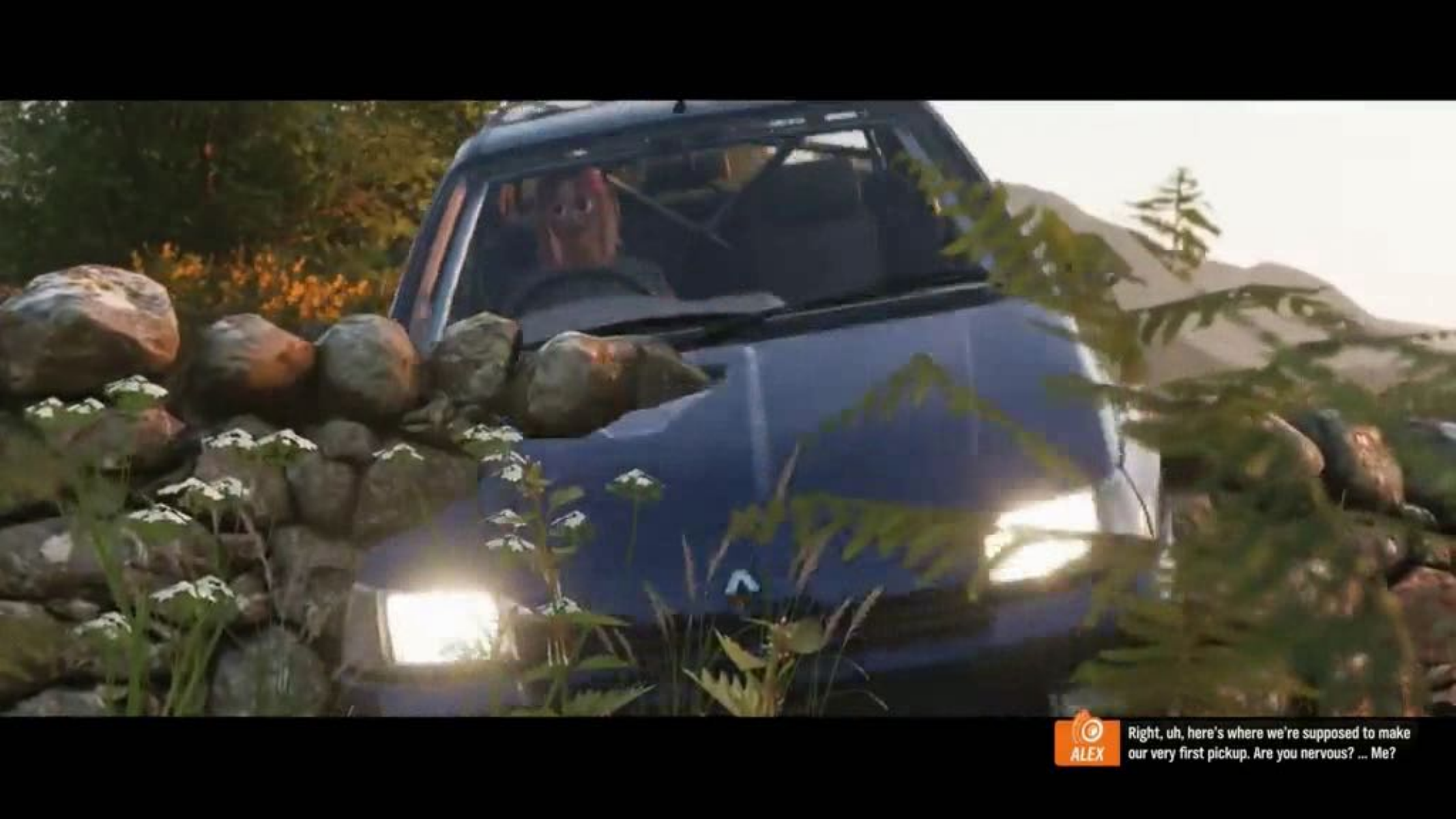}
\\
\raggedright
\textbf{\textcolor{darkblue}{Question:}} What is wrong with this image?
\\
\textbf{\textcolor{darkgreen}{Ground truth:}} A car is clipped inside a stone barrier.
\\
\textbf{\textcolor{darkred}{LLaVA-1.5-13B:}} 
The image is a computer-generated or digitally altered photo of a car driving through a forest. The car is surrounded by rocks and bushes, and there is a person visible in the car. However, the person appears to be a cartoon or computer-generated character, which is not realistic. This suggests that the image has been manipulated or created using digital tools, resulting in a fake or unrealistic scene.
\\
\textbf{\textcolor{darkyellow}{Llama-2 (Judge):}}
No, the generated description does not accurately convey the key information and themes of the ground truth. The ground truth states that a car is clipped inside a stone barrier, indicating a dangerous or precarious situation. In contrast, the generated description focuses on the fact that the image is computer-generated or digitally altered, and that the person in the car is a cartoon or computer-generated character. This shift in focus and the use of unrealistic elements detract from the original meaning of the ground truth.
\\
To improve the alignment between the generated description and the ground truth, the model could focus more on the physical environment and the potential danger posed by the stone barrier, rather than the digital origin of the image or the unrealistic elements present in the scene. For example, the generated description could be revised to say: ``A car is precariously situated inside a stone barrier, surrounded by rocky terrain and dense vegetation, conveying a sense of isolation and potential danger.'' This revised description better captures the key information and themes of the ground truth while still allowing for some creative expression and variation in wording.
\end{ebox}
\caption{LLaVA-1.5 detects some individual objects from the image, for example, the car and its driver, rocks and bushes, but it does not understand the clipping issue happening between the car and rocks (or stone barrier according to the ground truth).}
\label{suppxfig:clipping_issues_3}
\end{figure*}

\begin{figure*}[h] 
\begin{ebox}{{MiniGPT-v2 detects the presence of two people on the bed, however, it fails to recognize the clipping issue.}}
\centering
\includegraphics[width=1\textwidth]{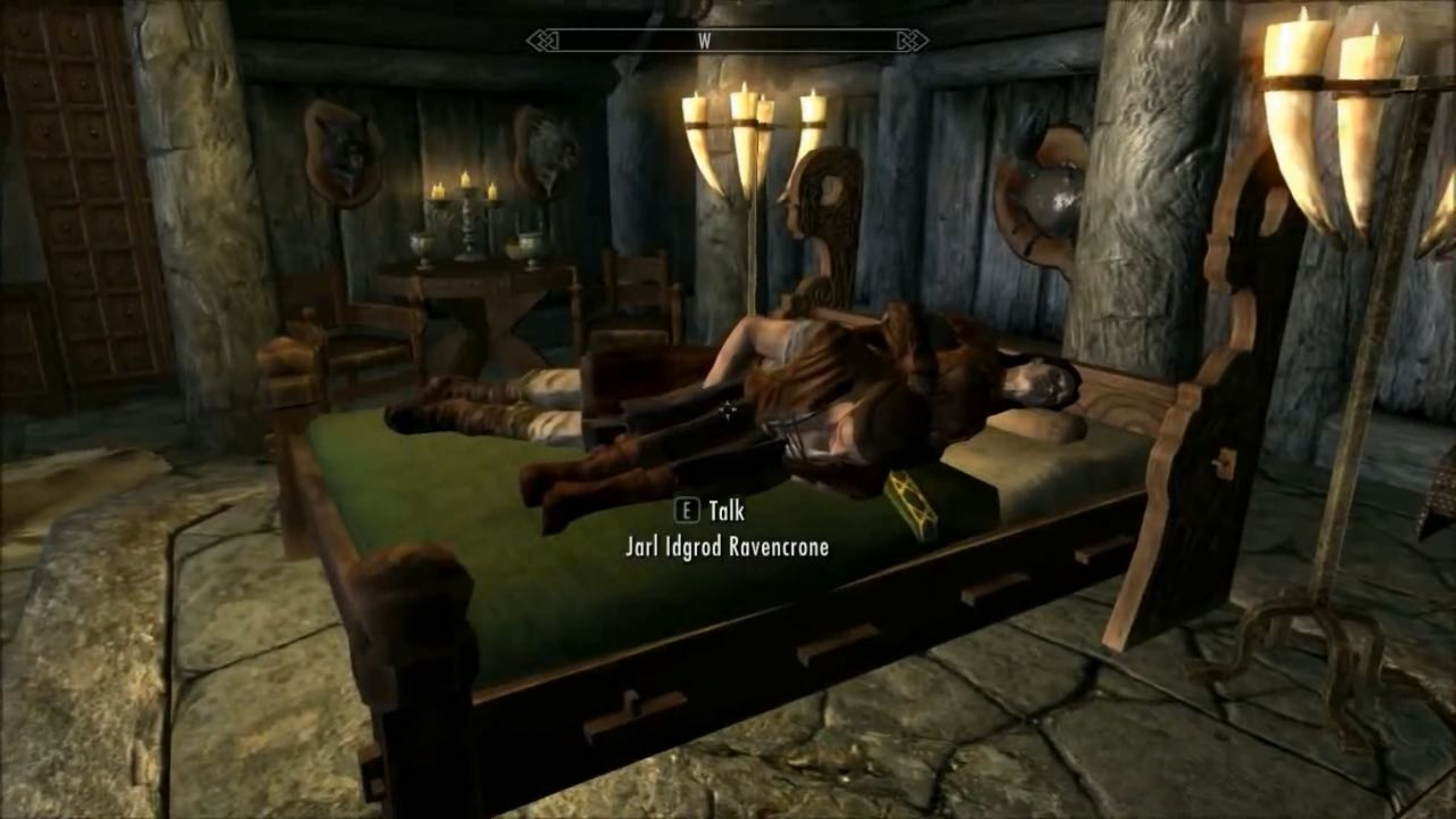}
\\
\raggedright
\textbf{\textcolor{darkblue}{Question:}} What is unusual about this image?
\\
\textbf{\textcolor{darkgreen}{Ground truth:}} Two characters on the bed are clipping into each other.
\\
\textbf{\textcolor{darkred}{MiniGPT-v2:}} 
In this scene, there are two men sleeping on a bed in a cave, and the setting is quite unusual since it takes place underground. hopefully, the characters are safe. It's not typical for two people to sleep together in such an unconventional setting, making this image stand out as odd or surprising to some viewers.
\\
\textbf{\textcolor{darkyellow}{Llama-2 (Judge):}}
The generated description conveys the key information of two characters sleeping on a bed, but it also includes some extraneous details that are not present in the ground truth, such as the location of the scene being underground and the characters' safety. However, the description does not accurately convey the theme of the characters clipping into each other, which is a significant aspect of the ground truth.
\\
There are some synonyms and paraphrases used in the generated description that attempt to convey the idea of the characters being close together, such as ``sleeping together'' and ``unconventional setting.'' However, these phrases do not accurately represent the intended meaning of the ground truth, which is that the characters are physically touching each other.
\\
Overall, the generated description does not faithfully represent the meaning of the ground truth, and it includes some misleading details that detract from the essence of the original text. Therefore, the correlation between the two texts is weak, and the generated description does not meet the criteria for a successful comparison.
\\
Reported answer: No.
\end{ebox}
\caption{MiniGPT-v2 provides a detailed description of the image but fails to notice that two people are placed in a strange way and are clipping into each other on the bed.}
\label{suppxfig:clipping_issues_4}
\end{figure*}

\begin{figure*}[h] 
\begin{ebox}{{InstructBLIP-13B fails to detect the paper coffee cup intersecting with the person's hand.}}
\centering
\includegraphics[width=1\textwidth]{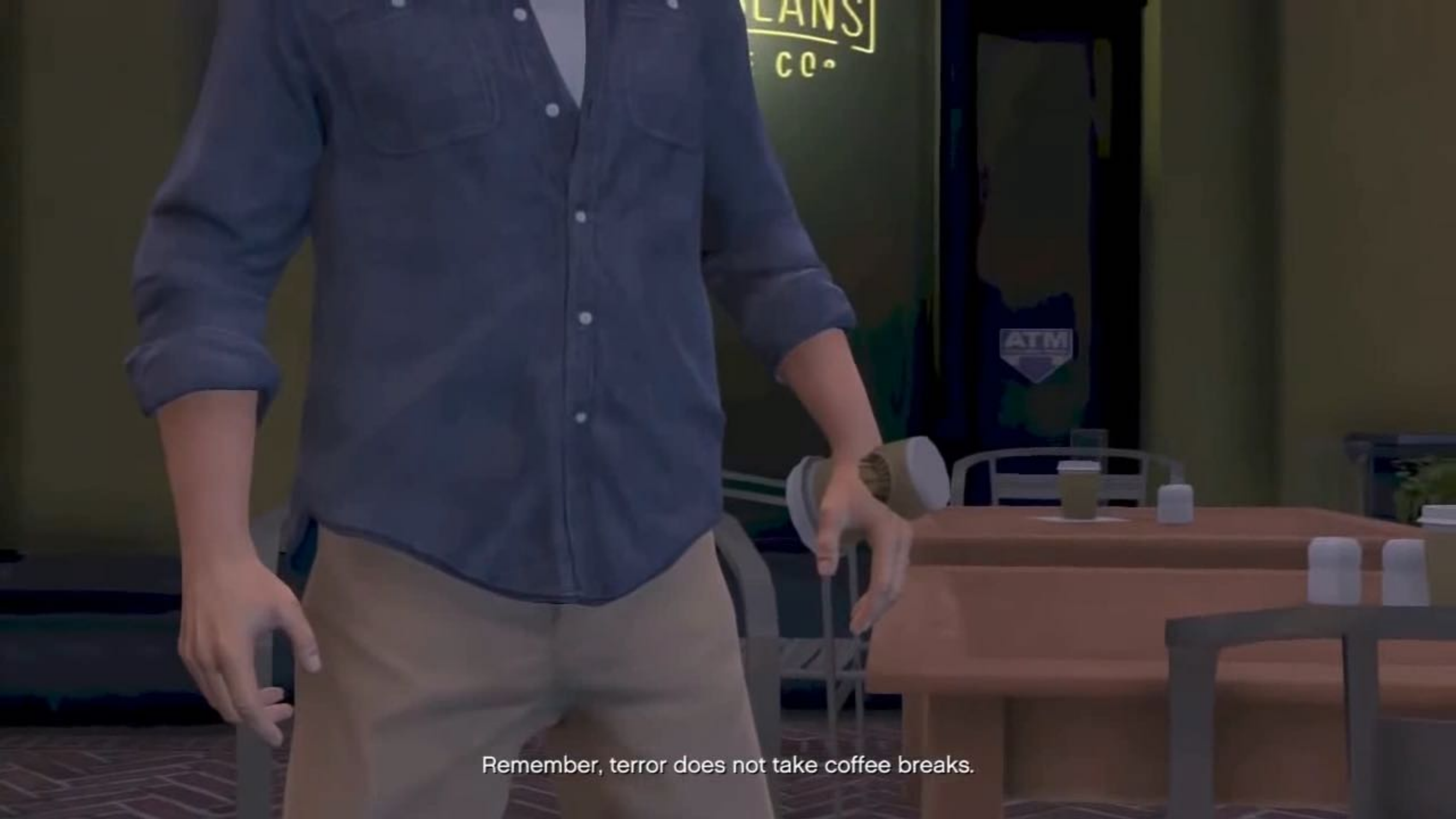}
\\
\raggedright
\textbf{\textcolor{darkblue}{Question:}} What is unusual about this image?
\\
\textbf{\textcolor{darkgreen}{Ground truth:}} A coffee cup is clipping and intersecting with a person's hand.
\\
\textbf{\textcolor{darkred}{InstructBLIP-13B:}} 
In this image, there is an unusual presence of a mannequin in the scene. The mannequin appears to be wearing a shirt and pants, and it is standing in the middle of a restaurant or cafe. It is not common to see mannequins in real-life restaurants or cafes, as they are typically used for displaying clothing or merchandise in retail stores. Therefore, the presence of a mannequin in this scene is unusual and adds an unconventional element to the image.
\\
\textbf{\textcolor{darkyellow}{Llama-2 (Judge):}}
No, the generated description does not accurately convey the key information and themes of the ground truth. The ground truth describes a scene where a coffee cup is intersecting with a person's hand, but the generated description does not mention anything about a coffee cup or a person's hand. Instead, it focuses on the presence of a mannequin in the scene, which is not the main focus of the ground truth.
\\
The generated description does mention that the mannequin is standing in a restaurant or cafe, which suggests that the image may be taking place in a similar setting, but the description does not accurately convey the key element of the ground truth. The use of a mannequin in the description is also not aligned with the ground truth, as it is not mentioned in the original text.
\\
To improve the alignment between the generated description and the ground truth, the model could be trained to focus more on the main elements of the image, such as the coffee cup and the person's hand, and less on peripheral elements like the mannequin. Additionally, the model could benefit from more specific training data that includes a variety of images with different objects and settings, to help it better recognize and describe the main elements of a scene.
\end{ebox}
\caption{The screenshot shows a person's body and a coffee cup intersecting and clipping with the person's hand. InstructBLIP mistakes the person for a mannequin and also fails to notice the coffee cup entirely}
\label{suppxfig:clipping_issues_5}
\end{figure*}

\begin{figure*}[h] 
\begin{ebox}{{Qwen-VL fails to detect that the helicopter is clipping inside the wall.}}
\centering
\includegraphics[width=1\textwidth]{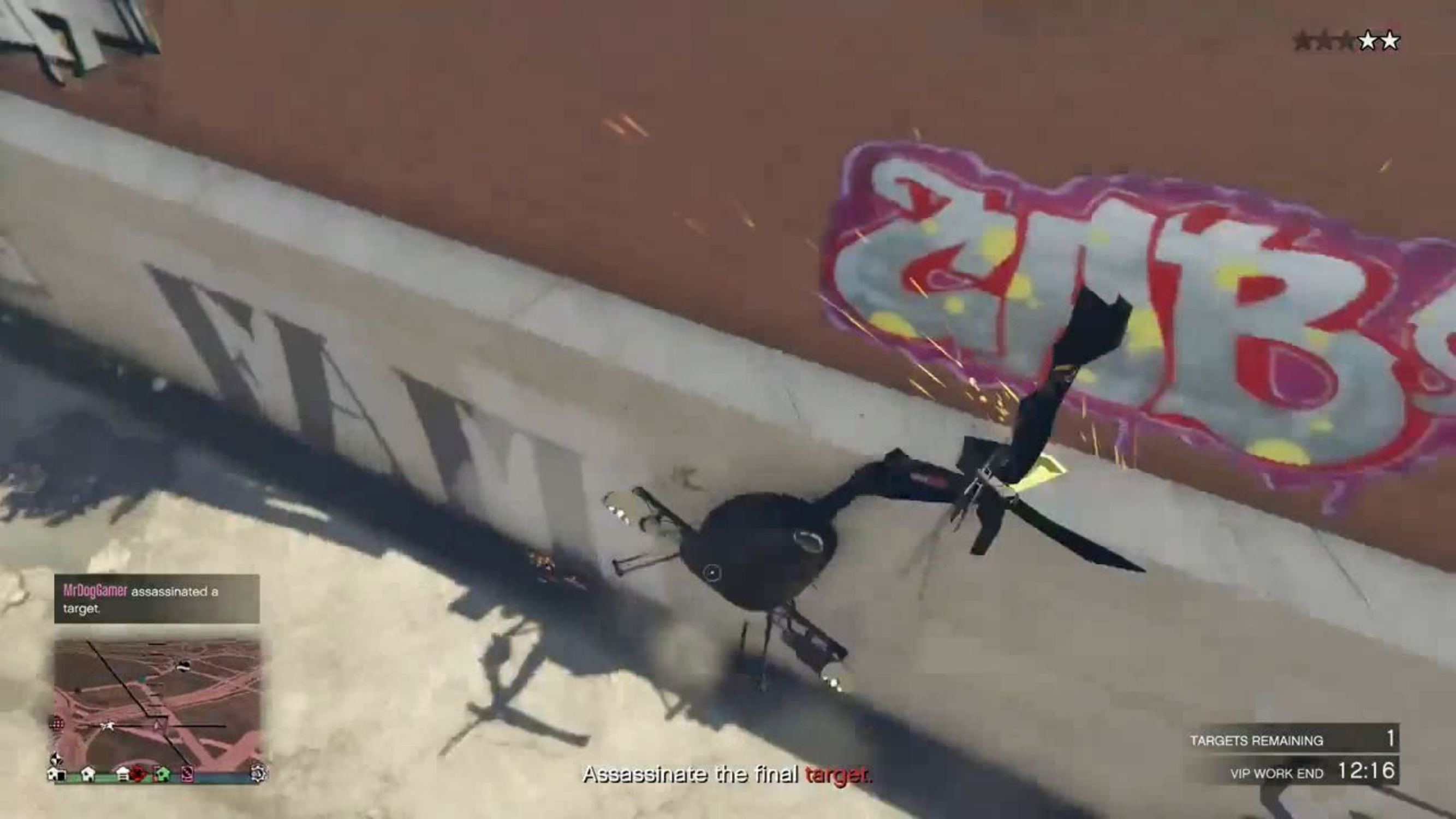}
\\
\raggedright
\textbf{\textcolor{darkblue}{Question:}} What is unusual about this image?
\\
\textbf{\textcolor{darkgreen}{Ground truth:}} A helicopter is clipping through the wall.
\\
\textbf{\textcolor{darkred}{Qwen-VL:}} 
The unusual aspect of this image is that it shows a first-person view of a person playing a video game, specifically Grand Theft Auto V. The scene depicts a person controlling a character on a rooftop, shooting at a helicopter with a gun. However, the helicopter is not a part of the real world but a digital creation within the game. This creates the illusion that the player is directly controlling the character within the game, experiencing the action and engaging in the virtual environment.
\\
\textbf{\textcolor{darkyellow}{Llama-2 (Judge):}}
No, the generated description does not accurately convey the key information and themes of the ground truth. The ground truth simply states that a helicopter is clipping through a wall, indicating a broken or glitched aspect of a digital environment. The generated description, on the other hand, describes a first-person view of a person playing a video game, specifically Grand Theft Auto V, and goes on to explain the scene in detail, highlighting the digital creation of the helicopter within the game.
\\
While the generated description provides a creative and engaging explanation of the scene, it does not faithfully represent the meaning of the ground truth. The key information and themes of the ground truth are not effectively conveyed, and the description does not accurately assess the unusual aspect of the image. Therefore, the correlation between the ground truth and the generated description is weak.
\\
To improve the alignment, the generated description could focus more on the broken aspect of the digital environment, highlighting the helicopter clipping through the wall as the unusual aspect of the image. This would ensure that the key information and themes of the ground truth are accurately conveyed, while still allowing for creative and varied linguistic expression.
\end{ebox}
\caption{Qwen-VL provides an inaccurate description of the image, including multiple hallucinations about a person standing on the rooftop. Ultimately, Qwen-VL fails to notice the helicopter clipping into the wall.}
\label{suppxfig:clipping_issues_6}
\end{figure*}


\clearpage
\subsection{Failures Related to Unusual Circumstances}
\label{sec:supp-unusual_setups}

\begin{figure*}[ht]
\begin{ebox}{{GPT-4V fails to detect multiple clipping glitches involving several cars.}}
\centering
\includegraphics[width=1\textwidth]{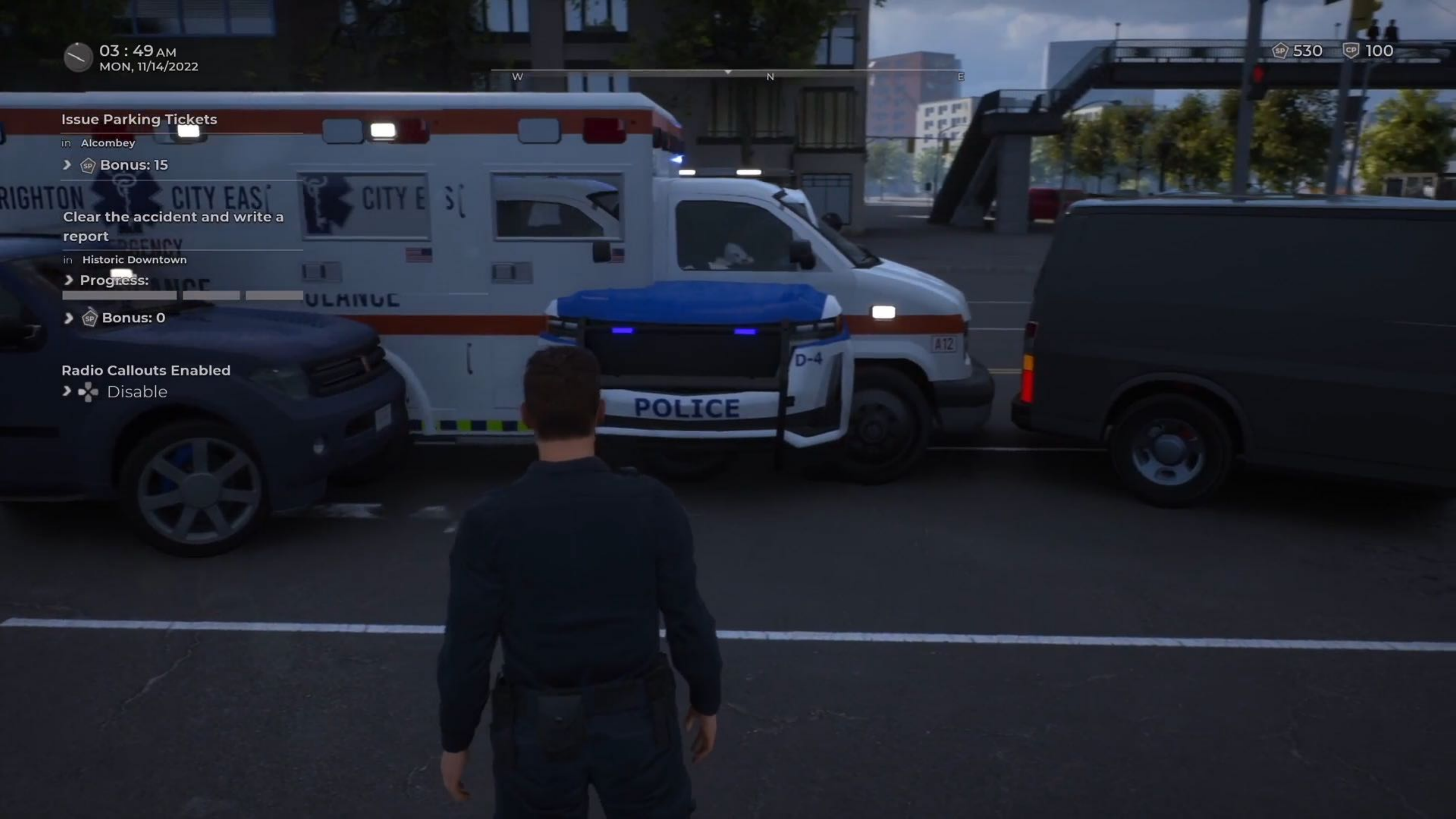}
\raggedright
\textbf{\textcolor{darkblue}{Question:}} What is wrong with this image?
\\
\textbf{\textcolor{darkgreen}{Ground truth:}} Several cars are colliding with each other.
\\
\textbf{\textcolor{darkred}{GPT-4V:}} 
This image appears to be from a video game, judging by the graphics and the user interface elements visible on the screen, such as the time, a mini-map, and game objectives related to issuing parking tickets and clearing an accident. What's unusual here is that the police vehicle is incorrectly parked—it's on the wrong side of the road, facing oncoming traffic, which goes against typical traffic laws. Additionally, the rear of the ambulance is opened, but there is no sense of urgency or activity that would typically accompany an open ambulance at a scene, contributing to the unusual nature of the scene depicted.
\\
\textbf{\textcolor{darkyellow}{Llama-2 (Judge):}}
No, the generated description does not accurately convey the key information and themes of the ground truth. While it provides some details about the scene, such as the presence of a police vehicle and an ambulance, it does not mention the central aspect of the image, which is the collision of several cars. Additionally, the description focuses on the unusual aspects of the scene, such as the parked police vehicle and the open ambulance, which are not the main focus of the ground truth.
\\
The generated description could be improved by first acknowledging the central theme of the image, which is the collision of multiple cars, and then providing additional details about the scene, such as the presence of the police vehicle and ambulance, while ensuring that the description accurately reflects the content of the image. A possible revised description could be: ``This image depicts a chaotic scene with multiple cars colliding with each other. In the foreground, a police vehicle is parked on the wrong side of the road, while an ambulance is parked nearby with its rear opened. The scene appears to be from a video game, as indicated by the graphics and user interface elements visible on the screen.''
\end{ebox}
\caption{The screenshot shows two ambulance cars and a police car colliding and intersecting with each other. Two ambulance cars are moving in the same direction, following the street traffic, but the police car is rotated 90 degrees, as if it was blocking the road. However, both ambulances are colliding with the police car. GPT-4V fails to detect these clipping issues.}
\label{suppfig:police_car_parked}
\end{figure*}

\begin{figure*}[ht]
\begin{ebox}{{GPT-4V is confused by the light emitted from the furnace and mistakes it for a thermal monitor.}}
\centering
\includegraphics[width=1\textwidth]{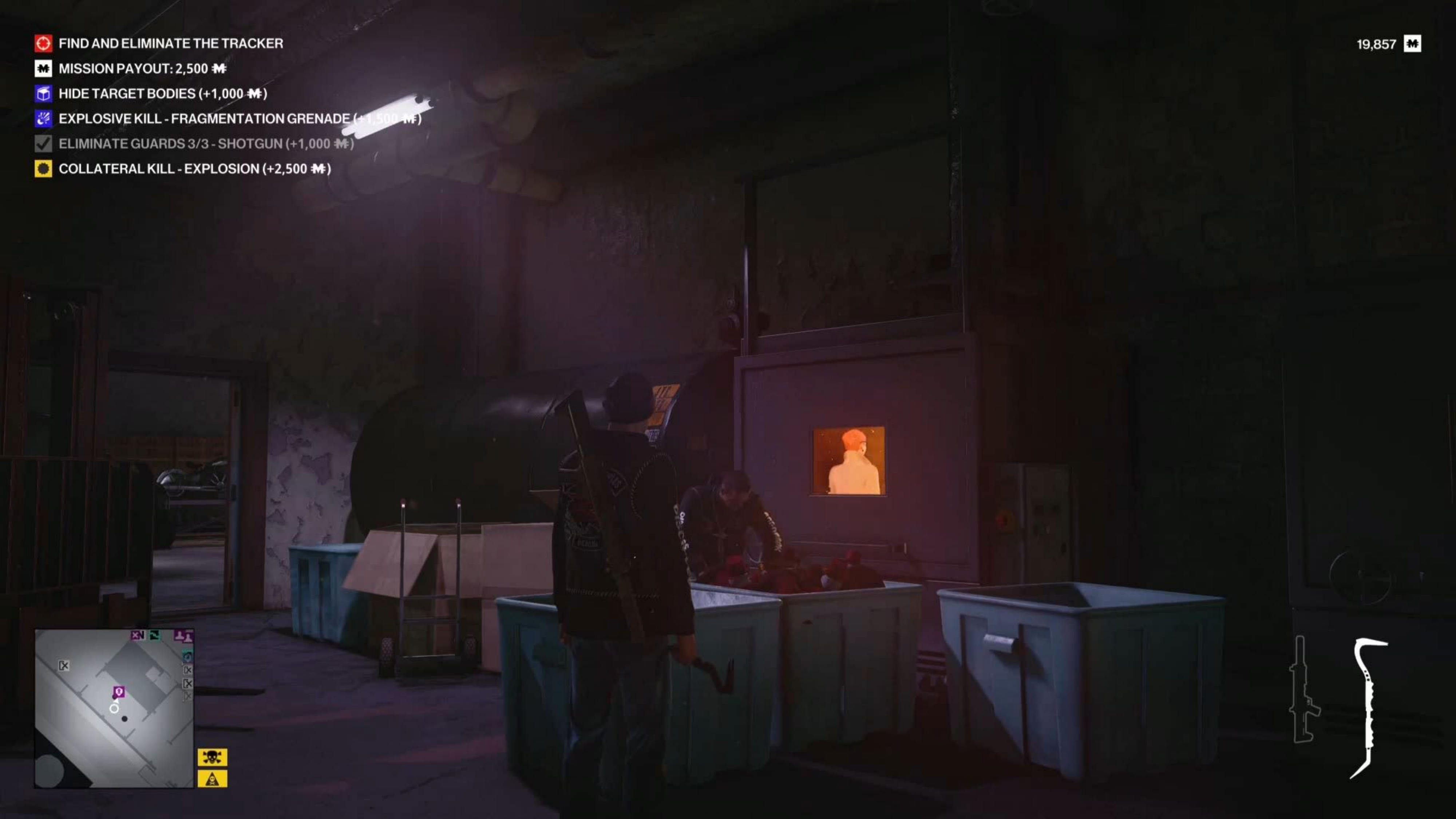}
\raggedright
\textbf{\textcolor{darkblue}{Question:}} What is unusual about this image?
\\
\textbf{\textcolor{darkgreen}{Ground truth:}} A person is standing in the furnace.
\\
\textbf{\textcolor{darkred}{GPT-4V:}} 
The image appears to be a screenshot from a video game, judging from the HUD (Heads-Up Display) elements present on the screen. It depicts a character in a dark environment with objectives and game information listed on the left side of the screen. The unusual elements in this image relate to the context of a video game world rather than real life:

\begin{enumerate}
  \item \textbf{Objective List}: On the left-hand side, there are specific objectives listed like ``FIND AND ELIMINATE THE TRACKER'' and ``HIDE TARGET BODIES,'' which are not real-world tasks but typical of assassination or stealth-based video games.

  \item \textbf{Currency and Scoring}: The top right corner shows a currency or score indicator with ``19,857,'' which suggests the player has amassed that amount of in-game currency or points.

  \item \textbf{Weapons Display}: The bottom right corner shows silhouetted images of weapons, indicating the player's current loadout or available equipment.

  \item \textbf{Thermal Image}: There is a thermal imaging display showing a figure in red, which suggests the use of some kind of thermal vision or scanning technology within the game to detect characters through obstacles or walls.

  \item \textbf{Radar}: In the bottom left corner, there is a mini-map or radar that shows the player's immediate surroundings, including the location of other characters, represented by red and white icons.
\end{enumerate}

The image as a whole is not reflective of a real-life scene but rather is constructed for an interactive gaming experience, where such displays and objectives are part of the gameplay mechanics.
\\
\textbf{\textcolor{darkyellow}{Llama-2 (Judge):}}
No, the generated description does not accurately convey the key information and themes of the ground truth reference.
\\
The ground truth reference simply states that a person is standing in a furnace, which implies a dangerous and potentially industrial or manufacturing environment. In contrast, the generated description focuses on the visual elements of a video game, such as the HUD, objectives, currency, weapons, thermal image, and radar. While it acknowledges that the image appears to be from a video game, it does not address the main theme of the person in the furnace.
\end{ebox}
\caption{The screenshot displays an unusual setting where a person stands in a furnace with glowing fire. GPT-4V fails to correctly detect and understand the scene, mistaking the furnace for a thermal monitor.}
\label{suppfig:thermals}
\end{figure*}

\begin{figure*}[ht]
\begin{ebox}{{GPT-4V fails to notice that a car is not being rendered at all, while its passengers are}}
\centering
\includegraphics[width=1\textwidth]{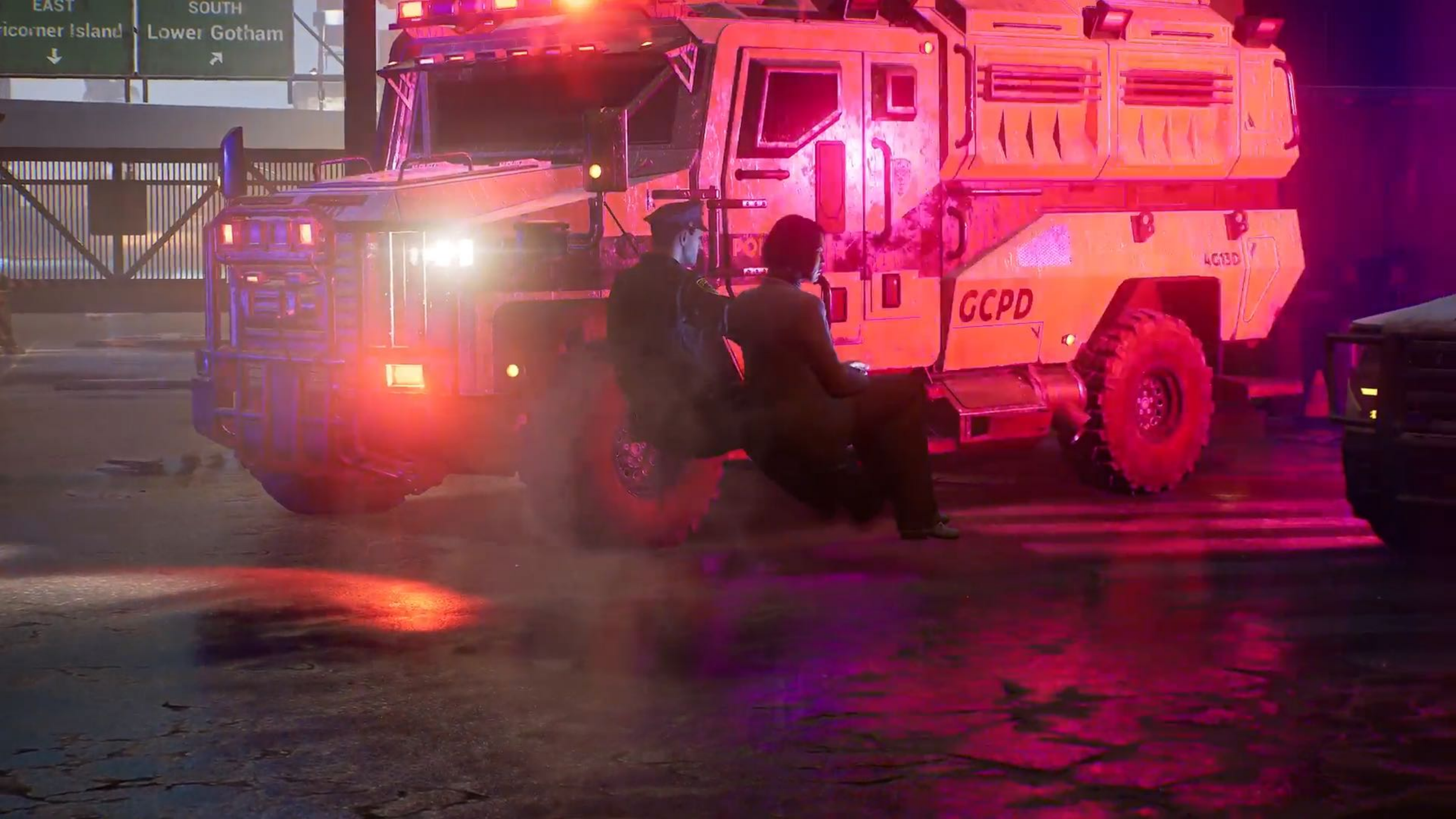}
\raggedright
\textbf{\textcolor{darkblue}{Question:}} What is wrong with this image?
\\
\textbf{\textcolor{darkgreen}{Ground truth:}} Two people driving an invisible car
\\
\textbf{\textcolor{darkred}{GPT-4V:}} 
The image seems to be a still from a video game or a CGI-rendered scene, showing a police armored vehicle with ``GCPD'' written on it, suggesting it is related to the fictional Gotham City Police Department from the Batman universe. There are two characters crouching beside the vehicle. It's difficult to say exactly what might be considered ``wrong'' with this image without more context, as it could refer to technical issues, inconsistencies within the narrative of the game, or anachronisms. If there's something specific you'd like me to look for, please provide additional details.
\\
\textbf{\textcolor{darkyellow}{Llama-2 (Judge):}}
No, the generated description does not faithfully represent the ground truth.
\\
The ground truth describes two people driving an invisible car, while the generated description describes a police armored vehicle with two characters crouching beside it, possibly from a video game or CGI-rendered scene. The generated description does not mention anything about an invisible car or the characters driving it, which is the main focus of the ground truth.
\\
The generated description does note that the image could be from a video game or CGI-rendered scene, which is a good start, but it does not provide any specific information about the content of the image that aligns with the ground truth. Additionally, the mention of technical issues, inconsistencies within the narrative of the game, or anachronisms suggests that the model is not sure what is wrong with the image, which is not helpful in evaluating its alignment with the ground truth.
\\
To improve the alignment, the model could be trained on a dataset that includes images of invisible cars and pulling information from a variety of sources to ensure that the information provided is accurate and relevant to the task. Additionally, the model could be designed to focus on the main elements of the image, such as the characters and the car, rather than peripheral details like the police armored vehicle.
\end{ebox}
\caption{The screenshot show two people  are driving an invisible car due to a rendering glitch. GPT-4V fails to notice the strange character postures, which suggest that they are participating in an animation where they are riding in a car.}
\label{suppfig:invisible_car}
\end{figure*}


\clearpage
\section{Sample Glitches that are Hard for Humans}
\label{sec:supp-sample_hard_for_humn}

In this section, we provide some samples for which human users find it difficult to detect or report the glitch correctly.
There are some glitches that are not easy for humans to report or detect. We can roughly categorize these glitches into two groups:

\begin{enumerate}
  \item Lay users cannot report them using the correct terminology. For example, ``Hall of Mirrors'', which refers to cases where textures and images are incorrectly reflected multiple times, creating a disorienting, mirror-like effect.
  \item Users may not notice glitches due to poor visibility, lighting, or rendering conditions.
\end{enumerate}

\begin{figure}[htp]
    \centering
    \includegraphics[width=\linewidth]{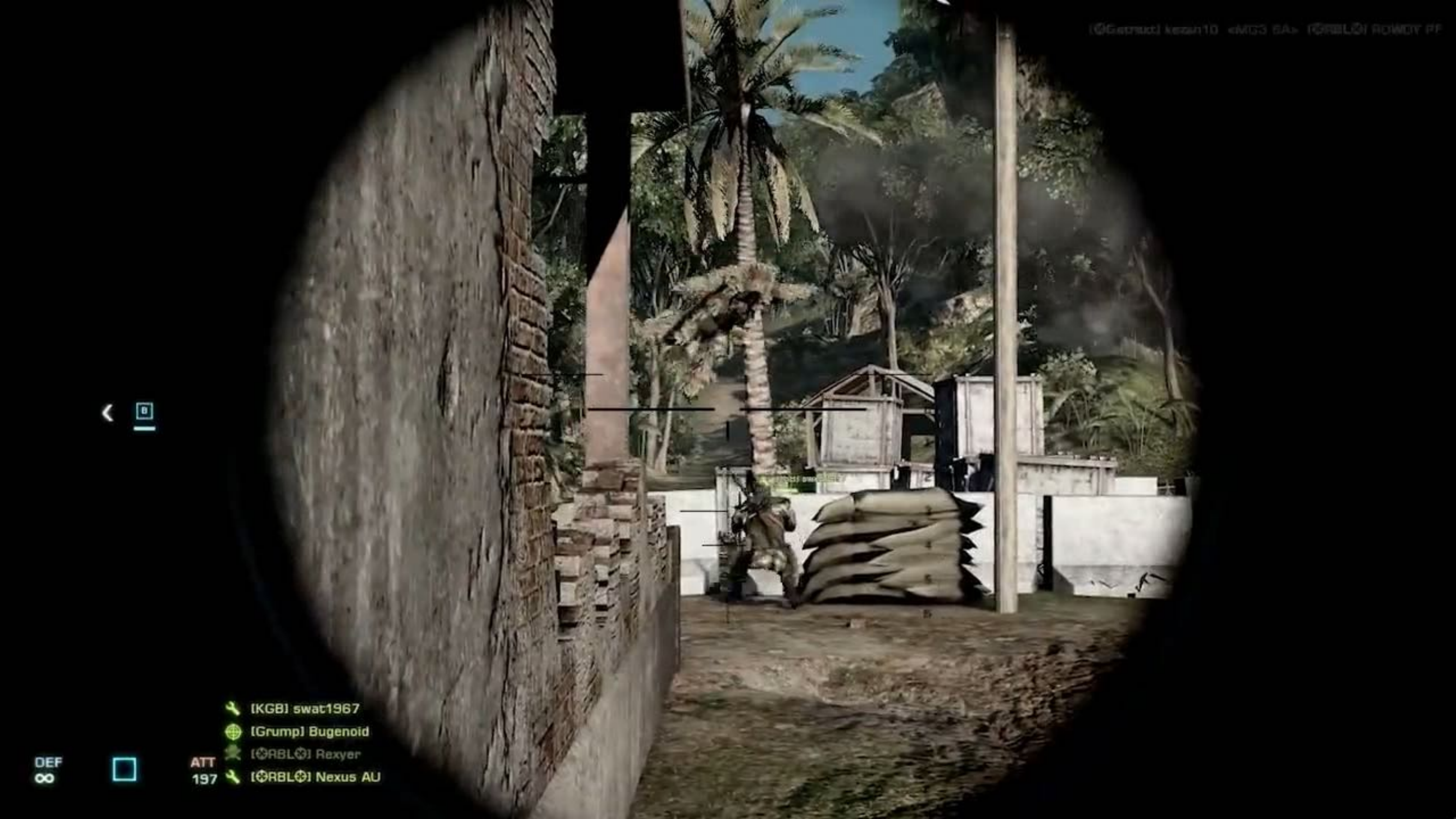}
\caption{The screenshot shows a person dressed in a sniper suit floating in the air near the center of the image (above the crosshair). Detecting the floating person can be challenging for some users due to the pattern used in the sniper suit, the background palm tree, and the overall color of the environment.}
    \label{suppxfig:hard_images_human_1}
\end{figure}

\begin{figure}[htp]
    \centering
    \includegraphics[width=\linewidth]{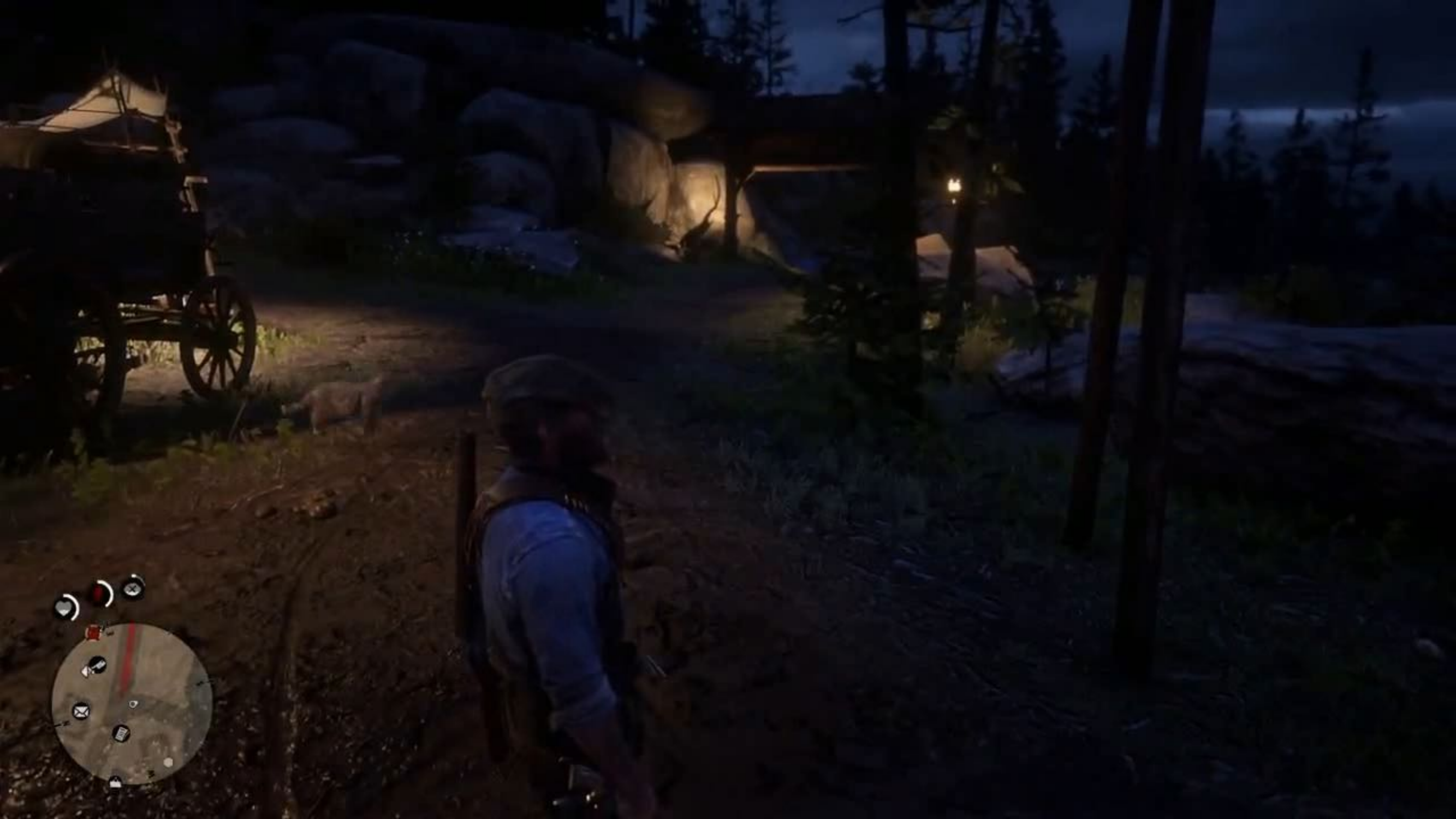}
\caption{In this image, the cat on the left side of the image is slightly above the ground and floating in mid-air. Due to the lighting conditions and distance of the cat from the camera, detecting the glitch is hard.}
    \label{suppxfig:hard_images_human_4}
\end{figure}

\begin{figure}[htp]
    \centering
    \includegraphics[width=\linewidth]{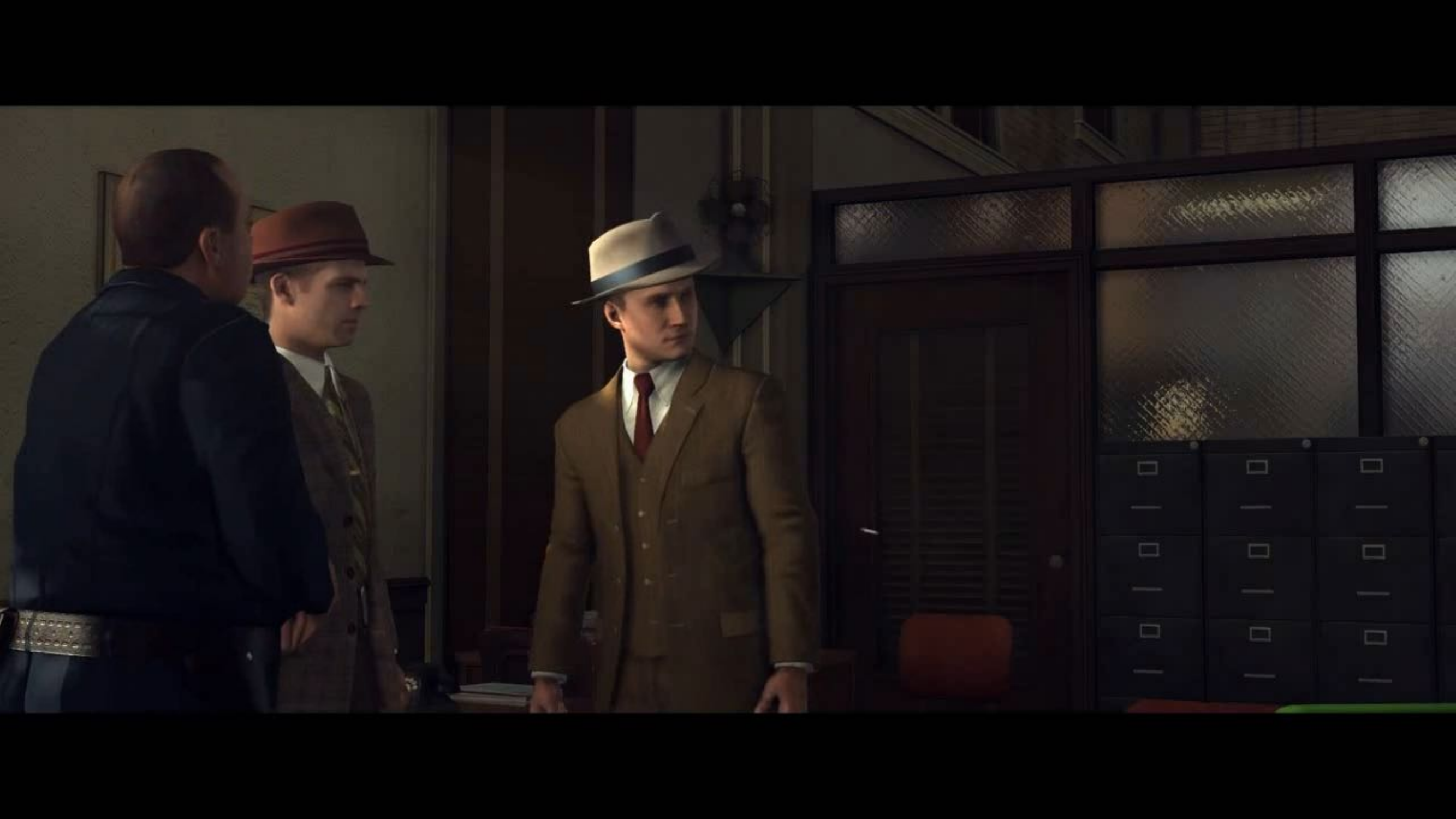}
\caption{In this image, there is a character smoking a cigarette on the right side, but due to a rendering glitch, the character is not rendered at all; only the cigarette is visible. Detecting the absence of the character can be challenging for some users.}
    \label{suppxfig:hard_images_human_5}
\end{figure}

\begin{figure}[htp]
    \centering
    \includegraphics[width=\linewidth]{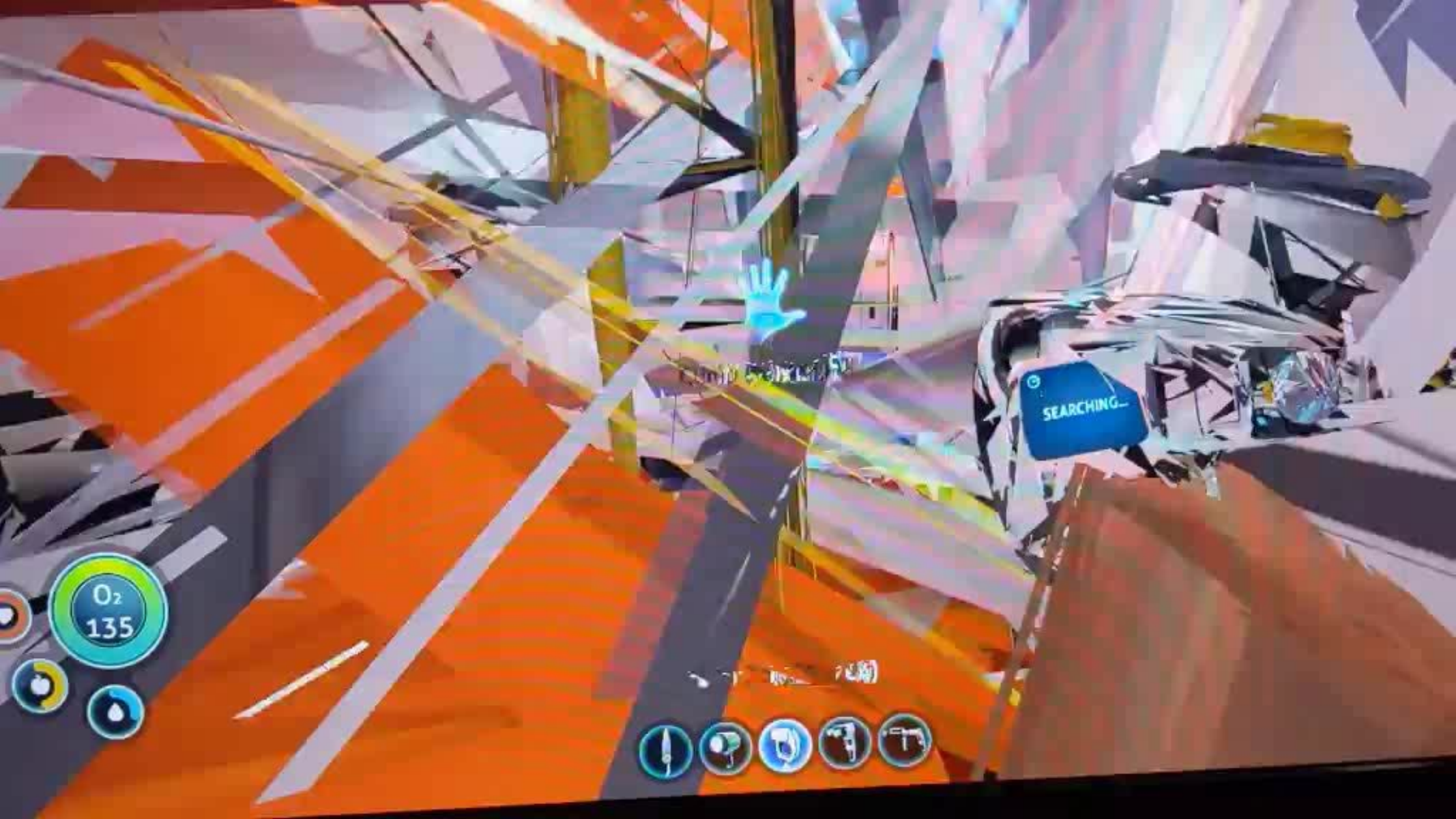}
\caption{The image shows a significant rendering glitch in which vertices and triangles of the object are completely corrupted. Describing what is wrong with this image can be challenging for some users as they do not know specific terminology.}
    \label{suppxfig:hard_images_human_3}
\end{figure}

\begin{figure}[htp]
    \centering
    \includegraphics[width=\linewidth]{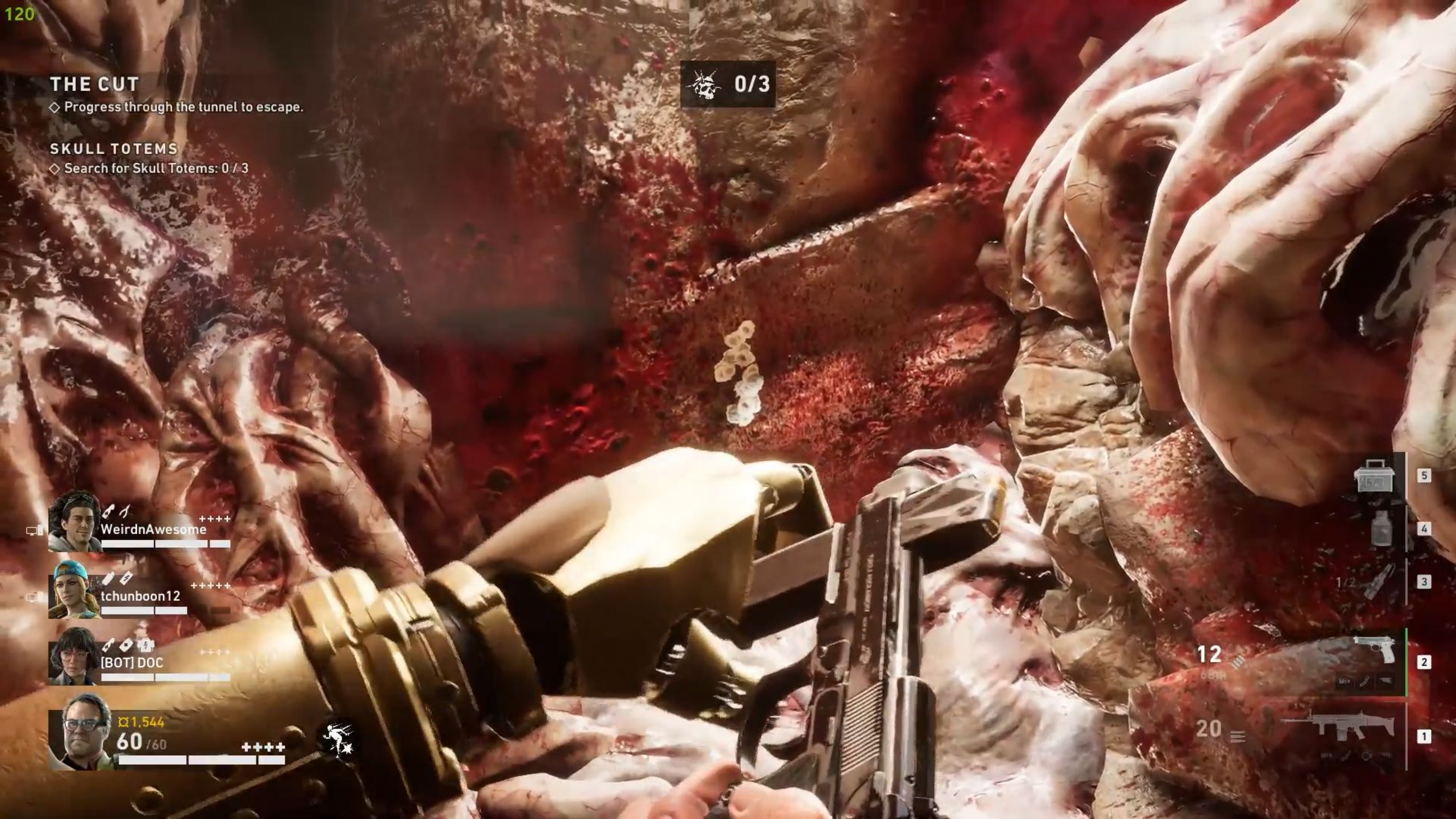}
\caption{The image shows a blood and gore setting in a zombie-related game, with zombie intestines all over the place. Some users fail to notice that the hands are reloading a gun, but the magazine is being put into the wrong part of the gun, resulting in a clipping glitch.}
    \label{suppxfig:hard_images_human_2}
\end{figure}

\begin{figure}[htp]
    \centering
    \includegraphics[width=\linewidth]{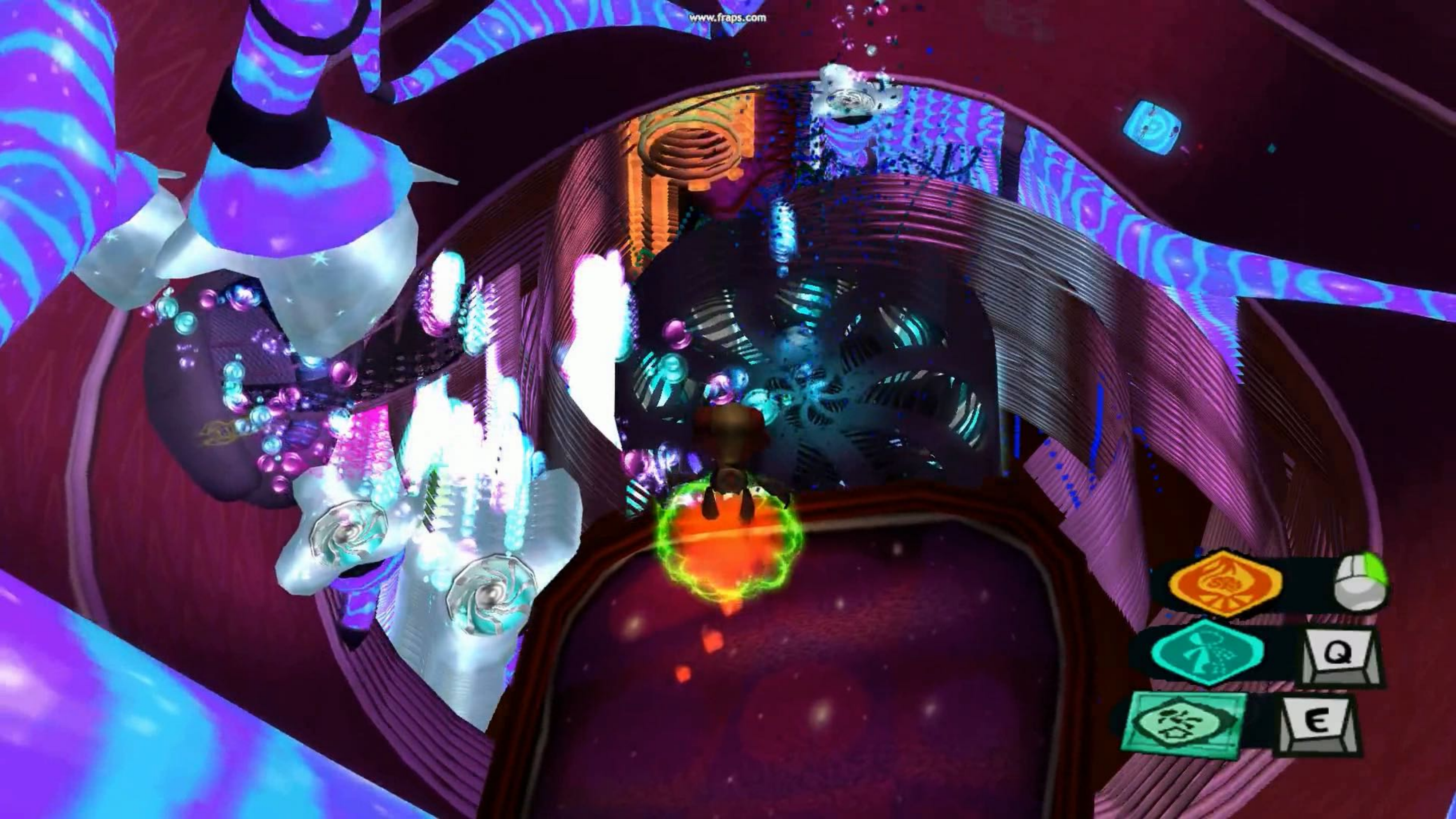}
\caption{The image depicts a rendering glitch known as ``Hall of Mirrors'' or ``ghosting'', which results in a trail of previously rendered frames appearing instead of a missing mesh or texture. While detecting that there are some issues with the image is easy for most users, using the correct terms can be challenging.}
    \label{suppxfig:hard_images_human_6}
\end{figure}


\clearpage
\section{Synthetic Sample Generated with Unity}
\label{sec:supp-unitysamples}

In this section, we provide samples of glitches generated inside Unity.

\begin{figure}[h!]
    \centering
    \begin{minipage}{.495\textwidth}
        \centering
        \includegraphics[width=\linewidth]{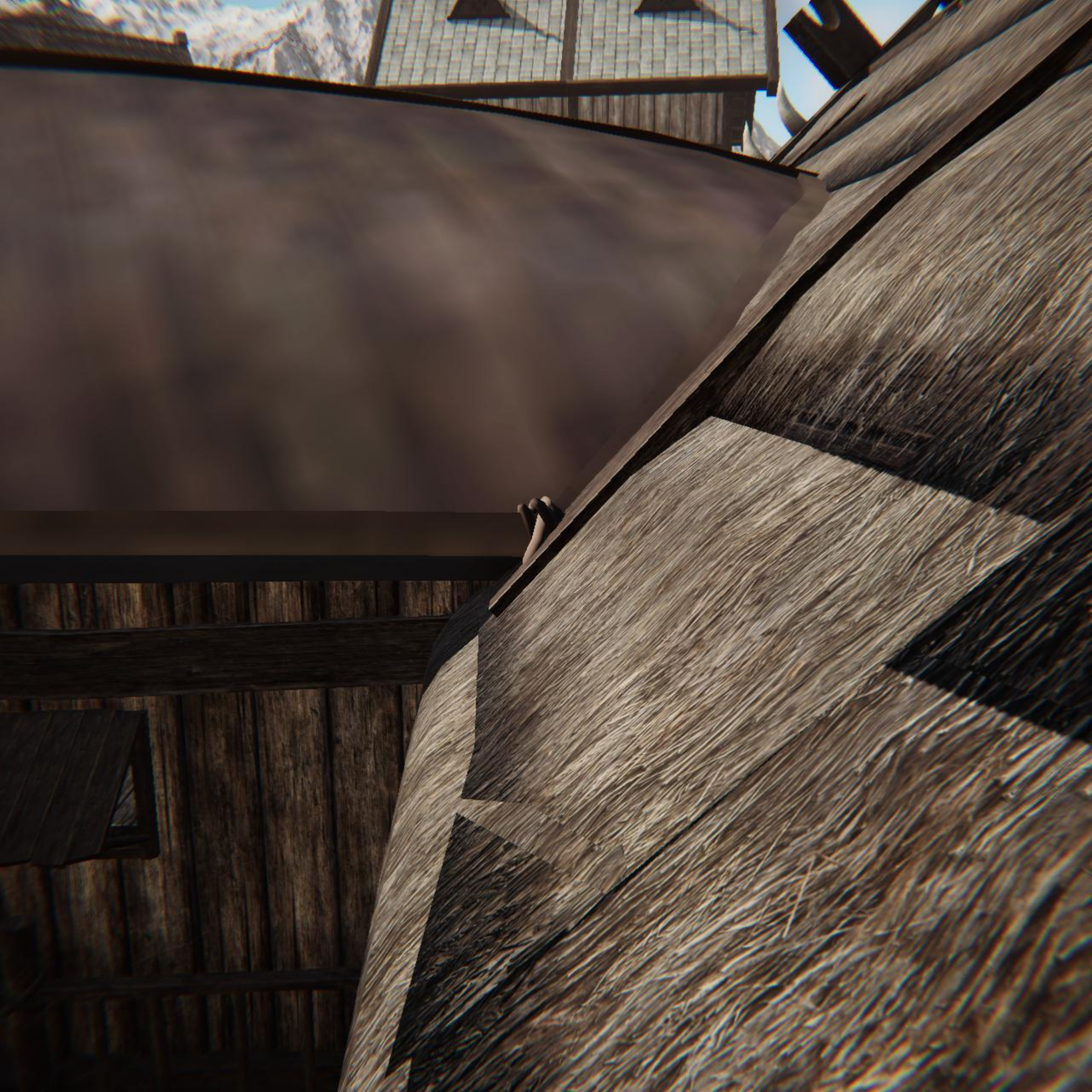}
        \caption{The roof has a low-resolution texture.}
    \end{minipage}
    \hfill
    \begin{minipage}{.495\textwidth}
        \centering
        \includegraphics[width=\linewidth]{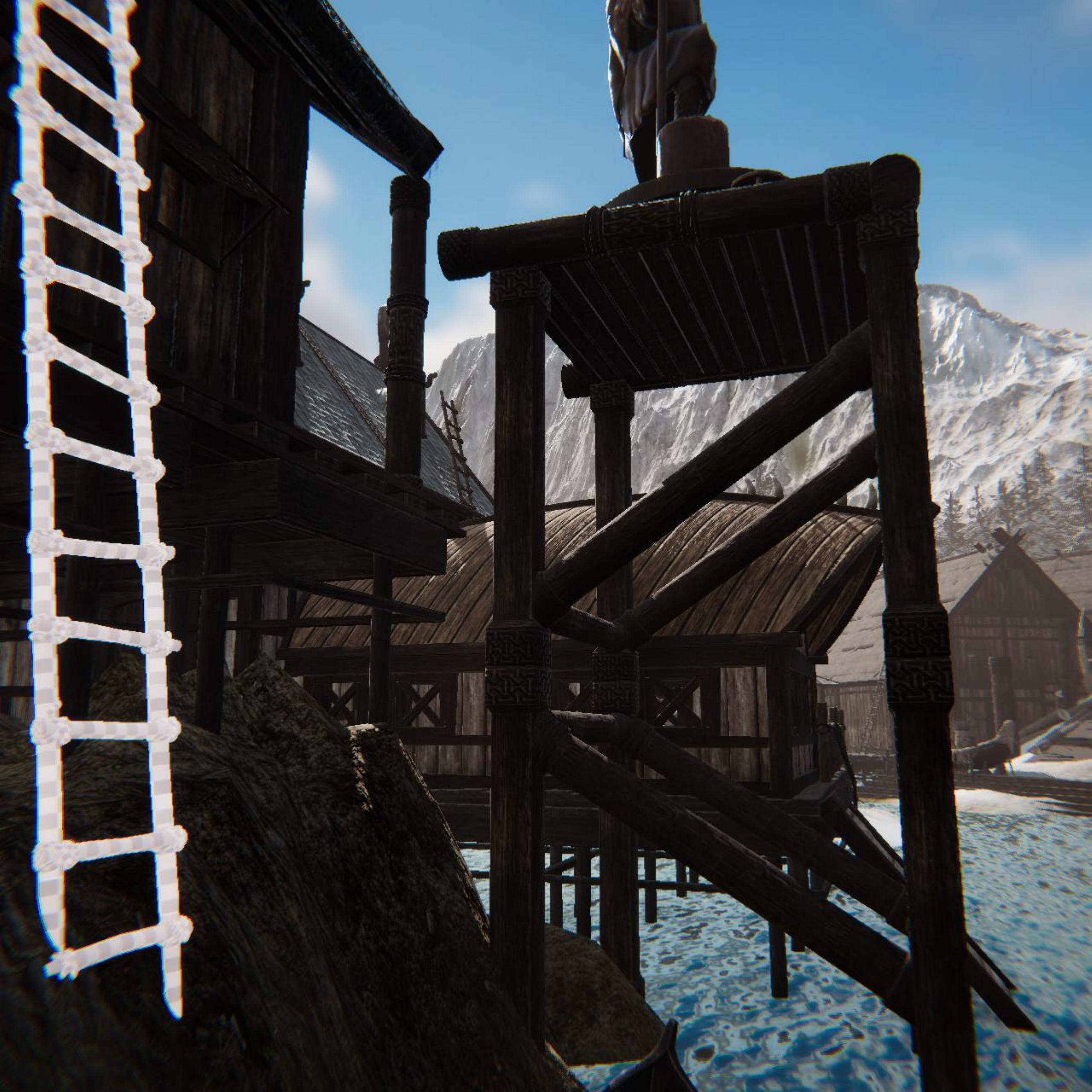}
        \caption{The ladder has a placeholder texture.}
    \end{minipage}
    \newline
    \begin{minipage}{.495\textwidth}
        \centering
        \includegraphics[width=\linewidth]{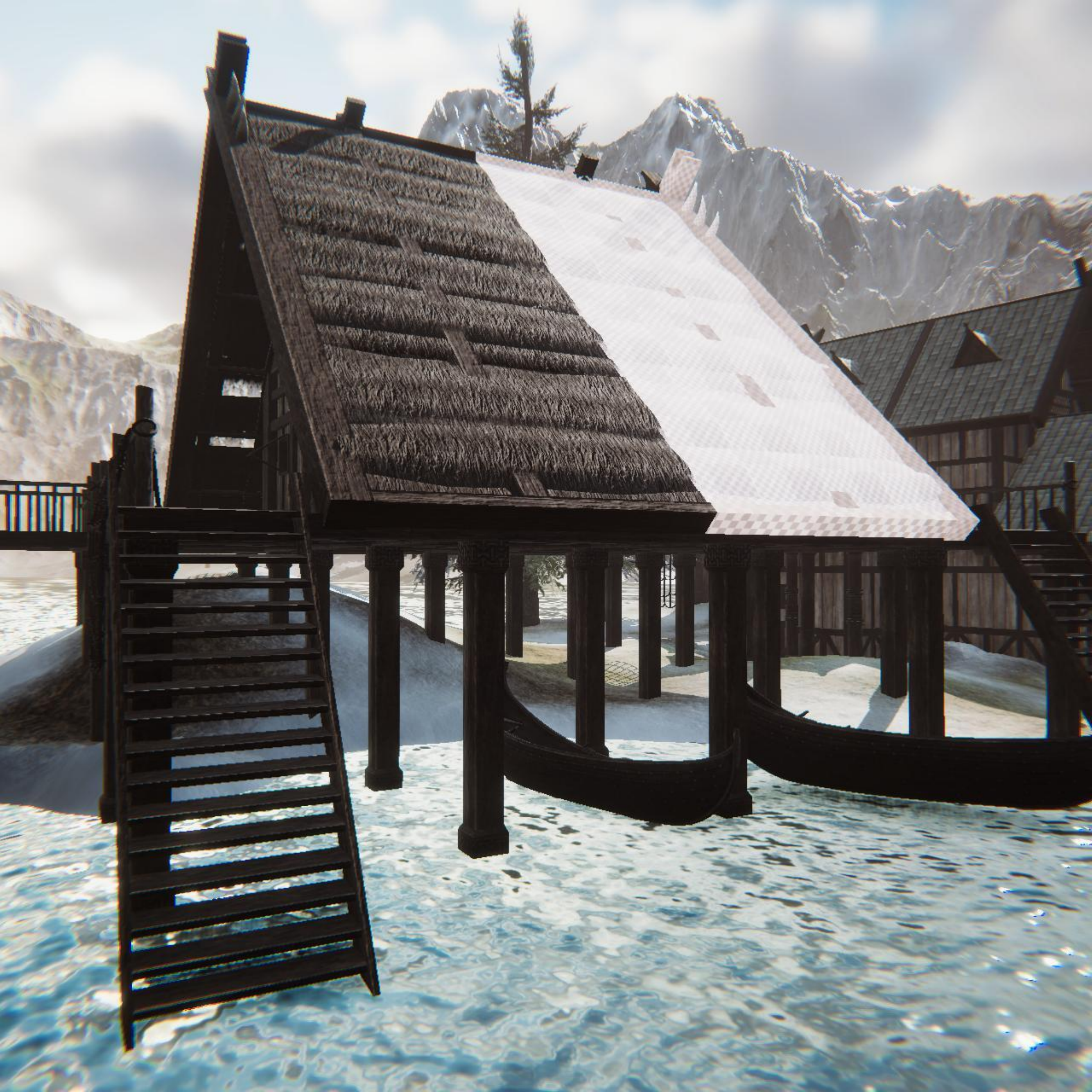}
        \caption{Part of the roof has a placeholder texture.}
    \end{minipage}
    \hfill
    \begin{minipage}{.495\textwidth}
        \centering
        \includegraphics[width=\linewidth]{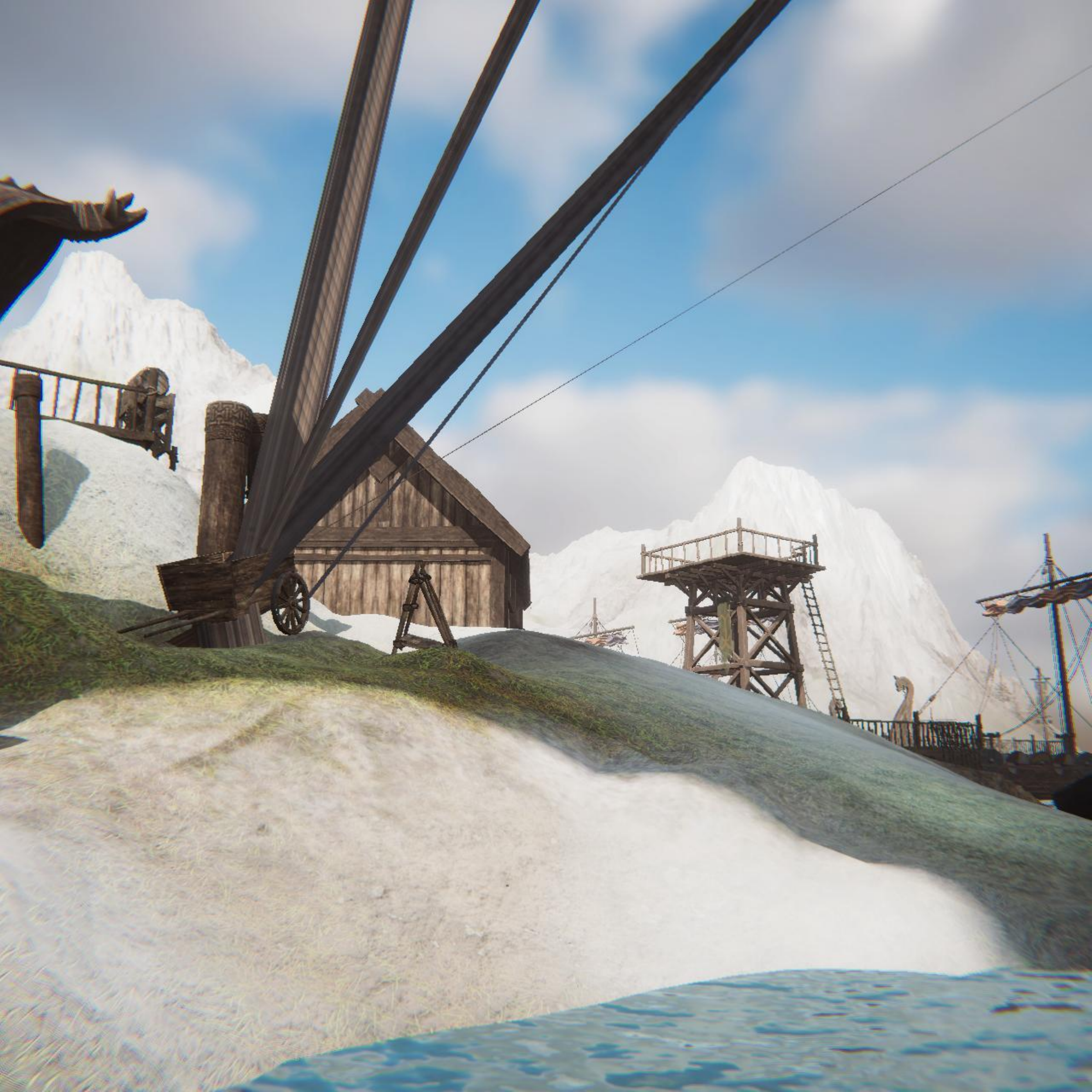}
        \caption{The carriage has a distorted mesh.}
    \end{minipage}
\end{figure}

\begin{figure}[h!]
    \centering
    \begin{minipage}{.495\textwidth}
        \centering
        \includegraphics[width=\linewidth]{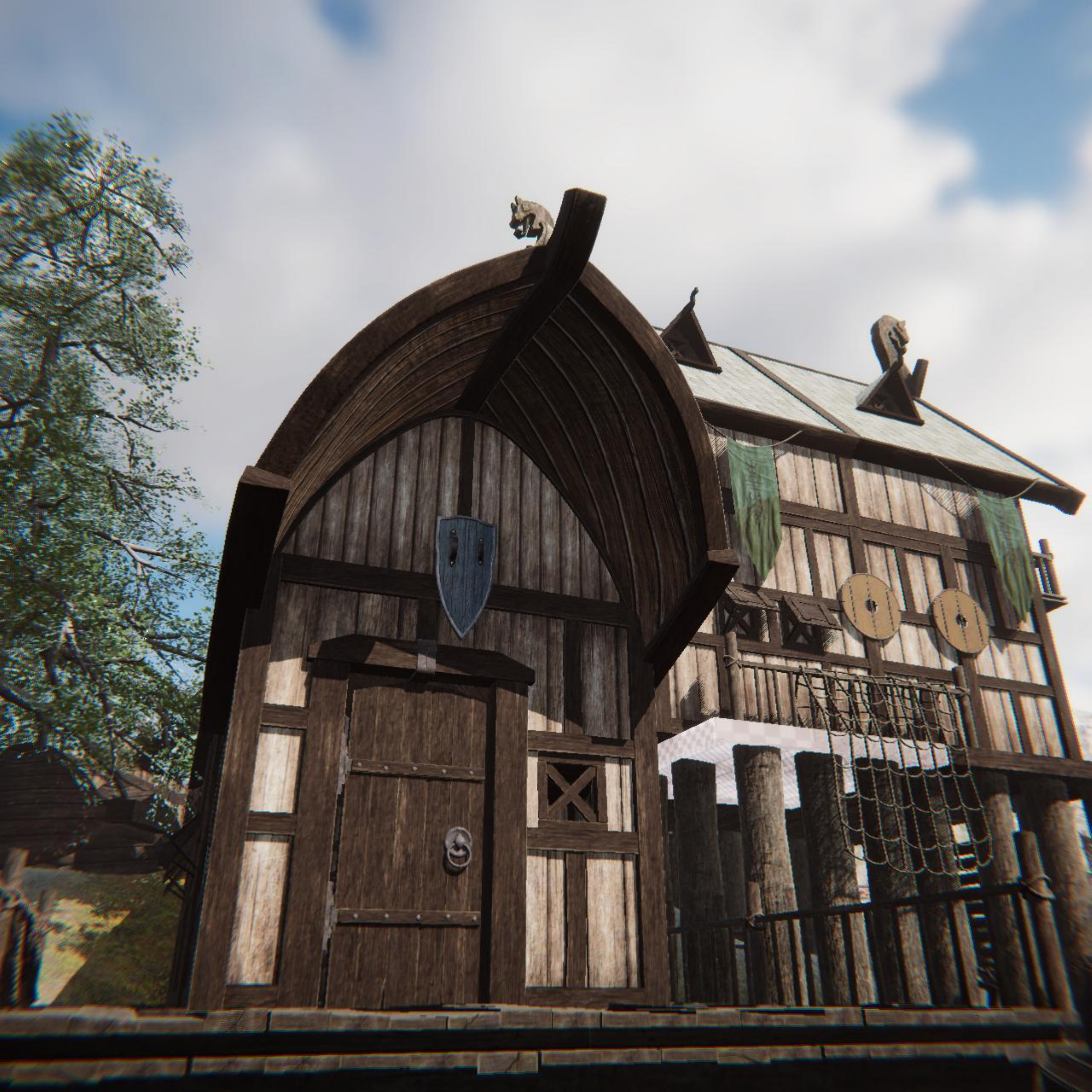}
        \caption{Part of the house structure has a placeholder texture.}
    \end{minipage}
    \hfill
    \begin{minipage}{.495\textwidth}
        \centering
        \includegraphics[width=\linewidth]{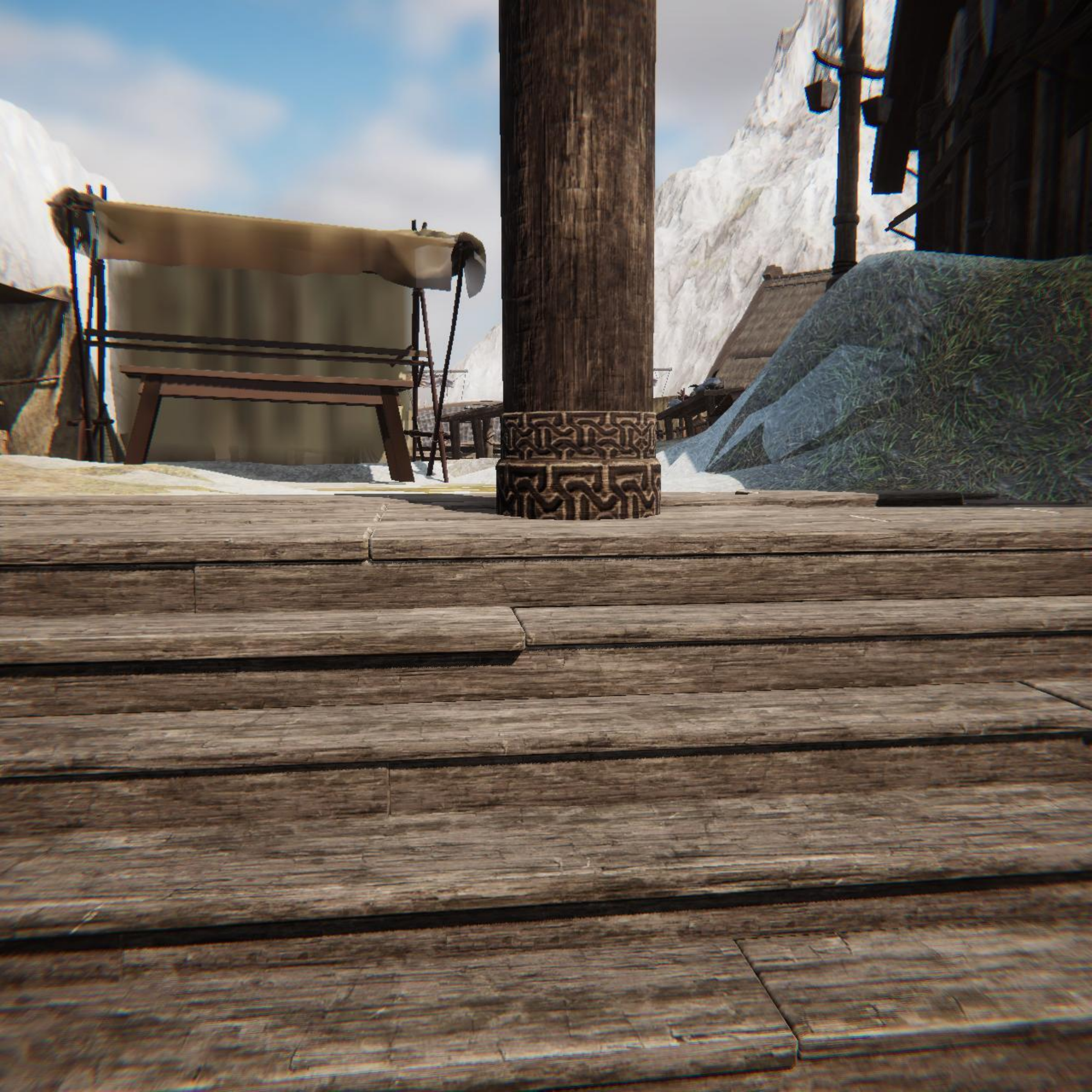}
        \caption{The canopy structure has a low-resolution texture.}
    \end{minipage}
    \newline
    \begin{minipage}{.495\textwidth}
        \centering
        \includegraphics[width=\linewidth]{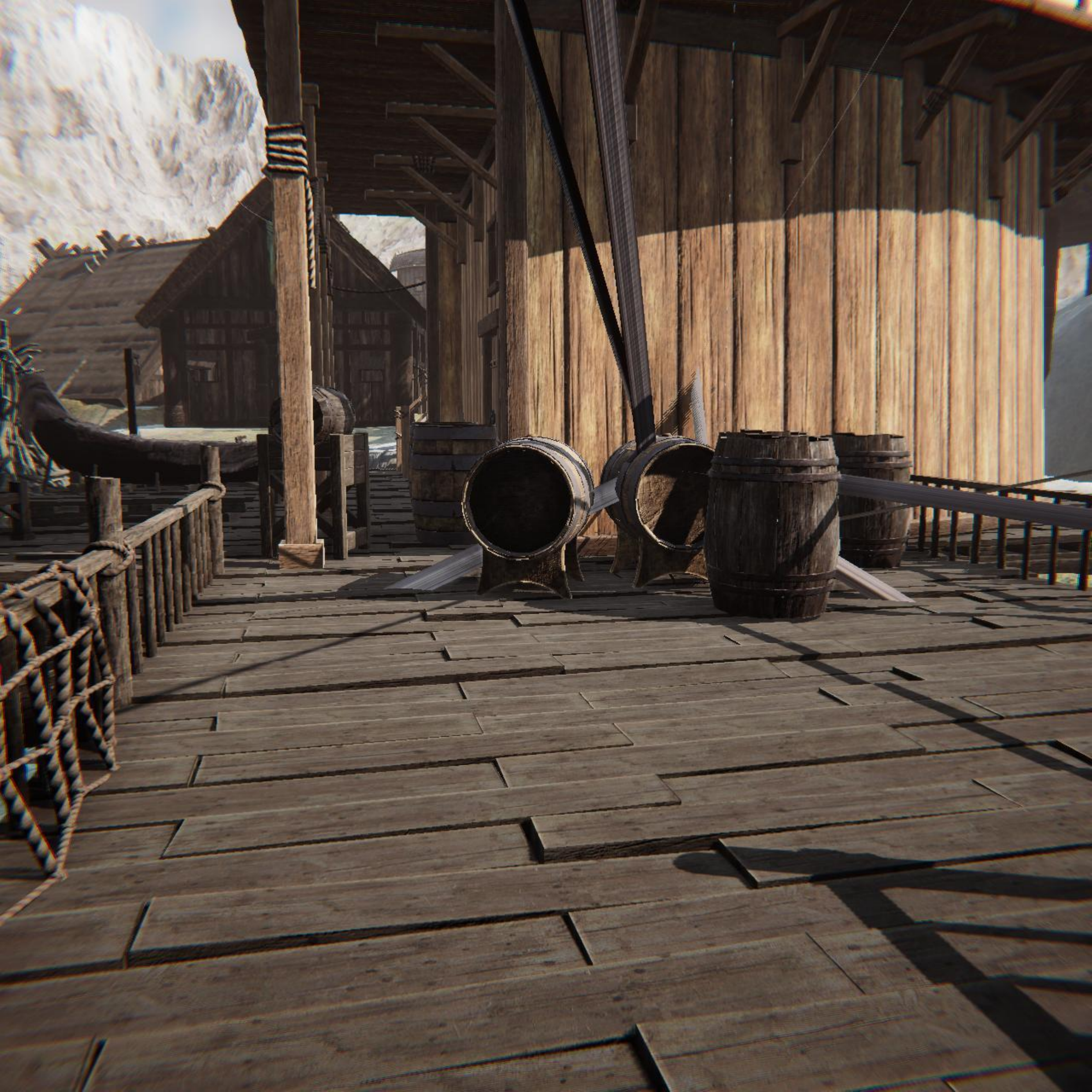}
        \caption{The barrel's mesh is stretched and distorted.}
    \end{minipage}
    \hfill
    \begin{minipage}{.495\textwidth}
        \centering
        \includegraphics[width=\linewidth]{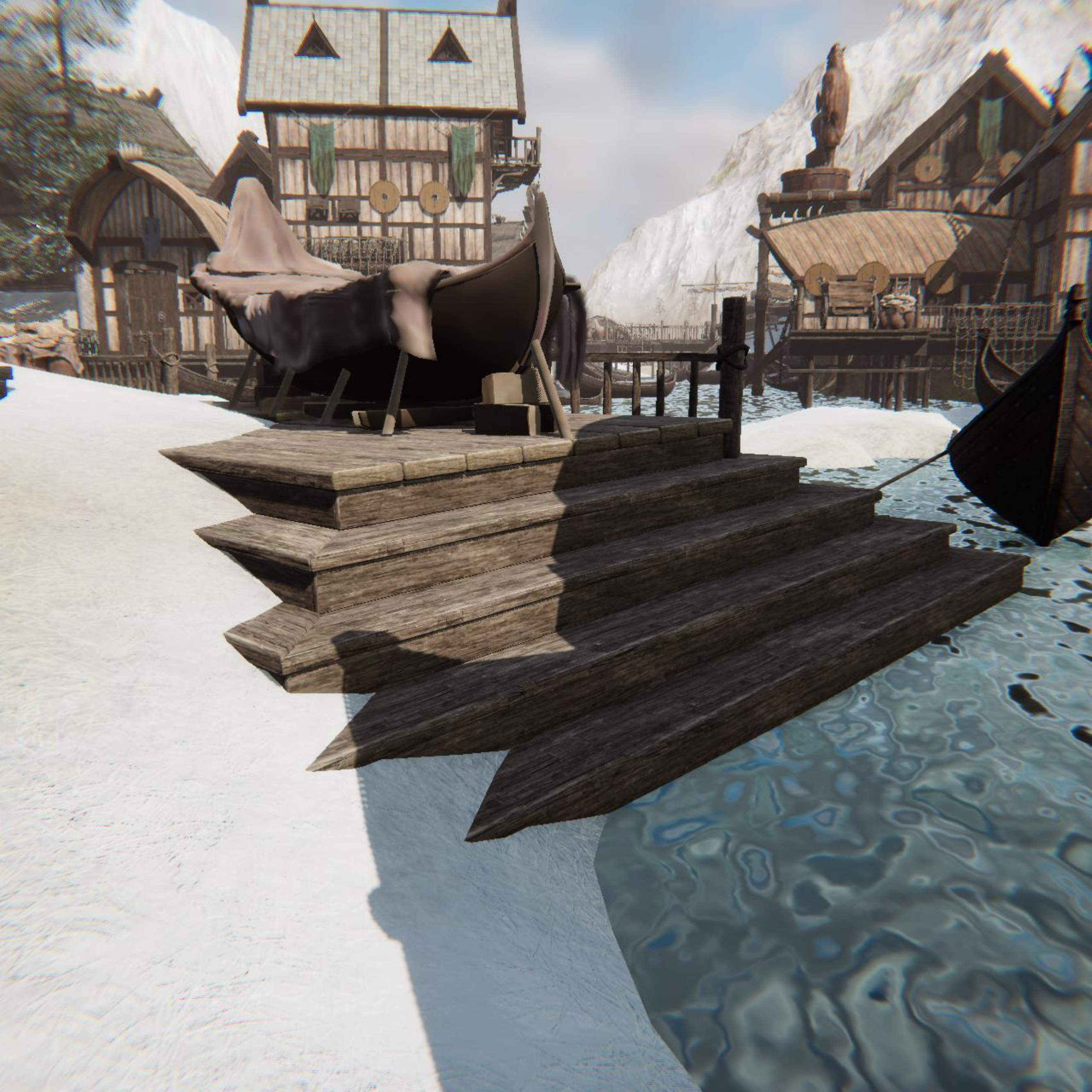}
        \caption{The boat has a low-resolution texture.}
    \end{minipage}
\end{figure}

\end{document}